\documentclass{ieeeaccess}
\usepackage{cite}
\usepackage{threeparttable}
\usepackage{amsmath,amssymb,amsfonts}
\usepackage{algorithmic}
\usepackage{graphicx}
\usepackage{textcomp}
\usepackage{rotating}
\usepackage{makecell} 
\usepackage{bm}
\usepackage{booktabs} 
\usepackage{hyperref} 
\usepackage{colortbl}
\usepackage[table,xcdraw]{xcolor}
\definecolor{DNNColor}{RGB}{220,240,255} 
\definecolor{SNNColor}{RGB}{255,230,230}  
\hypersetup{
  colorlinks=true,
  linkcolor=blue,
  citecolor=blue,
  filecolor=magenta,
  urlcolor=cyan,
}

\makeatletter
\AtBeginDocument{\DeclareMathVersion{bold}
\SetSymbolFont{operators}{bold}{T1}{times}{b}{n}
\SetSymbolFont{NewLetters}{bold}{T1}{times}{b}{it}
\SetMathAlphabet{\mathrm}{bold}{T1}{times}{b}{n}
\SetMathAlphabet{\mathit}{bold}{T1}{times}{b}{it}
\SetMathAlphabet{\mathbf}{bold}{T1}{times}{b}{n}
\SetMathAlphabet{\mathtt}{bold}{OT1}{pcr}{b}{n}
\SetSymbolFont{symbols}{bold}{OMS}{cmsy}{b}{n}
\renewcommand\boldmath{\@nomath\boldmath\mathversion{bold}}}
\makeatother

\def\BibTeX{{\rm B\kern-.05em{\sc i\kern-.025em b}\kern-.08em
    T\kern-.1667em\lower.7ex\hbox{E}\kern-.125emX}}

\begin{document}

\history{This work has been accepted for publication in IEEE Access.}
\doi{xx.xxxx/ACCESS.2025.xxxxxxx}

\title{Continual Learning with Neuromorphic Computing: Foundations, Methods, and Emerging Applications}
\author{\uppercase{Mishal Fatima Minhas}\authorrefmark{1}\authorrefmark{*}, 
\uppercase{Rachmad Vidya Wicaksana Putra}\authorrefmark{2}\authorrefmark{*}, \IEEEmembership{Member, IEEE}, \uppercase{Falah Awwad}\authorrefmark{1}, \IEEEmembership{Senior Member, IEEE}, \uppercase{Osman Hasan}\authorrefmark{3}, \IEEEmembership{Senior Member, IEEE}, and \uppercase{Muhammad Shafique}\authorrefmark{2}, \IEEEmembership{Senior Member, IEEE}}

\address[1]{Electrical and Communication Engineering Department, United Arab Emirates University (UAEU), Al Ain, United Arab Emirates (e-mail: mishal.fatima@uaeu.ac.ae, f\_awwad@uaeu.ac.ae)}
\address[2]{eBRAIN Lab, New York University (NYU) Abu Dhabi, United Arab Emirates (e-mail: rachmad.putra@nyu.edu, muhammad.shafique@nyu.edu)}
\address[3]{National University of Sciences and Technology (NUST), Pakistan (e-mail: osman.hassan@seecs.edu.pk)}

\tfootnote{* These authors contributed equally to this work}

\corresp{Corresponding author: Falah Awwad (e-mail: f\_awwad@uaeu.ac.ae).}

\begin{abstract}

The challenging deployment of compute- and memory-intensive methods from Deep Neural Network (DNN)-based Continual Learning (CL), underscores the critical need for a paradigm shift towards more efficient approaches. 
\textit{Neuromorphic Continual Learning (NCL)} appears as an emerging solution, by leveraging the principles of Spiking Neural Networks (SNNs) and their inherent advantages (e.g., sparse spike-driven operations and bio-plausible learning rules) for improving energy efficiency and performance, thereby enabling efficient CL algorithms (e.g., unsupervised learning approach) executed in dynamically-changed environments with resource-constrained computing systems. 
Though in its early stages, NCL is already a major research field with an increasing interest in novel SNN-based techniques for different CL methods (e.g., regularization-, replay-, and architecture-based). 
Motivated by the need for a holistic study of NCL, in this survey, we first provide a detailed background on CL, encompassing the desiderata, settings, metrics, scenario taxonomy, Online Continual Learning (OCL) paradigm, recent DNN-based methods proposed in the literature to address catastrophic forgetting (CF). 
Then, we analyze these methods based on their achieved CL desiderata, computational and memory costs, as well as network complexity, hence emphasizing the need for energy-efficient CL. 
After introducing the CL background and the energy efficiency challenges, we provide an extensive background of low-power neuromorphic computing systems including encoding techniques, neuronal dynamics, network architectures, learning rules, neuromorphic hardware processors, software and hardware frameworks, neuromorphic datasets, benchmarks, and evaluation metrics. 
Then, this survey comprehensively reviews and analyzes state-of-the-art works in the NCL field.
The key ideas, implementation frameworks, and performance assessments (including CL, OCL, neuromorphic hardware compatibility aspects) are provided.
This survey also covers several hybrid approaches that combine supervised and unsupervised learning paradigms and categorizes them into three main classes.
It also covers optimization techniques including SNN operations reduction, weight quantization, and knowledge distillation. 
Then, this survey covers the progress of real-world NCL applications categorized into adaptive robots and autonomous vehicles with a wide range of use-cases i.e., object recognition, robotic arm control, cars and road lane detection, Simultaneous Localization and Mapping (SLAM), people detection and robotic navigation and provides their specific case-studies with empirical results.
Finally, this paper provides a future perspective on the open research challenges for NCL, since the purpose of this study is to be useful for the wider neuromorphic AI research community and to inspire future research in bio-plausible OCL.

\end{abstract}


\titlepgskip=-21pt

\maketitle

%%%%%%%%%%%%%%%%%%%%%%%%%%%%%%%%%%%%%%%%%%%%%%%%%%%%%%%%%%%%%%%%%%%%%%%%%%%%%%%%%%%%%%
%%%%%%%%%%%%%%%%%%%%%%%%%%%%%%%%%%%%%%%%%%%%%%%%%%%%%%%%%%%%%%%%%%%%%%%%%%%%%%%%%%%%%%
\section{Introduction}
\label{sec:introduction}

\PARstart{T}{he} inherent cognitive flexibility of the human brain allows continuous learning from environmental interactions throughout life, exhibiting a remarkable ability to simultaneously acquire and retain multiple skills~\cite{bib1}.
For instance, a person fluent in English can also master Spanish and French, each with unique grammatical structures and vocabulary, without losing proficiency in their native language. 
This ability is rooted in the brains' delicate balance between \textit{plasticity} and \textit{stability}~\cite{bib2, bib3}. 
Plasticity allows for the acquisition of new skills, while stability ensures that existing knowledge is preserved. 
Naturally, we expect Artificial Intelligence (AI) systems \cite{russell2016ai, bib5, bib6}, wielding Neural Network (NN) algorithms \cite{bib7}, to develop a similar learning capability to effectively operate and adapt in real-world scenarios. 
While state-of-the-art AI systems excel in single task-based \textit{static environments}, such as image recognition~\cite{bib8}, face identification~\cite{bib9}, and speech recognition~\cite{bib10}, they often struggle in \textit{dynamic environments} where data or tasks may change in structure, distribution, or characteristics over time. 
For instance, when a new task is encountered after a resource-intensive training, NNs are often retrained from scratch despite the high computational costs~\cite{bib44}.  
Therefore, robustness to multiple tasks and sequential experiences, remains a significant research challenge for AI systems. 
When faced with incremental learning of different tasks, most NNs underperform due to suffering from rapid performance degradation, a phenomenon known as \textbf{Catastrophic Forgetting (CF)} or \textbf{interference} \cite{bib15, bib16, bib17}.

\begin{figure} [t]
\centering
\includegraphics[width=0.5\textwidth]{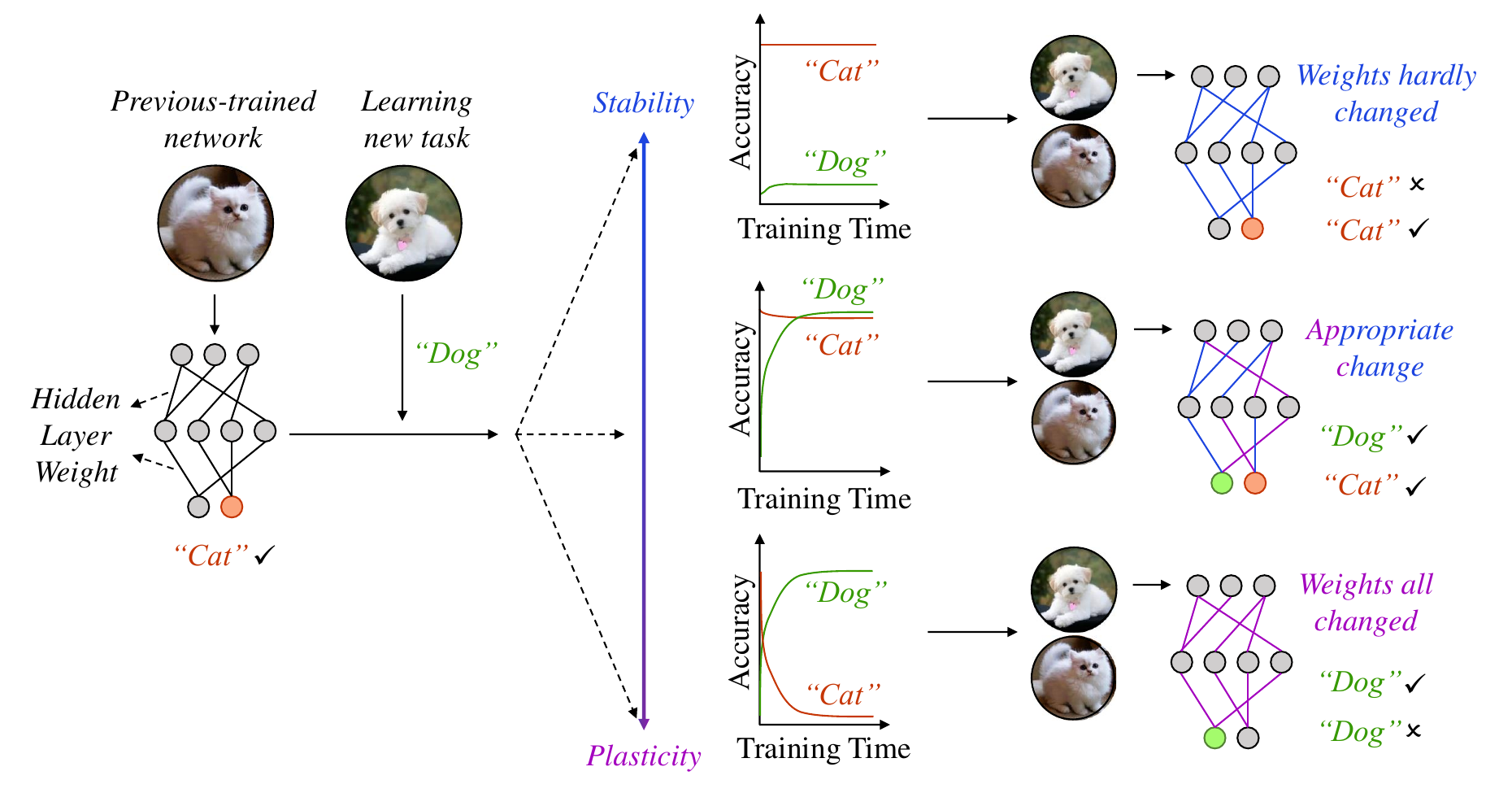} 
\caption{The key challenge of performing sequential task learning in NNs is balancing stability and plasticity of the weights. Failing to maintain this balance results in CF, leading to a significant performance decline on earlier tasks. It is illustrated by the sharp drop in accuracy on the previous ``cat'' task as the network learns a new task, such as recognizing ``dog''; adapted from studies in~\cite{bib23}.} 
\label{sp_dilemma}
\end{figure}

In recent years, \textbf{Continual Learning (CL)} \cite{bib18,bib19,bib20} emerges as a conceptual solution for addressing CF in AI systems. 
CL aims to balance the system's ability to learn new tasks without forgetting the previous knowledge, known as the \textit{stability-plasticity dilemma} \cite{bib21, bib22} (Fig.~\ref{sp_dilemma}). 
It also seeks to achieve generalizability across tasks.
In addition to \textit{avoiding CF}, the primary objectives of CL also include \textit{ensuring scalability of NNs}, \textit{minimizing reliance on old data}, \textit{realizing controlled forgetting}, and \textit{enabling rapid adaptation and recovery}, which will be further discussed in Section~\ref{sec_desiderata}. 

Numerous CL methods have been proposed to address CF in the conventional \textit{Artificial/Deep Neural Network (ANN/DNN)} domain, which showed notable performance improvements~\cite{bib24, bib25, bib26, bib27, bib28, bib29, bib30}. 
However, most methods are resource-intensive, demanding additional memory and computational power. 
Therefore, ANN-based CL methods often fail to account for storage usage during incremental training, necessitating a significant addition of memory for parameters~\cite{bib31}. 
These conditions are not suitable for embedded AI systems, that use portable-battery and need quick adaptation to new data within limited resources.
However, designing resource- and energy-efficient systems with CL capabilities (i.e., \textit{CL systems}) is still a major challenge.

Recently, the brain-inspired \textbf{Neuromorphic Computing (NC)} paradigm \cite{bib47, bib48, bib49, bib50, bib51, Putra_SNNonCNP_arXiv25} has emerged as a promising field for enabling efficient and low-power information processing and decision-making, i.e., by emulating the architecture and functionality of brains with \textbf{Spiking Neural Network (SNN)} algorithms~\cite{bib46, bib53, bib54, bib55, bib56, bib57}. 
The event-driven nature of SNNs supports energy-efficient learning of both static and non-static data streams~\cite{bib46, bib53, bib54, bib55, bib56, bib57}. 
Furthermore, SNNs can perform unsupervised learning due to their bio-plausible learning rules (e.g., \textit{Spike-Timing-Dependent Plasticity (STDP)}~\cite{bib58}), thus enabling continuous adaptation to dynamically changing environments and efficiently learning spatio-temporal data online without labels~\cite{bib296, bib297}. 
In this manner, \textit{NC aligns seamlessly with the desired characteristics (desiderata) of CL}, called \textbf{Neuromorphic Continual Learning (NCL)}. 

\begin{table*}[t]
\centering
\scriptsize
\caption{Qualitative comparison of our survey and the existing ones. Note: ''\checkmark'', ''$\approx$'', and ''$\times$'' mean full, partial, and no consideration, respectively.}
\setlength{\tabcolsep}{1pt}
\begin{tabular*}{\textwidth}{@{\extracolsep\fill} cccccccccccccccc}
\hline%
\rotatebox{45}{\textbf{Surveys}} & \rotatebox{45}{\textbf{Year}} & \rotatebox{45}{\textbf{Pages \#}} & \rotatebox{45}{\textbf{References \#}} & \multicolumn{2}{c}{\rotatebox{45}{\textbf{Scope}}} & \rotatebox{45}{\textbf{Taxonomy}} & \rotatebox{45}{\textbf{HW Platforms}} & \rotatebox{45}{\textbf{SW Frameworks}} & \rotatebox{45}{\textbf{Architectures}} & \rotatebox{45}{\textbf{Benchmarks}} & \rotatebox{45}{\textbf{Datasets}} & \rotatebox{45}{\textbf{Metrics}} & \rotatebox{45}{\textbf{Challenges}} & \rotatebox{45}{\textbf{Quant. Analysis}} & \rotatebox{45}{\textbf{Case Studies}}\\%\cmidrule{4-5}
\cmidrule{5-6}
& &  & & \textbf{CL} & \textbf{NCL} & & & & & & & & \\ 
\hline
\hline
\cite{bib20} & 2019 & $21^*$ & 207 & \checkmark & $\times$ & $\times$ & $\times$ & $\times$ & $\times$ & \checkmark & $\checkmark$ & \checkmark & \checkmark & $\times$ & $\times$ \\
\hline
\cite{bib36} & 2021 & 25 & 344 & \checkmark & $\times$ & \checkmark & $\times$ & $\times$ & $\times$ &  \checkmark & \checkmark & \checkmark & $\times$ & \checkmark & \checkmark \\
\hline
\cite{bib42} & 2021 & 18 & 102 & \checkmark & $\times$ & \checkmark & $\times$ & $\times$ & $\times$ & \checkmark & \checkmark & \checkmark & \checkmark & $\times$ & $\times$\\
\hline
\cite{bib38} & 2022 & 18 & 159 & \checkmark & $\times$ & $\times$ & $\times$ & $\times$ & $\times$ & $\approx$ & \checkmark & \checkmark & $\approx$ & \checkmark & \checkmark  \\
\hline
\cite{bib39} & 2022 & 18 & 124 & \checkmark & $\times$ & $\times$ & $\times$ & \checkmark & \checkmark & $\times$ & \checkmark & \checkmark & $\times$ & \checkmark & \checkmark \\
\hline
\cite{bib40} & 2022 & $32^*$ & 249 & \checkmark & $\times$ & $\times$ &  $\times$ &  $\times$ & $\times$ & \checkmark & $\times$ & $\times$ & $\times$ & $\times$ & $\times$ \\
\hline
\cite{bib41} & 2022 & 7 & 113 & \checkmark & $\times$ & $\approx$ & $\times$ & $\times$ & $\times$ & $\times$ & $\times$ & \checkmark & $\times$ & $\times$ & $\times$ \\
\hline
\cite{bib45} & 2022 & 6 & 78 & $\times$ & SNNs & $\times$ & \checkmark & \checkmark & $\approx$ & $\times$ & $\approx$ & $\times$ & \checkmark &  $\times$ & $\times$\\
\hline
\cite{bib43} & 2022 & 30 & 79 & \checkmark & $\times$ & $\times$ & $\times$ & $\times$ & $\times$ & $\times$ & $\times$ & $\times$ & \checkmark & $\times$ & \checkmark \\
\hline
\cite{wang2024comprehensive} & 2023 & 15 & 281 & \checkmark & $\times$ & \checkmark & $\times$ & $\times$ & $\times$ & $\times$ & $\times$ & $\times$ & \checkmark & $\times$ & $\approx$\\
\hline
\cite{bib46} & 2023 & 11 & 83 & $\times$ & \checkmark & $\times$ & \checkmark & $\times$ & $\approx$ & $\approx$ & $\times$ & \checkmark & $\times$ & \checkmark & \checkmark \\
\hline
\cite{bib59} & 2023 & 18 & 173 & \checkmark & $\times$ & \checkmark & $\times$ & $\times$ & $\times$ & $\times$ & $\times$ & \checkmark & \checkmark & \checkmark & $\times$ \\
\hline
\cite{zhou2024class} & 2024 & 16 & 289 & \checkmark & $\times$ & \checkmark & $\times$ & $\times$ & $\times$ &  \checkmark & \checkmark & \checkmark & $\times$ & \checkmark & \checkmark \\
\hline
\cite{bib44} & 2024 & 13 & 177 & \checkmark & $\times$ & $\times$ & $\times$ & $\times$ & $\times$ & $\times$ & $\times$ & \checkmark & \checkmark & $\times$ & $\times$ \\
\hline
\cite{bib19} & 2024 & 20 & 527 & \checkmark & $\times$ & \checkmark & $\times$ & $\times$ &  $\times$ & $\times$ & $\approx$ & $\times$ & \checkmark & $\times$ & \checkmark \\
\hline
\cite{bib334} & 2024 & $28^*$ & 66 & $\approx$ & \checkmark & $\times$ & $\times$ & $\times$ & \checkmark & $\times$ & $\approx$ & $\times$ & \checkmark & $\times$ & \checkmark \\
\hline
\cite{shi2024continual} & 2024 & $25^*$ & 304 & \checkmark &  $\times$ & \checkmark & $\times$ & $\times$ & $\times$ & \checkmark &  \checkmark & \checkmark & \checkmark & $\times$ & $\times$  \\
\hline
\cite{zhou2024continual} & 2024 & 7 & 80 & \checkmark & $\times$ & \checkmark &  $\times$ &  $\times$ &  $\times$ & \checkmark & \checkmark &  $\approx$ & \checkmark & \checkmark & $\times$  \\
\hline
\cite{li2024continual} & 2024 & 17 & 167 & \checkmark & $\times$ & \checkmark & $\times$ & $\approx$ & $\approx$ & $\approx$ & \checkmark & $\approx$ & $\times$ & \checkmark & $\times$\\
\hline
\cite{yu2024recent} & 2024 & 14 & 144 & \checkmark & $\times$ & \checkmark & $\times$ &  $\times$ & \checkmark & \checkmark & \checkmark & \checkmark & \checkmark &  $\times$ &  $\times$\\
\hline
\cite{yang2025recent} & 2025 & 27 & 233 & \checkmark & $\times$ & $\times$ & $\times$ & $\times$ & $\times$ & \checkmark & \checkmark & \checkmark & \checkmark & $\times$ & $\times$ \\
\hline
\textbf{This work} & \textbf{2025} & 39 & 413 & \textbf{\checkmark} & \textbf{\checkmark} & \textbf{\checkmark} & \textbf{\checkmark} & \textbf{\checkmark} & \textbf{\checkmark} & \textbf{\checkmark} & \textbf{\checkmark} & \textbf{\checkmark} & \textbf{\checkmark} & \textbf{\checkmark} & \textbf{\checkmark}\\
\hline
\hline
\end{tabular*}
\begin{tablenotes} \footnotesize
\item \textit{*} Single-column pages.
\end{tablenotes}
\label{table_1} 
\vspace{-5pt}
\end{table*}

%%%%%%%%%%%%%%%%%%%%%%%%%%%%%%%%%
\subsection{Surveys on CL and NCL}

\textit{To fully understand the foundations and recent advancements in CL and NCL, a comprehensive survey is imperative}, unlike other surveys that focus on a single area (either CL, NCL, bio-plausible learning rules, or applications). Table~\ref{table_1} compares surveys studying multiple aspects of CL, NCL or SNNs, like our work. 
Most of the existing CL surveys~\cite{bib19, bib20, bib36, wang2024comprehensive, bib38, zhou2024class, bib39, bib40, bib41, bib42, bib43, bib44, bib45} have particularly focused on addressing CF in DNNs. 
For instance, studies in \cite{bib19} provided an up-to-date CL survey in DNNs. 
Meanwhile, other surveys focused on categorizing CL methods~\cite{zhou2024class, bib39, bib36, bib41}, studying the challenges of forgetting in DNNs~\cite{bib44, wang2024comprehensive, bib42}, reviewing computational requirements of CL (e.g., regularization)~\cite{bib44}, discussing the benefits of forgetting in CL~\cite{wang2024comprehensive}, providing experimental comparisons of 11 CL methods~\cite{bib42}, discussing \textit{Online Continual Learning (OCL)} paradigm~\cite{bib43}, and discussing the importance of adopting brain-inspired data representations~\cite{bib59}. 
A more recent line of surveys have also explored CL in large pre-trained models, focusing on paradigms such as prompt tuning, adapters, and distillation for mitigating forgetting in large language models (LLMs)~\cite{shi2024continual, zhou2024continual, li2024continual, yu2024recent, yang2025recent}. While these methods offer solutions for foundation models, \textit{they are fundamentally different from resource-constrained, biologically-inspired approaches like NCL for enabling energy-efficient CL and OCL, which remain underexplored in existing surveys}. In particular, \textit{our survey differs from the study} \cite{bib19} in the following ways (see overview in Table~\ref{table_1} and an in-depth comparison in Table~\ref{table_sc}).

\begin{itemize}
    \item \textbf{Focus:}
    Survey~\cite{bib19} focused on categorizing DNN-based CL methods, and provided theoretical concepts of stability-plasticity trade-offs, generalizability in CL, and application domains. 
    Our survey builds on that foundation to highlight the limitations of DNN-based CL methods, i.e., computational and memory overhead for resource-constrained systems. Motivated by these challenges, it provides a specialized perspective on state-of-the-art NCL methods, categorizing the existing literature into: 1) enhancements on unsupervised STDP learning, 2) predictive coding, 3) active dendrites, 4) bayesian continual learning, 5) architecture, 6) replay, 7) regularization and 8) hebbian learning based methods, along with optimization techniques proposed by these works, and SNN-specific application use-cases for enabling energy-efficient CL and OCL in embedded AI systems.
    \item \textbf{Comparative Quantitative Analysis:} 
    Our survey distinguishes itself from \cite{bib19} by providing a comparative quantitative analysis considering design factors (i.e., network complexity) and evaluation metrics (i.e., accuracy, memory footprint, latency, power/energy usage) of the reviewed NCL methods with relevant DNN-based CL methods. It also provides a detailed quantitative trade-off analysis (accuracy vs. efficiency) for various NCL methods. Additionally, it provides a comparison of prominent neuromorphic datasets and a comparative analysis of NCL frameworks with standard DNN-based CL frameworks, which are not provided in \cite{bib19}.
    \item \textbf{Application Domains:} 
    Our survey systematically analyzes the current progress of NCL in real-world applications by categorizing existing works into: 1) adaptive robots, and 2) autonomous vehicles/agents. It provides specific case studies with empirical results (latency, power/energy, accuracy), and OCL capability assessment, to demonstrate the real-world feasibility of NCL in embedded AI systems. Furthermore, it reports the datasets, learning rules, software/hardware platforms, and discusses hardware deployment challenges.
    These aspects are not covered by~\cite{bib19} survey, as it mainly focuses on the scenario complexity and task-specific challenges.
\end{itemize}

\begin{table}[t]
\centering
\scriptsize

\caption{Comparison of our survey with study \cite{bib19}. Note: ''\checkmark'', ''$\approx$'', and ''$\times$'' mean full, partial, and no consideration, respectively.}
\begin{tabular}{|p{3.8cm}|p{2.1cm}|p{1.2cm}|}
\hline
\textbf{Aspect} & \textbf{This Work} & \cite{bib19} \\
\hline
\hline
Taxonomy \& categorization & NCL methods \& their real-world use-cases & CL methods \\
\hline
Neuromorphic datasets \& SNN-based evaluation metrics & \checkmark & $\times$ \\
\hline
SNN-based CL, OCL capability, \& hardware compatibility & \checkmark & $\times$ \\
\hline
Hardware deployment constraints & \checkmark & $\approx$ \\
\hline
Quantitative trade-off analysis & \checkmark & $\times$ \\
\hline
Comparative analysis of NCL frameworks with
standard DNN-based CL frameworks & \checkmark & $\times$ \\
\hline
Comparative quantitative analysis of NCL methods with DNN-based CL methods & \checkmark & $\times$ \\
\hline
\end{tabular}
\label{table_sc}
\end{table}

The remarkable CL capability of humans has sparked interest in the emerging of NCL paradigm, i.e., investigation on how SNN architectures and learning rules can be utilized to develop efficient CL systems.
For instance, studies in~\cite{bib46} explored biological mechanisms for NCL and emphasized the importance of CL, but did not provide a comparative quantitative analysis of the reviewed NCL methods with relevant DNN-based CL methods. It also did not provide details of neuromorphic datasets or a taxonomy that categorizes the recent literature on NCL methods and SNN-specific real-world application use-cases. 
The other survey~\cite{bib334} provided the background theories on sparse and predictive coding and their relation to Hebbian and bio-plausible learning, hence it only focused on learning rules and application aspect of NCL. It did not cover the aspects such as OCL capability, software and hardware frameworks, hardware compatibility, neuromorphic datasets \& benchmarks, performance and computational efficiency trade-offs of existing NCL works. 
Therefore, \textit{a unified survey that covers the foundations of CL, recent advances in both the DNN and SNN domains, provides a comparative and application-driven view of NCL with detailed NCL design and optimization aspects, assessing suitability for OCL and neuromorphic hardware compatibility, has not been provided.}

\begin{figure*}[t]
\centering
\includegraphics[width=\textwidth]{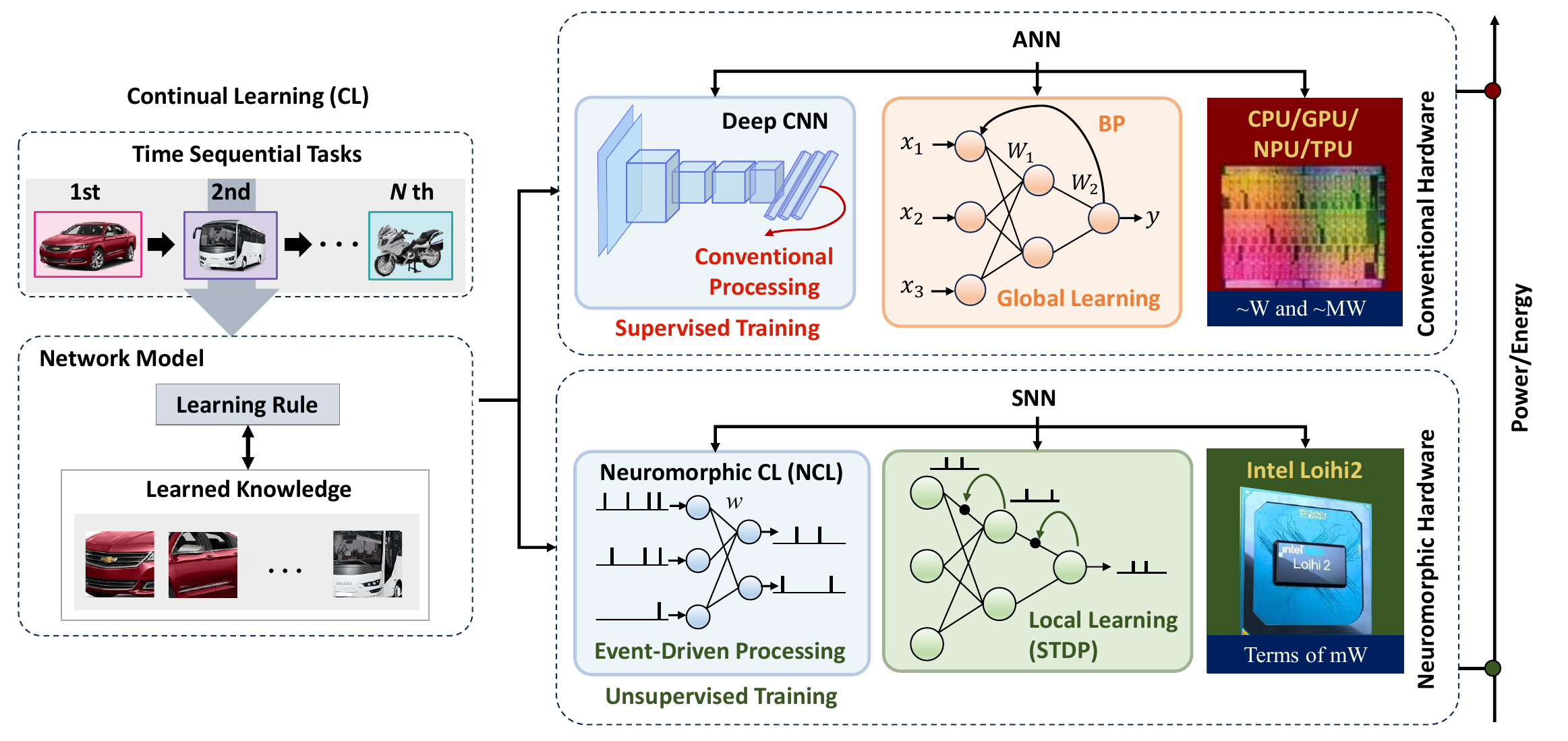}
\caption{Overview of our survey: CL in ANNs/DNNs and SNNs (i.e., NCL). ANNs/DNNs rely on global learning, supervised training, and computationally intensive processing on conventional hardware, thus making them less suitable for OCL. SNNs leverage event-based processing, local learning rules, and low-power neuromorphic hardware which enable energy-efficient CL, thus making them suitable for OCL.}
\label{survey_focus}
\end{figure*}

%%%%%%%%%%%%%%%%%%%%%%%%%%%%%%%%%%%%%
\subsection{Scope and Contribution}
\label{contribution}

\textbf{Our survey addresses the following key questions}.
\begin{itemize}
    \item [Q1:] What are the CL desiderata, settings, evaluation metrics, and taxonomy of scenarios?
    \item [Q2:] What are the different categories of existing CL methods in the ANN/DNN domain, how they overcome CF, and their achieved desiderata?
    \item [Q3:] What are the deployment challenges of these CL methods (i.e., computational/memory overhead) that necessitate energy-efficient approaches? 
    \item [Q4:] What is the technical background of low-power neuromorphic systems, current software and hardware frameworks/platforms, datasets, benchmarks and evaluation metrics? 
    \item [Q5:] How do neuromorphic frameworks compare with standard deep learning (DL) and CL frameworks?
    \item [Q6:] What methods do the state-of-the-art propose to enable energy-efficient CL with SNNs (i.e., NCL), their OCL capability, and which efficiency enhancement techniques they propose?
    \item [Q7:] What is the compatibility of reviewed methods with neuromorphic hardware platforms, and their on-chip learning capabilities?
    \item [Q8:] Is there a standardized evaluation framework or metrics for benchmarking NCL and what additional metrics can be used for evaluating NCL methods?
    \item [Q9:] What are the hardware deployment challenges of NCL methods considering accuracy-efficiency trade-offs?
    \item [Q10:] What are the specific use-cases in adaptive robots and autonomous vehicles that benefit from NCL, their OCL capability, and quantitative performance analysis?
    \item [Q11:] What are the identified open challenges and proposed future research directions?
\end{itemize}

\smallskip
By addressing the above key questions, \textbf{our paper covers all the key aspects discussed above and makes the following contributions} (see overview in Fig.~\ref{survey_focus}): 

\begin{enumerate}
    \item We provide the preliminaries of CL including formulation, desiderata, settings, taxonomy of learning scenarios, evaluation metrics, and OCL paradigm. We discuss how different categories of existing CL methods which are pre-dominantly from ANN/DNN domain overcome CF, and which of the desiderata they achieve.
    Furthermore, we emphasize their challenges (computational/memory overhead) for practical applications in resource-constrained systems, outlining the need for energy-efficient approaches.  
    \item We discuss the technical background of low-power neuromorphic computing systems covering the key aspects (i.e., encoding schemes, neuronal dynamics, learning rules, software/hardware frameworks, neuromorphic processor architectures, an elaborate description of neuromorphic datasets, benchmarks \& metrics), suggesting the need for standardized NCL benchmarks and additional metrics.
    We provide a comparative analysis of NCL frameworks with conventional DNN-based CL frameworks.
    Then, we provide a comprehensive review of state-of-the-art works for NCL, categorizing them into: 1) enhancements on unsupervised STDP learning, 2) predictive coding, 3) active dendrites, 4) bayesian continual learning, 5) architecture, 6) replay, 7) regularization and 8) hebbian learning based methods, assessing their CL performance, OCL capabilities, and neuromorphic hardware compatibility. We discuss hybrid approaches that combine supervised and unsupervised learning paradigms to address CF and improve CL performance. We categorize these approaches into three main classes: 1) self-supervised pre-training hybrids, 2) STDP + supervised learning hybrids, and 3) generative-discriminative hybrid models. We discuss efficiency enhancement techniques for NCL proposed by state-of-the-art including SNN operations reduction, quantization, and knowledge distillation. Moreover, we discuss the performance and computational efficiency trade-offs of NCL considering hardware implementation constraints and provide a comparative quantitative analysis considering design factors and evaluation metrics of the reviewed NCL with relevant DNN-based CL methods.
    \item We categorize the current progress of NCL in real-world applications into: 1) adaptive robots, and 2) autonomous vehicles/agents,  studying a range of emerging use-cases including object recognition, robotic arm control, cars and road lane detection, Simultaneous Localization and Mapping (SLAM), people detection and robotic navigation, and report their specific case-studies with empirical results, implementation frameworks, and OCL capabilities to assess real-world feasibility. 
    Toward the end, we discuss the identified open research challenges and provide a perspective on future research directions.     
\end{enumerate}

Table~\ref{tab:abbreviations} summarizes the abbreviations used throughout the paper for clarity and consistency.

\begin{table}
\centering
\caption{List of Abbreviations}
\begin{tabular}{|l|l|}
\hline
\textbf{Abbreviation} & \textbf{Full Form} \\
\hline
\hline
CL & Continual Learning \\
DL & Deep Learning \\
ANN & Artificial Neural Network \\
DNN & Deep Neural Network \\
CNN & Convolutional Neural Network\\
NNs & Neural Networks \\
CF & Catastrophic Forgetting \\
NCL & Neuromorphic Continual Learning \\
SNN & Spiking Neural Network \\
CSNN & Convolutional SNN \\
OCL & Online Continual Learning \\
TA & Task Agnostic \\
FWT & Positive Forward Transfer\\
BWT & Positive Backward Transfer\\
RL & Reinforcement Learning \\
STDP & Spike-Timing-Dependent Plasticity \\ 
SLAM & Simultaneous Localization and Mapping \\
Task-IL & Task-Incremental Learning\\
Domain-IL & Domain-Incremental Learning \\
Class-IL & Class-Incremental Learning \\
DTA & Discrete Task Agnostic \\
CTA & Continuous Task Agnostic \\
OWTA & Open-World Task Agnostic \\
SSL & Self-Supervised Learning \\
CPT & Continual Pre-Training \\
BP & Backpropagation \\
EWC & Elastic Weight Consolidation \\
FIM & Fisher Information Matrix \\
SI & Synaptic Intelligence\\
GEM & Gradient Episodic Memory \\
GPM &  Gradient Projection Memory \\
MAS & Memory Aware Synapses \\
RWalk & Riemannian Walk \\
MLP & Multi-layer Perceptron\\
FC & Fully-Connected \\
CFN & Controlled Forgetting Network \\
ASP & Adaptive Synaptic Plasticity\\
ViT & Vision Transformer \\
KD & Knowledge Distillation \\
ER & Experience Replay \\
HW & Hardware \\
SW & Software \\
TTFS & Time-to-first Spike \\
HH & Hodgkin–Huxley \\
IF & Integrate-and-Fire \\
LIF & Leaky IF \\
AdExIF & Adaptive Exponential IF \\
SDSP & Spike-Driven Synaptic Plasticity \\
R-STDP & Reward-Modulated STDP \\
PES & Prescribed Error Sensitivity \\
BPTT & Backpropagation Through Time \\ 
SG & Surrogate Gradient\\
STBP & Spatio-Temporal Backpropagation \\ 
DECOLLE & Deep Continuous Local Learning \\
ST-LRA & Spike-Triggered Local Representation Alignment \\
BrainCog & Brain-inspired Cognitive Intelligence Engine \\
CPU & Central Processing Unit\\
GPU & Graphic Processing Unit\\
MCU & Microcontroller Unit\\
TPU & Tensor Processing Unit \\
CMOS & Complementary Metal-Oxide-Semiconductor\\
FPGA & Field-Programmable Gate Array\\
ASIC & Application-Specific Integrated Circuit \\
ROLLS & Reconfigurable On-line Learning Spiking \\
& Neuromorphic Processor \\
PIM & Processing-In-Memory\\
CIM & Compute-In-Memory\\
NVM & Non-Volatile Memory\\
RRAM & Resistive Random Access Memory\\
MRAM & Magnetic RAM\\
PCM & Phase Change Memory\\
\hline
\end{tabular}
\label{tab:abbreviations}
\end{table}

\addtocounter{table}{-1}
\begin{table}

\caption{Continued}
\begin{tabular}{|l|l|}
\hline
\textbf{Abbreviation} & \textbf{Full Form} \\
\hline
\hline
DVS & Dynamic Vision Sensor \\
ATIS & Asynchronous Time-based Image Sensor \\
DAS & Dynamic Audio Sensor \\
NVS & Neuromorphic Vision Sensor \\
BAE &  Biologically Plausible Auditory Encoding \\
mAP & Mean Average Precision \\
MAE & Mean Absolute Error \\
$MAE_L$ &  MAE for Localisation \\
$MAE_M$ &  MAE for Mapping \\
MSE & Mean Square Error \\
RMSE & Root Mean Square Error \\
AOC & Average of top-1 accuracy over classes\\
ITAE & Integral of Time-weighted Absolute Error \\
SNR & Signal-to-Noise Ratio \\
SpNCN & Spiking Neural Coding Network \\
DSD-SNN & Dynamic Structure Development of SNN \\
SOR-SNN & Self-Organized Regulation SNN \\
LR & Latent Representation/Replay \\
HLOP & Hebbian Learning-based Orthogonal Projection \\
NEF & Neural Engineering Framework\\
NSM & Neural State Machine\\
IK & Inverse Kinematics \\
PID & Proportional Integral Derivative \\
PES & Prescribed Error Sensitivity \\
WCE & Weighted Binary Cross Entropy \\
IoU & Intersection over Union\\
FMCW & Frequency Modulated Continuous Wave \\
HNN & Hybrid Neural Network (SNN-ANN) \\
FPS/W & Frames Per Second per Watt \\
DoF & Degrees of Freedom \\
FID & Fréchet distance\\
\hline
\end{tabular}
\label{tab:abbreviations}
\end{table}

\begin{figure*}[t]
\centering
\includegraphics[width=0.75\textwidth]{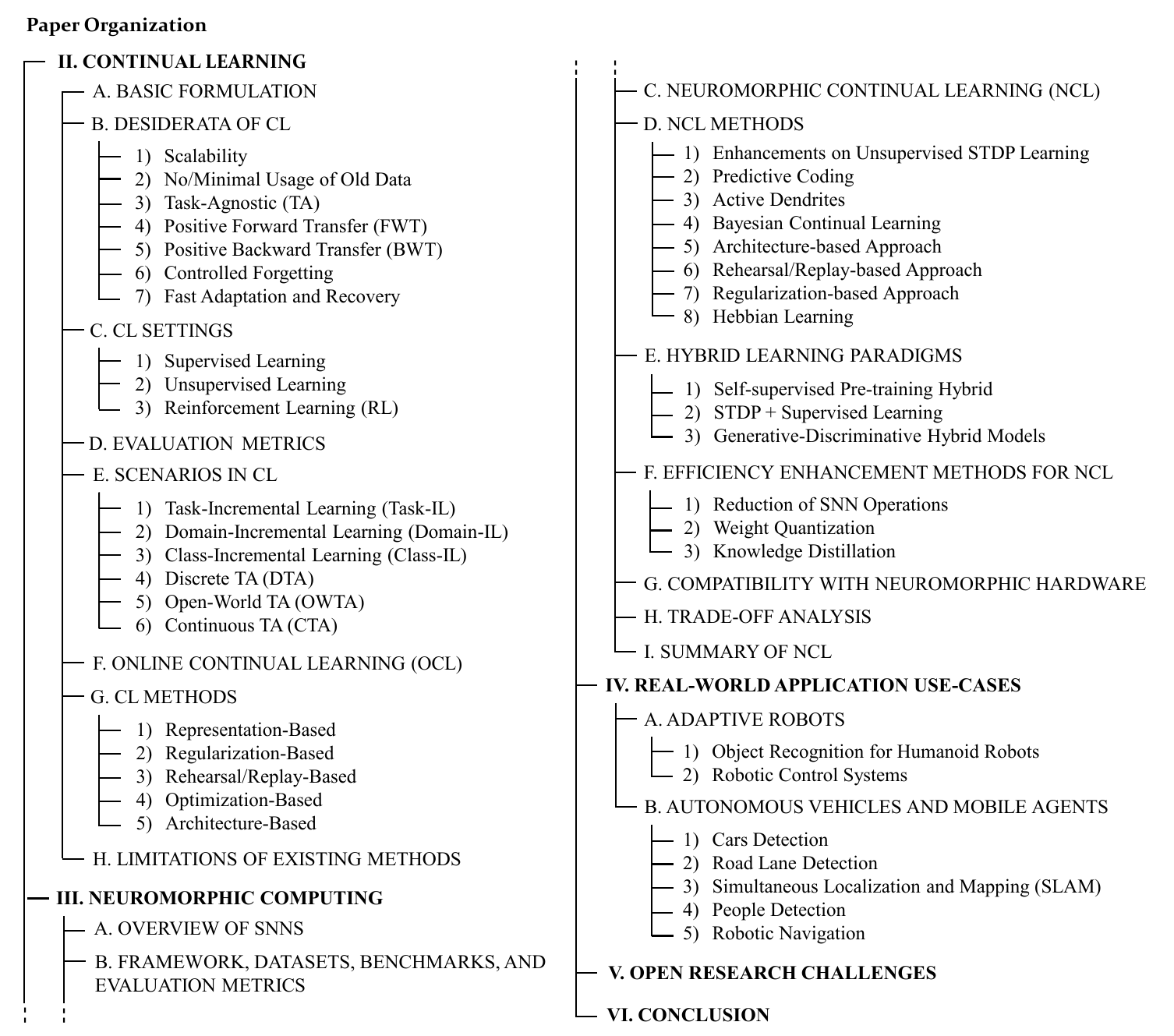} 
\caption{Organization of our survey paper} 
\label{paper_org}
\end{figure*}

%%%%%%%%%%%%%%%%%%%%%%%%%%%%%%%%%%%
\subsection{Paper Structure}

The rest of this survey paper is organized as shown in Fig.~\ref{paper_org}. 
Section~\ref{sec: CL} delves into the CL background. 
Section~\ref{sec3} provides detailed technical background of neuromorphic computing systems across multiple aspects and comprehensively reviews 
state-of-the-art NCL methods, with comparative quantitative and trade-off analysis.  
In Section~\ref{sec_usecases}, several application use-cases of NCL are provided and analyzed. 
Then, Section~\ref{sec_openchallenges} discusses open challenges for meeting the CL desiderata with SNNs and proposes future research directions for each challenge. 
Finally, Section~\ref{sec_conclusion} presents the conclusion. 

%%%%%%%%%%%%%%%%%%%%%%%%%%%%%%%%%%%%%%%%%%%%%%%%%%%%%%%%%%%%%%%%%%%%%%%%%%%%%%%%%%%%%%
%%%%%%%%%%%%%%%%%%%%%%%%%%%%%%%%%%%%%%%%%%%%%%%%%%%%%%%%%%%%%%%%%%%%%%%%%%%%%%%%%%%%%%
\section{Continual Learning}
\label{sec: CL}

In recent years, ANNs have revolutionized the field of AI~\cite{bib8, bib43, bib112}, due to their effective training technique (i.e., gradient descent-based backpropagation), while assuming the training data are independently and identically distributed (i.i.d)~\cite{bib113,bib114}. 
However, a critical question arises: \textit{how can we integrate new data into existing models}? 
One approach is to retrain with new data, while using current parameters as the initial state~\cite{bib114}. 
However, successive training tends to cause NNs to suffer from CF~\cite{bib115, bib116}. 
Although retraining from scratch effectively tackles CF issues, but it is inefficient and almost impossible in some cases, such as real-time learning scenarios where the model must update and adapt to continuous data streams instantly with limited resources (e.g., developmental learning for autonomous agents~\cite{bib117}). 

CL (also known as \textit{incremental learning} or \textit{lifelong learning}) addresses CF by accumulating knowledge from a continuous data stream throughout their lifetime~\cite{bib1}. 
As stability-plasticity dilemma requires trade-off between preserving past knowledge (\textit{stability}) and adapting to new experiences (\textit{plasticity})~\cite{bib121}, researchers have developed several CL methods to provide the same capability~\cite{bib122, bib123, bib124}. 
While many CL methods focused on the offline learning with supervised task-based incremental learning, which operates under the assumption of i.i.d. data and constant task identification, these assumptions often diverge from real-world scenarios. 
In practical applications, input data streams are typically not i.i.d. and task identity may not be available. 
It highlights the complexity of CL problems, showing the importance of understanding the CL foundations.  
Therefore, in this section, we delve into the basic formulation and desiderata of CL, and explore various learning scenarios, methods, settings, and evaluation metrics. 

%%%%%%%%%%%%%%%%%%%%%%%%%%%%%%%%
\subsection{Basic Formulation} 

\begin{figure*}[h]
\centering
\includegraphics[width=\textwidth]{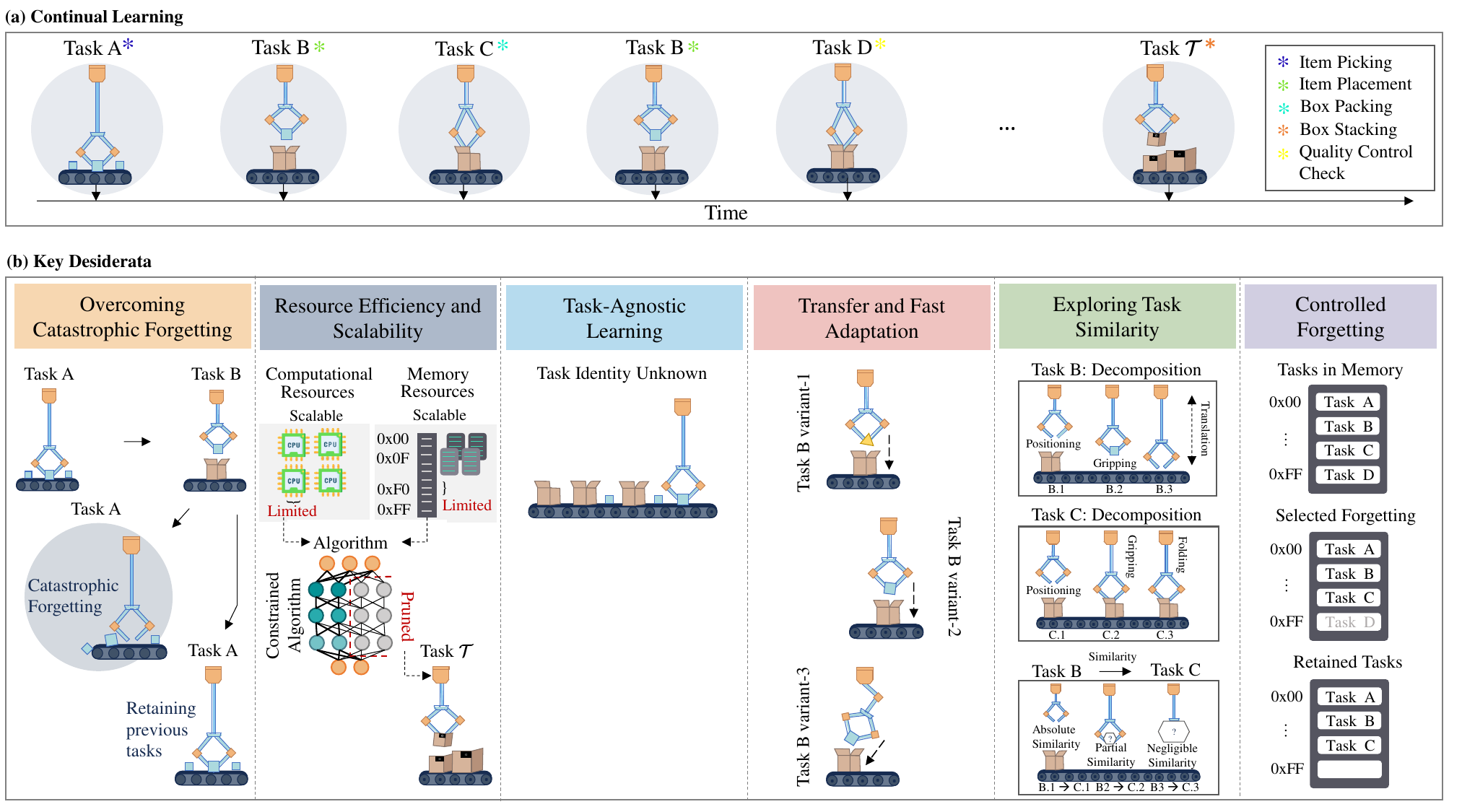} 
\caption{(a) A robotic arm being trained to perform a variety of tasks sequentially, and is subsequently able to select from its repertoire of learned skills to apply in different situations. (b) Key desiderata for CL; adapted from studies in~\cite{bib1}.} 
\label{CL_desiderata}
\end{figure*}

A CL algorithm must adapt to new tasks without full access to previous training data while maintaining performance on respective test sets. 
Formally, defined by parameters $\theta = \cup_{t=1}^{\mathcal{T}} \theta^{(t)} $, where $\theta^{(t)} = \{e^{(t)}, \psi\}$, $e^{(t)}$ is the task-specific parameters and $\psi$ is the task-sharing parameters for a task \textit{t}. 
A batch of training samples for a task \textit{t} is denoted as $\mathcal{D}_{t,b} = $\{$\mathcal{X}_{t,b}$, $\mathcal{Y}_{t,b}$\} where $\mathcal{X}_{t,b}$ represents input data, $\mathcal{Y}_{t,b}$ denotes data labels, and \textit{t} $\in$ $\mathcal{T}$ = \{1,···,$\mathcal{T}$\} signifies the task identity, with \textit{b} $\in$ $\mathcal{B}_t$ as the batch index ($\mathcal{T}$ and $\mathcal{B}_t$ indicate their respective spaces). 
Task \textit{t} is defined by its training samples $\mathcal{D}_t$, with distribution $\mathbb{D}_t$ $:=$ \textit{p}($\mathcal{X}_t,\mathcal{Y}_t$), $\mathcal{D}_t$ encompasses the entire training set, omitting the batch index and assuming no distribution discrepancy between training and test sets. 
However, in practical scenarios, data labels $\mathcal{Y}_t$ and task identity (Task ID) \textit{t} may not always be available. 
CL accommodates varying batch sizes for each task’s training samples (i.e., \{\{$\mathcal{D}_{t,b}\}_{b \in \mathcal{B}_t}\}_{t \in \mathcal{T}}$) or simultaneous arrival (i.e.,  \{$\mathcal{D}_t\}_{t \in \mathcal{T}}$) \cite{bib19}.   

%%%%%%%%%%%%%%%%%%%%%%%%%%%%%%%%%%
\subsection{Desiderata of CL}
\label{sec_desiderata}
 
In this sub-section, we address the key question Q1 by explaining the desired characteristics (desiderata) of a CL algorithm, and show an overview of key requirements to achieve CL capability using a robotic arm as an example in Fig.~\ref{CL_desiderata}. 

%%%%% -----
\subsubsection{Scalability} \label{2.2.1}

A CL algorithm should effectively train on a large or potentially unlimited number of tasks without expanding NNs excessively. 
The increasing of NN capacity and computational costs have to be minimal or sub-linear, thereby ensuring the \textit{scalability} of a CL algorithm~\cite{bib59}. 
To achieve high scalability, the major challenges include \textit{scalability of network model size}~\cite{bib19,bib60} and \textit{scalability of regularization terms} (e.g., weight regularization~\cite{bib61}). 

%%%%% -----
\subsubsection{No/Minimal Usage of Old Data} 

Minimizing or eliminating the reliance on data from previous tasks is crucial for a CL algorithm, because of storage constraints and privacy concerns. 
Recent research has made progress to achieve this goal~\cite{bib65, bib66, bib67, bib68, bib69, bib70}. 
For instance, previous studies used mutual information maximization and compressed gradients to minimize the reliance on old data~\cite{bib65, bib66, bib67}, and employed representative subsets to preserve knowledge~\cite{bib68}.

%%%%% -----
\subsubsection{Task Agnostic (TA)}

A CL algorithm should function independently from pre-defined task boundaries during the training and inference phases, known as \textit{Task Agnostic} (TA). 
For instance, a robot in an ever-changing environment should adapt to new tasks, such as picking up different objects or navigating new terrains without explicit information of task identity. 
It has to learn from environment and experiences, and dynamically adjust to new tasks. 
State-of-the-art attempted to achieve TA using methods studied in~\cite{bib82, bib88, bib84, bib85, bib81, bib77, bib87, bib76}.

%%%%% -----
\subsubsection{Positive Forward Transfer (FWT)}

A CL algorithm should leverage previous knowledge to enhance performance on new tasks, known as \textit{Positive Forward Transfer (FWT)}. 
Formally, $\text{FWT}_k$ evaluates the average influence of all old tasks on the current $k$-th task, and can be stated as:
\begin{equation} \label{eq1} 
\text{FWT}_k = \frac{1}{k-1} \sum_{j=2}^{k} (a_{j-1,j} - \tilde{a}_j)
\end{equation} 
where, $\tilde{a}_j$ is the accuracy of a randomly-initialized model trained with data $\mathcal{D}_{j}$ for the $j$-th task~\cite{bib19}. The parameter $a_{j-1,j}$ denotes accuracy on task $j$ after learning task $j-1$, but before task $j$ and $\tilde{a}_j$ denotes baseline performance on task $j$ without prior learning. High FWT indicates that the model generalizes well to unseen tasks by reusing learned knowledge.
Recent research explored different methods to enable FWT, including studies in~\cite{bib92, bib101, bib102, bib99, bib100, bib103, bib106, bib59, bib104, bib105, bib91}.

%%%%% -----
\subsubsection{Positive Backward Transfer (BWT)}

A CL algorithm should transfer knowledge from later tasks to past tasks for improving performance on the past tasks, known as \textit{Positive Backward Transfer (BWT)}. 
Negative BWT reflects CF and zero BWT suggests zero forgetting~\cite{bib59}. 
Formally, $\text{BWT}_k$ evaluates the average influence of learning the $k$-th task on all old tasks, and can be stated as:
\begin{equation} \label{eq2}
\text{BWT}_k = \frac{1}{k-1} \sum_{j=1}^{k-1} (a_{k,j} - a_{j,j}) 
\end{equation}
where, $a_{k,j}$ is the accuracy of a model at $j$-th task after incremental learning of the $k$-th task~\cite{bib19}. 
The parameter $a_{j,j}$ denotes accuracy of task $j$ when it was first learned.

%%%%% -----
\subsubsection{Controlled Forgetting}

A CL algorithm should forget old yet insignificant knowledge to make room for learning new information, known as \textit{controlled forgetting}. 
Although humans do not experience sudden memory loss, a gradual decrease in memory (forgetting) over time is natural, hence controlled forgetting is beneficial for managing memory retention~\cite{bib71}.
Related methods include~\cite{bib109, bib107, bib62} for ANNs and~\cite{bib72, bib108, bib296, bib297} for SNNs.

%%%%% -----
\subsubsection{Fast Adaptation and Recovery}

A CL algorithm should quickly learn new tasks while minimizing the loss of previous knowledge (\textit{fast adaptation}), and regain previous performance after encountering new tasks that cause degradation in learned knowledge (\textit{recovery}). 
Related methods include regularization~\cite{bib61} and \textit{gradient-based meta-learning} or \textit{fast optimization}~\cite{bib110}. 

%%%%%%%%%%%%%%%%%%%%%%%%%%%%%%%
\subsection{CL Settings} 
\label{CLsettings}

In this sub-section, we address the key question Q1 by introducing the definition of learning settings in which the CL methods operate, based on how the data label and information are utilized for learning process, including the \textit{supervised}, \textit{unsupervised}, and \textit{reinforcement learning} settings.

%%%%----
\subsubsection{Supervised Learning}  

\textit{It is an ML setting that trains models with labeled datasets}.  
Therefore, in the \textbf{Supervised CL} setting, the goal is to train a model on a sequential data stream. 
Its challenge is to efficiently perform the training, since the supervised-based learning typically requires huge memory and energy requirements. 
Moreover, obtaining a large labeled dataset can be time consuming and expensive. 

%%%%----
\subsubsection{Unsupervised Learning}

\textit{It is an ML setting that trains models with unlabeled data (e.g., identifying patterns, structures, or relationships)}~\cite{bib133}. 
Therefore, in the \textbf{Unsupervised CL} setting, the goal is to develop a model that can discover similarity in samples that arrive sequentially~\cite{bib133}. 
Its challenge is to extract meaningful features without explicit labels. 
This setting is particularly useful for tasks where data labeling is impractical or expensive.

%%%%----
\subsubsection{Reinforcement Learning (RL)}

\textit{It is an ML setting that trains models to make sequential decisions based on the feedback (i.e., rewards or penalties)}~\cite{bib250,bib255}.  
Through trials and errors, the model learns to associate actions with outcomes, aiming to discover the optimal strategy or policy to achieve its objectives over time.
Its challenge is to efficiently perform the training, as RL typically requires high memory and energy requirements. 
Moreover, the nature of trials and errors can also be time-consuming and expensive. 

%%%%%%%%%%%%%%%%%%%%%%%%%%%%%%%%%%
\subsection{Evaluation Metrics}  
\label{eval_metrics}

To evaluate the performance of an efficient CL system, metrics are essential. 
This sub-section addresses the key question Q1 for evaluation metrics.
To assess the overall performance of a CL algorithm, \textit{average accuracy (AA)}~\cite{bib134,bib100}, \textit{average incremental accuracy (AIA)}~\cite{bib135,bib136} and \textit{forgetting (F)} are often used, where F is the average maximum drop in classification accuracy over time, with lower values indicating better retention. \textit{BWT} and \textit{FWT} are also used to quantify memory stability and learning plasticity, respectively, in CL.
\textit{Model size} is used to measure the memory size for storing the model parameters~\cite{bib137}, while \textit{sample storage size (SSS)} is used to measure the storage size for storing data samples.
To evaluate the computational resources required for the training and inference phases, metrics such as the number of \textit{floating-point operations-per-second (FLOPS)} and \textit{time complexity} are often used. 
Meanwhile, \textit{power/energy efficiency} measures the efficiency gains of a CL system during its operational life-time.

%%%%%%%%%%%%%%%%%%%%%%%%%%%%%%%%%%%%%%%%%
\subsection{Scenarios in CL} 
\label{CLscenarios}

\begin{figure} [h]
\centering
\includegraphics[width=0.5\textwidth]{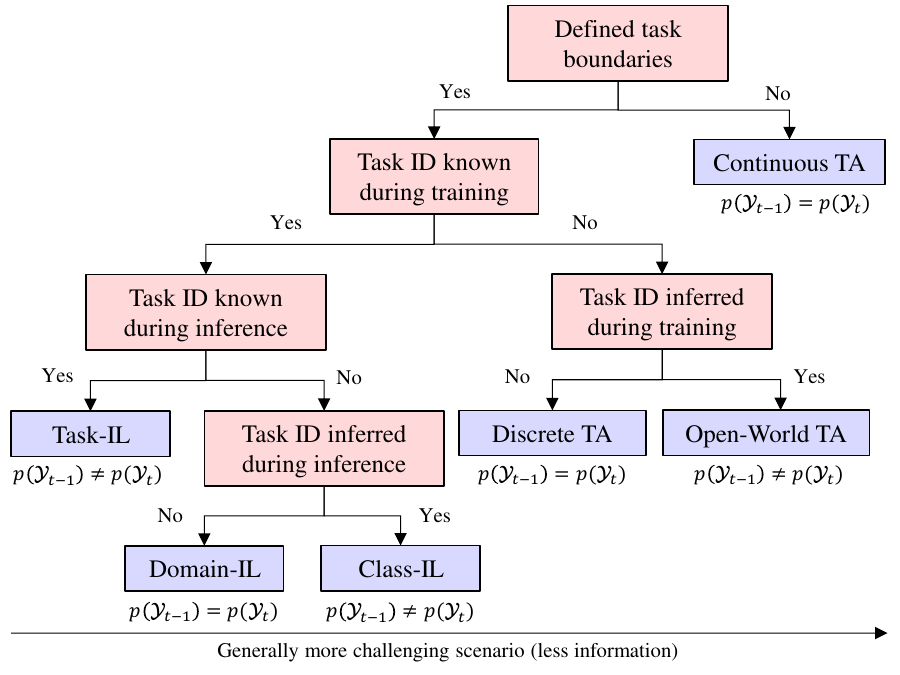} 
\caption{Taxonomy of the CL scenarios based on the nature of training and inference setups considering the task identity (Task ID); adapted from studies in~\cite{bib59}.} 
\label{CL_scenarios}
\end{figure}

\begin{figure*} [h]
\centering
\includegraphics[width=\textwidth]{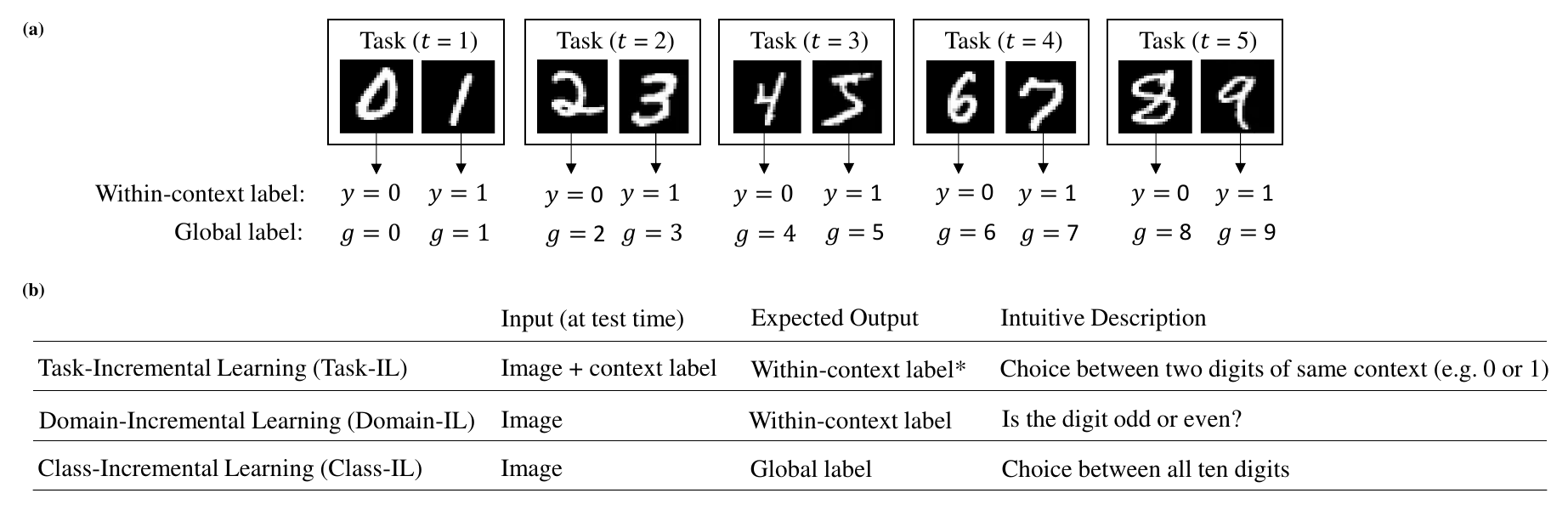} 
\caption{Split MNIST according to Task-IL, Domain-IL, and Class-ID. 
(a) Split MNIST is obtained by splitting the original MNIST into five contexts/tasks, each having two classes. 
(b) Overview of what is expected of the model at test time for each scenario.} 
\label{MNIST_scenarios}
\end{figure*}

In this sub-section, we address key question Q1 regarding the CL scenarios by presenting their definitions, taxonomy (shown in Fig.~\ref{CL_scenarios}), and formal comparison in Table~\ref{CL_Scenarios}.
The initially-identified scenarios are
\textit{Task-Incremental Learning (Task-IL)}, \textit{Domain-Incremental Learning (Domain-IL)}, and \textit{Class-Incremental Learning (Class-IL)}.
While these scenarios cover some desiderata of CL, they fail to meet the requirement of TA~\cite{bib59}; see Fig.~\ref{MNIST_scenarios}. 
They assume that tasks have clear and well-defined boundaries during training. 
Therefore, recent studies proposed expanding the CL scenarios by introducing \textit{Discrete TA (DTA)}~\cite{bib82}, \textit{Continuous TA (CTA)}~\cite{bib82}, and \textit{Open-World TA (OWTA)}~\cite{bib59}. 

\begin{table*}[h]
\centering
\footnotesize
\caption{The CL scenarios based on the difference between $\mathcal{D}_{t-1}$ and $\mathcal{D}_t$, (adapted from \cite{bib38}). \textit{p}($\mathcal{X}$) is the input data distribution; \textit{p}($\mathcal{Y})$ is the target label distribution; \{$\mathcal{Y}_{t-1} \neq \mathcal{Y}_t\}$ denotes that output space are from a disjoint space which is separated by task identity (Task ID).}
\begin{tabular*}{\textwidth}{@{\extracolsep\fill} ccccccc}
\hline%
\textbf{CL} & 
\multicolumn{3}{c}{\textbf{Difference between $\mathcal{D}_{t-1}$ and $\mathcal{D}_t$}} 
& \multicolumn{2}{c}{\textbf{Task ID}} 
& \textbf{Online} \\
\cline{2-6}
\textbf{Scenario} & \begin{tabular}[c]{@{}c@{}} \textit{p}$(\mathcal{X}_{t-1}) \neq$ \textit{p}$(\mathcal{X}_t)$ \end{tabular} & 
\begin{tabular}[c]{@{}c@{}} \textit{p}$(\mathcal{Y}_{t-1}) \neq$  \textit{p}$(\mathcal{Y}_t)$ \end{tabular} & 
\begin{tabular}[c]{@{}c@{}} $\{\mathcal{Y}_{t-1} \neq$ $\mathcal{Y}_t$\} \end{tabular} & \textbf{Train} & \textbf{Test} &  \textbf{Learning} \\

\hline
Task-IL   & \checkmark & \checkmark & \checkmark & Known         & Known          & No \\ 
Domain-IL & \checkmark & $\times$   & $\times$   & Known         & Not required   & Optional \\
Class-IL  & \checkmark & \checkmark & $\times$   & Known         & To be inferred & Optional \\
DTA       & \checkmark & $\times$   & $\times$  & Not required   & Not required    & Yes \\
OWTA      & \checkmark & \checkmark & $\times$  & To be inferred & To be inferred    & Yes \\
CTA       & $\times$   & $\times$   & $\times$  & Unknown  & Unknown   & Yes \\
\hline
\end{tabular*}
\label{CL_Scenarios} 
\end{table*}

\subsubsection{Task-Incremental Learning (Task-IL)}
\label{2.3.1}

In Task-IL, a model sequentially learns to solve a number of distinct tasks~\cite{bib128, bib129, bib130}. 
Each task has disjoint output spaces $\{\mathcal{Y}_{t-1} \neq \mathcal{Y}_t\}$ and Task IDs are known in training and testing phases. 
Typically, Task-IL employs \textit{multi-headed model} (i.e., an output head for each task). 
An NN model equipped with a multi-headed output layer can accommodate various tasks, making \textit{p}($\mathcal{Y}_{t-1}) \neq \textit{p}(\mathcal{Y}_t)$ and \textit{p}($\mathcal{X}_{t-1}) \neq \textit{p}(\mathcal{X}_t)$ true (Table~\ref{CL_Scenarios}). 

%%%%% -----
\subsubsection{Domain-Incremental Learning (Domain-IL)} 
\label{2.3.2}

In Domain-IL, a model learns to solve the same problem in different tasks~\cite{bib128, bib129, bib130}.
Tasks have the same data label space $\{\mathcal{Y}_{t-1}=\mathcal{Y}_t\}$ but different input distributions $\textit{p}(\mathcal{X}_{t-1}) \neq \textit{p}(\mathcal{X}_t)$.  
The model does not require Task IDs in the inference phase, as each task has the same possible outputs (e.g., same classes in each task). 
The use of a \textit{single-headed model} (i.e., same output head for every task) ensures that the output space remains consistent. 
An example of Task-IL is incrementally learning to recognize objects midst varying lighting conditions (e.g., indoor and outdoor)~\cite{bib131}.

%%%%% -----
\subsubsection{Class-Incremental Learning (Class-IL)}
\label{2.3.3}

In Class-IL, a model must distinguish the global labels (classes)~\cite{bib128, bib129, bib130}. 
Here, Task IDs are only provided in the training phase, as they will be inferred in the inference phase. 
Due to the multi-class property, $\textit{p}(\mathcal{Y}_{t-1}) \neq \textit{p}(\mathcal{Y}_t)$ is a natural consequence, thereby posing more challenges as compared to Task-IL and Domain-IL. 

%%%%% -----
\subsubsection{Discrete Task Agnostic (DTA)}
\label{2.3.4}

In DTA, a model learns from a sequence of distinct tasks, but without the need for inferring the Task ID during training and inference phases~\cite{bib59, bib82}.
The input distributions differ between tasks, making \textit{p}($\mathcal{X}_{t-1}) \neq \textit{p}(\mathcal{X}_t)$ true. 
Meanwhile, the target label across tasks have the same distributions, and the output space is the same across tasks, making \textit{p}($\mathcal{Y}_{t-1}) \neq \textit{p}(\mathcal{Y}_t)$ and $\{\mathcal{Y}_{t-1} \neq \mathcal{Y}_t\}$ false.

%%%%% -----
\subsubsection{Open-World Task Agnostic (OWTA)}
\label{2.3.5}

In OWTA, a model learns from a sequence of distinct tasks, but needs to infer the Task ID during training and inference phases~\cite{bib59, bib82}.
It requires the CL algorithm to handle an open set of tasks and classes, potentially facing unknown categories that are not encountered during initial training. 
This makes the OWTA as one of the most challenging scenarios.   

%%%%% -----
\subsubsection{Continuous Task Agnostic (CTA)}
\label{2.3.6}

In CTA, data stream is represented as a continuous function over time \textit{without explicit task boundaries}~\cite{bib59, bib82}, which makes CTA as the most challenging scenario.
Here, a CL algorithm must learn from evolving distribution without any task-specific guidance, as neither task boundaries are clear nor Task ID is known during the training phase~\cite{bib59}, making \textit{p}($\mathcal{X}_{t-1}) \neq \textit{p}(\mathcal{X}_t)$, \textit{p}($\mathcal{Y}_{t-1}) \neq \textit{p}(\mathcal{Y}_t)$ and \{$\mathcal{Y}_{t-1} \neq \mathcal{Y}_t\}$ false. 

%%%%%%%%%%%%%%%%%%%%%%%%%%%%%%%%%%%%%%%%%
\subsection{Online Continual Learning (OCL)}

In OCL, a model learns from a continuous data stream, so that each input sample during inference is used as training data for updating the models' knowledge~\cite{bib125,bib126,bib38}. 
As a new sample arrives, it is immediately used for inference, and for updating the model on-the-fly. 
\textit{This capability is highly desired for systems that face highly dynamic environments (e.g., autonomous mobile agents), where new data is encountered in real-time and immediate updates to the model are required}. 
Its challenges include integrating new knowledge and retaining previous information, while considering limited memory and power/energy budgets.
In this survey, we emphasize that OCL is the expected capability for embedded AI systems.

%%%%%%%%%%%%%%%%%%%%%%%%%%%%%%%%%%%%%%%%
\subsection{CL Methods} 
\label{sec2.5}

In this sub-section, we address the key question Q2, by providing an overview of the existing methods that address CF in various CL scenarios, and the desiderata they fulfill; summarized in Table~\ref{table_md}. 
These methods are categorized into five approaches based on \textit{representation}, \textit{regularization}, \textit{rehearsal/replay}, \textit{optimization}, and \textit{architecture}~\cite{bib19}, as follows. 

%%%%----
\subsubsection{Representation-based Approach}

\begin{figure}[t]
\centering
\includegraphics[width=0.5\textwidth]{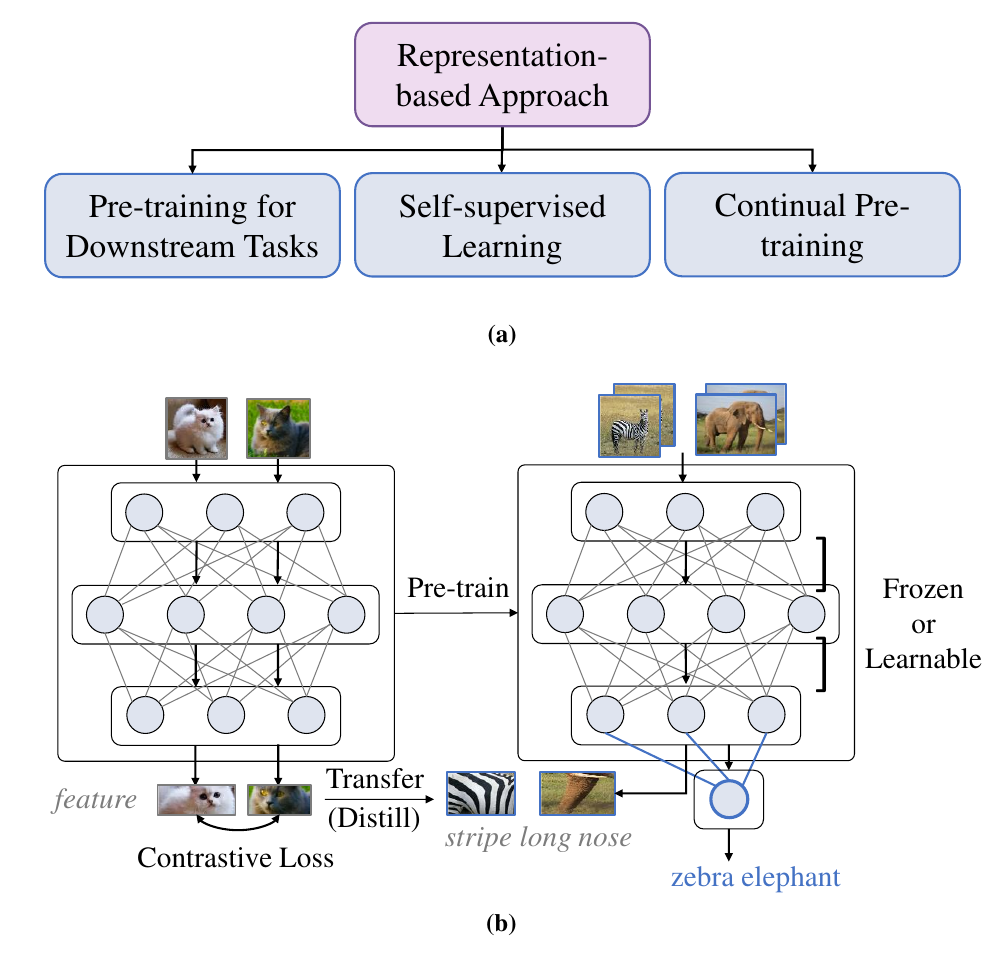} 
\caption{An illustration of the taxonomy of (a) representation-based approach and (b) its respective methods; adapted from studies in~\cite{bib19}.} 
\label{rep}
\end{figure}

This approach is characterized by exploiting the strengths of data representations~\cite{bib19}; see Fig.~\ref{rep}. 
The related methods are as follows.

\smallskip
\begin{enumerate}
    \item \textbf{Self-Supervised Learning (SSL)}: 
    It makes the model learn to generate useful data representations without relying on explicit labels, by creating supervisory signals from the input data to uncover the underlying structure of the data~\cite{bib167,bib168}. 
    Related studies are discussed in~\cite{bib141, bib142, bib143, bib144, bib145, bib146, bib171}.
    \smallskip
    \item \textbf{Pre-training for Downstream CL}: 
    It is the initial phase of model training on a large diverse dataset to learn general-purpose representations, which can be fine-tuned for specific downstream tasks, hence having strong knowledge transfer~\cite{bib101, bib148, bib149, bib150}. 
    To maintain generalizability for future tasks, several strategies have been developed, such as 
    \textit{Pre-trained Representations}~\cite{bib151, bib152, bib106, bib154, bib156}, 
    \textit{Task-adaptive Prompts}~\cite{bib157, bib158, bib159, bib160, bib161, bib298}, 
    \textit{Saving Prototypes and Enhancing Classifiers}~\cite{bib162,bib163,bib164}, and
    \textit{Optimizing an Updatable Backbone}~\cite{bib165, bib174, bib101, bib175, bib176}.
    \smallskip
    \item \textbf{Continual Pre-Training (CPT)}: 
    It includes techniques studied in~\cite{bib178, bib180, bib182, bib183, chen2021incremental, toda2023growing, bib179}.
\end{enumerate}

\subsubsection{Regularization-based Approach}

\begin{figure}[t]
\centering
\includegraphics[width=0.5\textwidth]{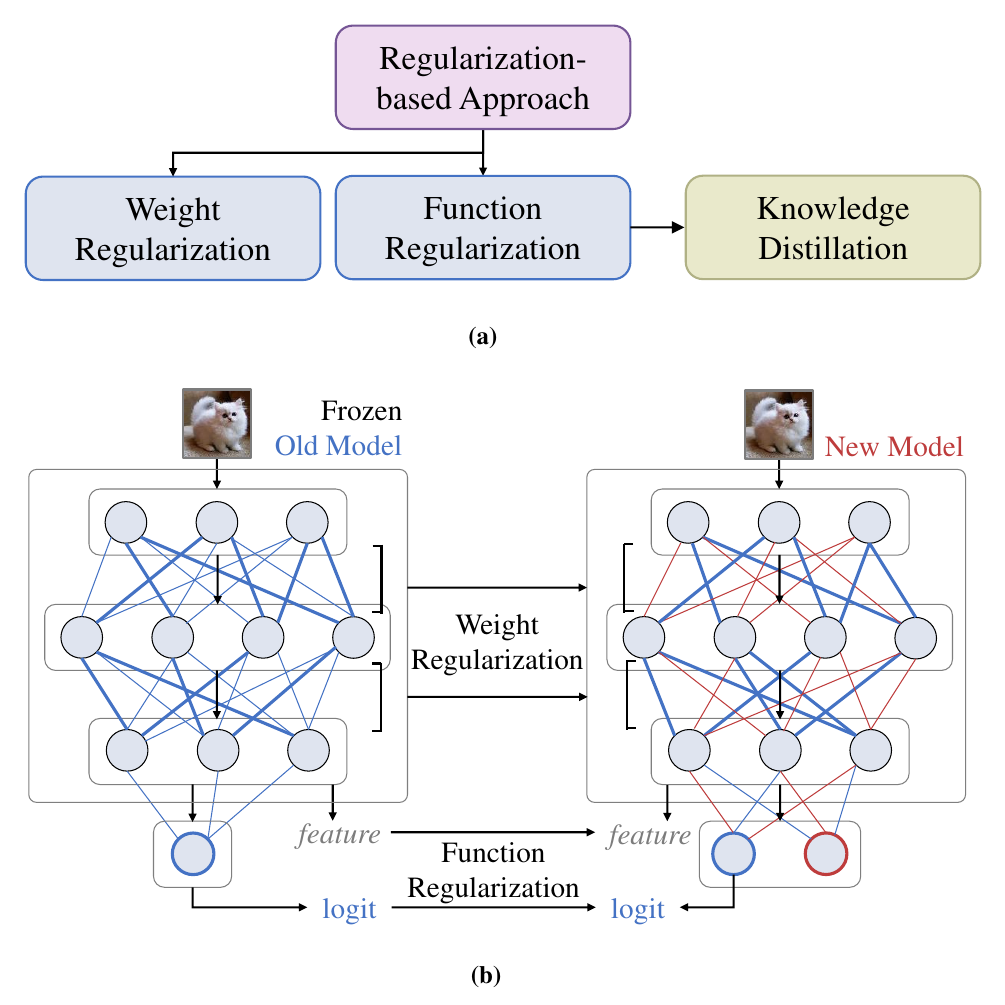} 
\caption{An illustration of the taxonomy of (a) regularization-based approach and (b) its respective methods; adapted from studies in~\cite{bib19}.} 
 \label{reg}
\end{figure}

This approach is characterized by adding explicit regularization terms to balance the old and new tasks~\cite{bib19}, which usually requires storing a frozen copy of the old model for reference; see Fig.~\ref{reg}. 
The related methods are as follows.

\begin{enumerate}
    \item \textbf{Weight Regularization}:  
    It includes techniques such as \textit{Elastic Weight Consolidation (EWC)}~\cite{bib61}, \textit{Synaptic Intelligenc (SI)}~\cite{bib122}, \textit{Memory Aware Synapses (MAS)}~\cite{bib187}, \textit{Riemannian Walk (RWalk)}~\cite{bib134}.
    Meanwhile, other techniques are categorized into  
    \textit{Expansion-Renormalization}~\cite{bib62, bib192, bib193, bib103}, 
    \textit{Quadratic Penalty Refinement}~\cite{bib189, bib190, bib191}, 
    and \textit{Online Variational Inference}~\cite{bib196, bib197, bib198, bib199, bib194, bib109, bib195}. 
    \smallskip
    \item \textbf{Function Regularization}: 
    It regulates the variations in models' function $f_\theta$ over time, and applies constraints directly to the function. 
    A generalized form of function regularization can be expressed as:

    \begin{equation} \label{eq3}
    \boldsymbol{\theta}_t = \arg \min_{\boldsymbol{\theta}} \left( \frac{1}{N_t} \sum_{i=1}^{N_t} \mathcal{L}_t\big(f_{\boldsymbol{\theta}}(\mathbf{x}_i), y_i\big) + \lambda \, \mathcal{R}_f\big(f_{\boldsymbol{\theta}}, \mathcal{D}_{1:t-1}\big) \right)
    \end{equation}

    where, $\boldsymbol{\theta} \in \mathbb{R}^d$ denotes the model parameters (weights) after learning task $t$.
    The term $\frac{1}{N_t} \sum_{i=1}^{N_t} \mathcal{L}_t(f_\theta(\mathbf{x}_i), y_i)$ calculates the average loss on the current task $t$, where $N_t$ is the number of samples in task $t$, $\mathbf{x}_i \in \mathbb{R}^n$ is the input feature vector for the $i^{\text{th}}$ training sample for task $t$, $y_i$ is the corresponding ground truth label for $\mathbf{x}_i$, $f_\theta(\mathbf{x}_i)$ is the model’s prediction for input $\mathbf{x}_i$ using parameters $\boldsymbol{\theta}$ and $\mathcal{L}_t$ is the task-specific loss function (e.g., cross-entropy for classification).
    $\mathcal{R}_f(f_\theta, \mathcal{D}_{1:t-1})$ is the function regularization term that penalizes changes in the models' function $f_\theta$ based on its performance on previous data $\mathcal{D}_{1:t-1}$ from tasks 1 to $t-1$ and $\lambda$ is a hyperparameter controlling the strength of the regularization i.e., the balance between stability and plasticity. The goal is to find the set of weights $\boldsymbol{\theta}_t$ that minimize both the loss on the current task and the deviation from prior knowledge.
    \smallskip
    \item \textbf{Knowledge Distillation (KD)}: 
    It utilizes the previously-trained model as the teacher and the currently-trained model as the student~\cite{bib19}. 
    It includes techniques studied in~\cite{bib177, bib135, bib200, bib201, bib202}. 
    \smallskip
    \item Besides the above prominent methods, several alternatives have also been proposed: \textbf{Sequential Bayesian Inference over Function Space}~\cite{bib203, bib204, bib205} and \textbf{Conditional Generation}~\cite{bib206, bib207, bib208}.
\end{enumerate}

\begin{figure}[h]
\centering
\includegraphics[width=0.5\textwidth]{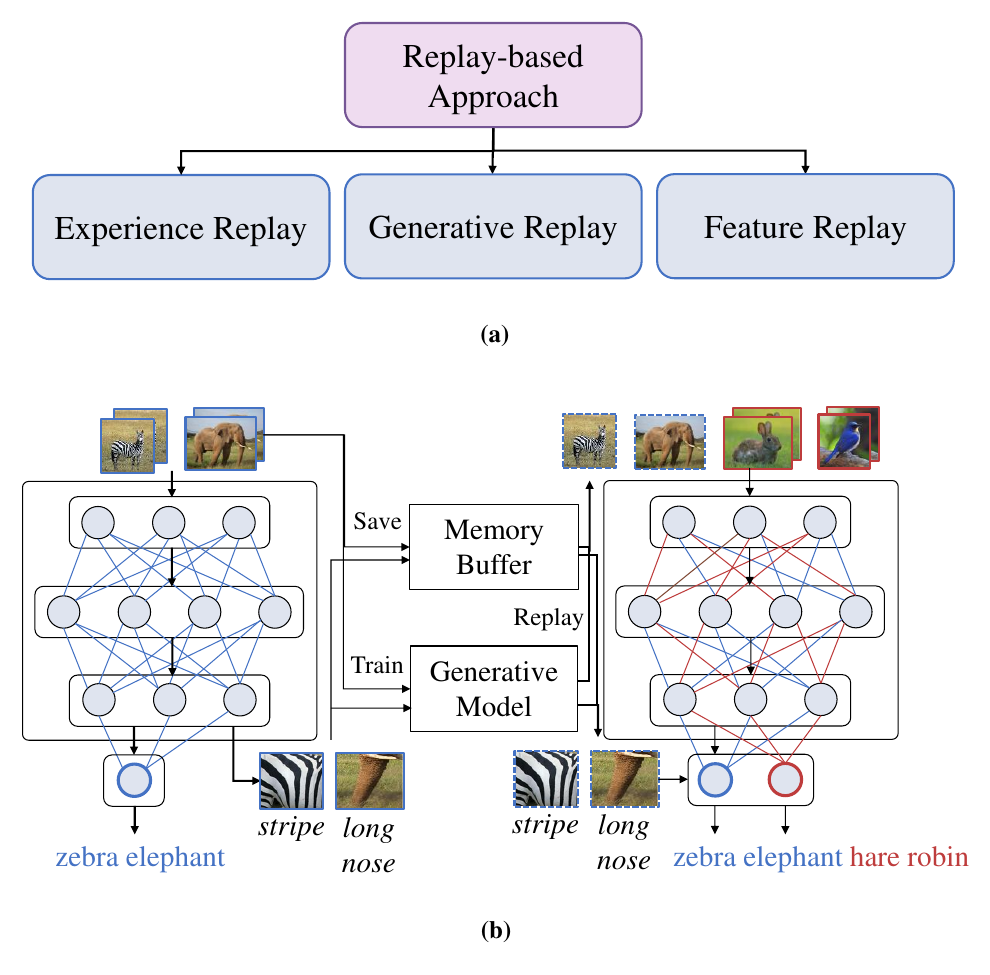} 
\caption{An illustration of the taxonomy of (a) replay-based approach and (b) its respective methods; adapted from studies in~\cite{bib19}.} 
 \label{replay}
\end{figure}

\subsubsection{Rehearsal/Replay-based Approach} 
\label{2.5.3}

This approach is characterized by approximating and recovering old data distributions~\cite{bib19}; see Fig. \ref{replay}. 
The related methods are discussed in the following.

\smallskip
\begin{enumerate}
    \item \textbf{Experience Replay (ER)}: 
    It stores a few old training samples in a small memory buffer. 
    It includes several strategies such as 
    \textit{Sample Selection (SS)}~\cite{bib99, bib209, bib100, bib135}, 
    \textit{Gradient/Optimization-based Techniques}~\cite{bib125, bib210, bib211, bib212, bib213, bib214}, 
    \textit{Storage Efficiency}~\cite{bib215, bib216, bib124, bib213, bib217, xie2022isvm}, 
    \textit{Integration of ER with KD}~\cite{bib135, bib218}, 
    \textit{Mitigation of Data Imbalance}~\cite{bib136, bib219, bib221, bib207, bib222, bib146, bib224, bib225, bib226, de2021continual}, 
    \textit{Learning Plasticity Enhancement}~\cite{mai2021supervised, bib228}, and
    \textit{Overfitting Alleviation}~\cite{bib229, bib230, bib231, bib232, boschini2022class}.
    \item \textbf{Generative Replay}: It uses an additional generative model such as \textit{Generative Adversarial Networks (GANs)}~\cite{bib238} or \textit{Variational Autoencoders (VAEs)}~\cite{bib239} to replay the previously-learned data~\cite{bib24}.
    There are several GANs-based techniques as studied in~\cite{bib64, bib206, bib123, bib233, bib154}.
    However, they usually suffer from label inconsistency~\cite{bib123}.
    Toward this, other techniques employ \textit{autoencoder-based strategies}~\cite{bib236, ayub2021eec, bib24}.

    \smallskip
    \item \textbf{Feature Replay}: It includes techniques studied in~\cite{bib25, bib26, bib240, bib241, bib242, bib243, wang2021acae, bib245}.
\end{enumerate}

\subsubsection{Optimization-based Approach} 

\begin{figure} [h]
\centering
\includegraphics[width=0.5\textwidth]{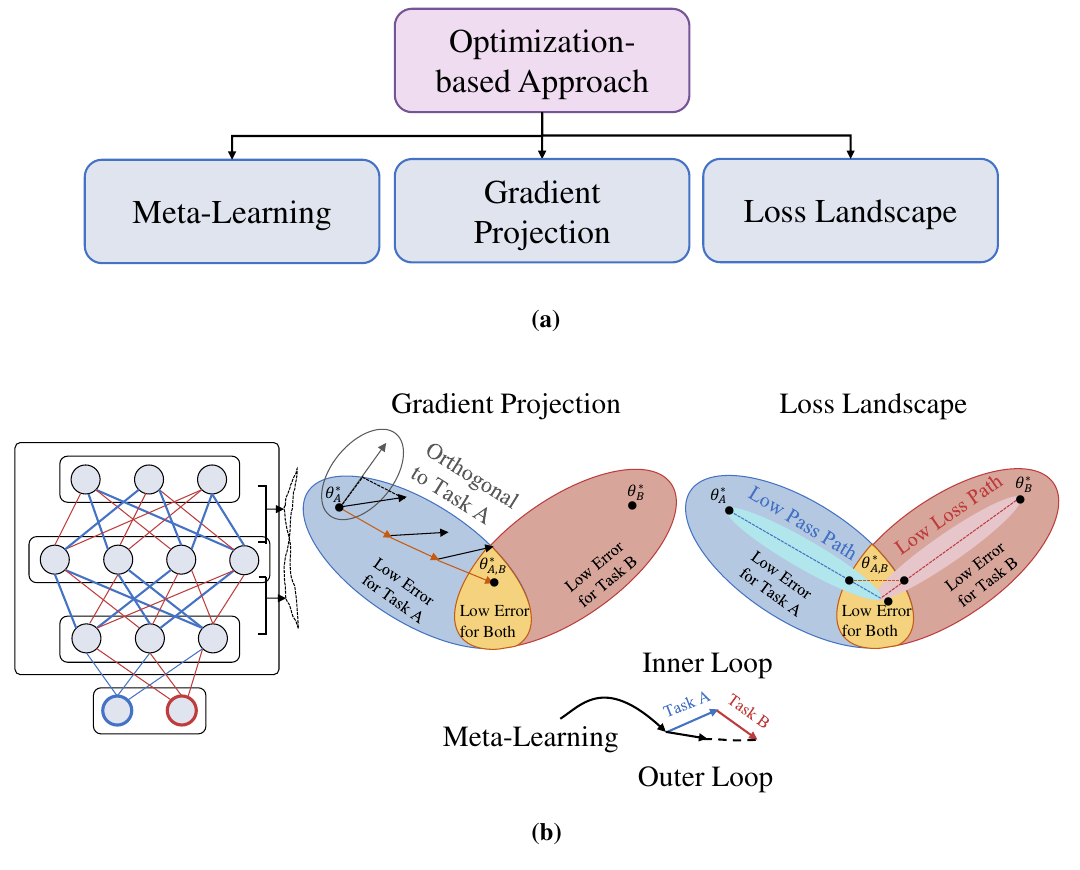} 
\caption{An illustration of the taxonomy of (a) optimization-based approach and (b) its respective methods; adapted from studies in~\cite{bib19}.} 
\label{opt}
\end{figure}

This approach is characterized by explicitly designing and manipulating the optimization programs~\cite{bib19}; see Fig.~\ref{opt}. 
It includes the following methods.
\smallskip
\begin{enumerate}
    \item \textbf{Gradient Projection}: It includes techniques studied in~\cite{bib96, bib94, bib277, bib63, bib278, bib95, bib279, bib280, bib282, bib281} and the ones with \textit{Replay-based Strategies}~\cite{bib100, bib275, bib276, bib99}.
    \smallskip
    \item \textbf{Loss Landscape}: The related techniques are studied in~\cite{bib283, bib285, bib286}.
    \smallskip
    \item \textbf{Meta-Learning}: 
    It is also known as \textit{learning-to-learn} for CL, which attempts to obtain a data-driven inductive bias for various scenarios, rather than designing it manually~\cite{bib18}. 
    It includes techniques studied in~\cite{bib102, bib288, bib289, bib99, bib83, bib290, bib291, bib292, bib293, bib294, bib295}.
\end{enumerate}

%%%%----
\subsubsection{Architecture-based Approach} 

\begin{figure} [t]
\centering
\includegraphics[width=0.46\textwidth]{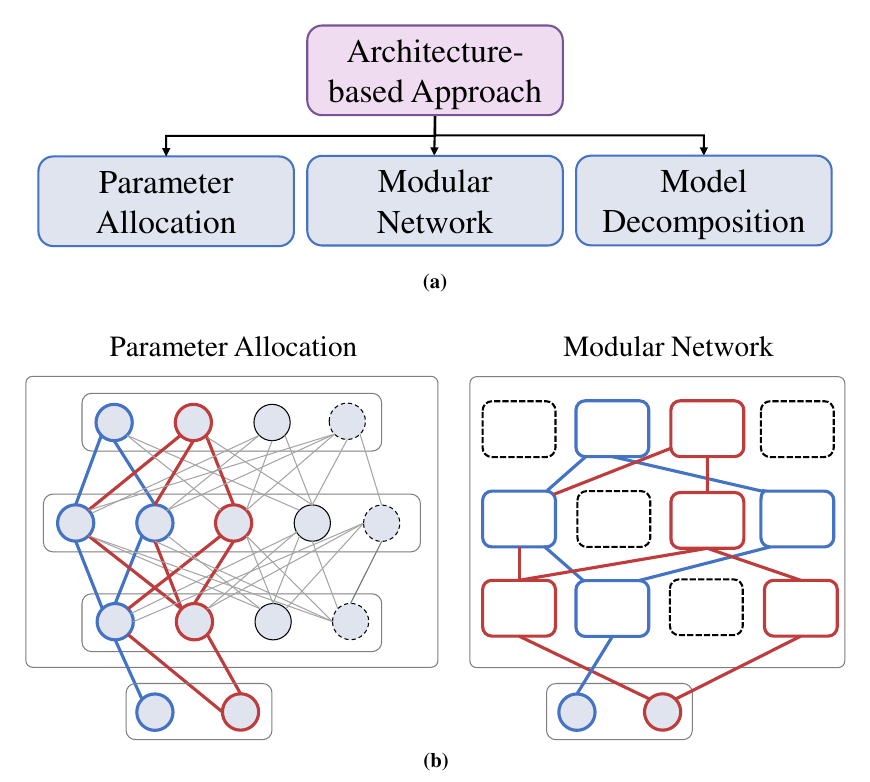} 
\caption{An illustration of the taxonomy of (a) architecture-based approach and (b) its respective methods; adapted from studies in~\cite{bib19}.} 
 \label{arch}
\end{figure}

This approach is characterized by constructing task-specific parameters, that can explicitly resolve the inter-task interference caused by incremental task learning with shared parameters~\cite{bib19}; see Fig.~\ref{arch}.
It encompasses the following methods.

\smallskip
\begin{enumerate}
    \item \textbf{Parameter Allocation}:
    It isolates parameter subspace, which is dedicated to each task throughout the model, where the architecture can be \textit{fixed} or \textit{dynamic} in size. 
    It includes strategies like \textit{Fixed Architecture}~\cite{bib60, bib107, bib195, bib246, bib247, bib248, bib249, bib109} and \textit{Dynamic Architecture}~\cite{bib256,bib257}.
    \smallskip
    \item \textbf{Modular Network}: 
    It leverages parallel sub-networks to learn new tasks one-by-one, without pre-defined task-sharing or task-specific components.
    The related techniques are studied in~\cite{bib104, bib268, bib105, bib269, bib270, bib271, bib272, bib273, bib274}.
    \smallskip
    \item \textbf{Model Decomposition}: 
    It separates a model into the task-sharing and task-specific components~\cite{bib19} (Fig.~\ref{arch1}).
    The relate techniques are studied in~\cite{bib259, bib260, bib261, bib262, bib263, bib264, bib265, bib266}. 
\begin{figure} [t]
        \centering
        \includegraphics[width=0.5\textwidth]{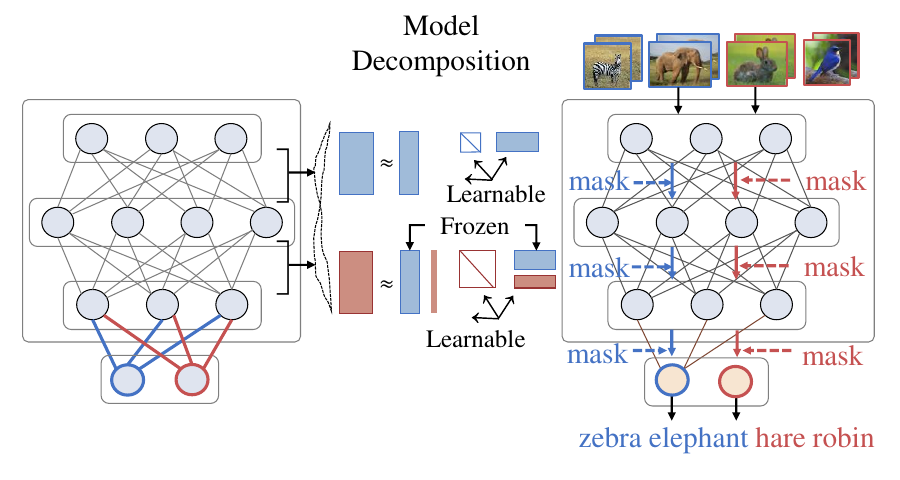} 
        \caption{An illustration of the model decomposition method exhibiting two types; corresponding to parameters (i.e., low-rank factorization) and representations (i.e., masks of intermediate features); adapted from studies in~\cite{bib19}.} 
        \label{arch1}
    \end{figure}
\end{enumerate}

\subsection{Limitations of Existing CL Methods}
The CL desiderata fulfilled by the existing CL methods (discussed in Section~\ref{sec2.5}) are presented in Table~\ref{table_md}. 
This table shows that, \textit{different CL methods feature different advantages and disadvantages}. Furthermore, the limitations of existing CL methods in DNNs and the challenges that impede their direct application to SNNs are discussed as follows, answering key question Q3. A summary of these limitations is provided in Table~\ref{Limitations}. For example, weight regularization with methods like EWC~\cite{bib61} and SI~\cite{bib122} requires estimating and storing importance measures (e.g., FIM calculations) for every synapse, adding a quadratic penalty term for each learned task, which is resource-intensive in high-dimensional, event-driven SNNs, therefore has a linearly increasing computational cost~\cite{bib36}. Moreover, such importance scores assume a static activation landscape, which is unstable in SNNs due to variable spiking activity and dynamic thresholds.
Replay-based methods (e.g., iCaRL~\cite{bib135}, DGR~\cite{bib64}, GEM~\cite{bib100}) require storing and reprocessing raw input samples or features. This is not well suited for neuromorphic hardware, which operates under strict memory and power constraints and lacks the infrastructure to store dense data or continuously replay high-dimensional inputs efficiently. 
Many CL methods in DNNs rely on precise gradient computations through differentiable activation functions. In contrast, SNNs involve discontinuous spike events and temporal dynamics, making conventional BP infeasible or biologically implausible. Although SG methods exist, they are computationally expensive.
Moreover, spike-based learning involves temporally extended input encoding. Mapping gradient-based loss functions from DNN-CL to temporally sparse spike trains introduces ambiguity in assigning credit across time, which is non-trivial and often requires costly unrolling (e.g., through BPTT). 
Most DNN-based CL methods assume synchronous, frame-based computation. SNNs operate asynchronously, responding to sparse events. This fundamental mismatch makes it difficult to translate CL mechanisms designed for batch updates and full forward/backward passes into the SNN regime without re-engineering their entire workflow.

These limitations in adapting conventional CL strategies to SNNs underscore the critical need for a paradigm shift towards neuromorphic-specific CL algorithms that can achieve higher performance while minimizing energy consumption.
This is especially important for deployments in embedded AI systems (e.g., mobile robots/agents and IoT-Edge) where compute and memory budgets are limited, highlighting the reasons why the energy-efficient CL algorithms are crucial for resource-constrained embedded AI systems posed by the key question Q3.

Motivated by the limitations of existing CL methods, NCL concept has emerged as a potential solution as it has characteristics that align with the desiderata of CL. 
Specifically, NCL leverages SNN computation models which has event-driven operations, thus enabling energy-efficient learning.
Furthermore, SNNs can perform unsupervised learning due to their bio-plausible learning rules.
Details of NCL concept and review of NCL methods that are explicitly designed to operate under the constraints of event-driven computation, limited memory, and hardware variability inherent to neuromorphic systems will be discussed in Section~\ref{sec3}.

\begin{sidewaystable*}
\caption{Representation-, Regularization-, Rehearsal/Replay- and Architecture-based CL methods that potentially fulfill ``+'' or have fulfilled ``\checkmark'' the desiderata of CL.}
\label{table_md}
\centering
\begin{tabular*}{\textheight}{@{\extracolsep\fill}cccccccccccc}
\cline{1-10}

& & & \multicolumn{7}{@{}c@{}}{\textbf{Desiderata of CL}} & \\
\cline{4-10}
\textbf{Approach} & \textbf{Methods} & \textbf{Work,(Year)} &\\
& & & Scalability & No/Minimal & Task & Positive & Positive & Controlled & Fast & & \\
& & & & Old Data Use & Agnostic & FWT & BWT & Forgetting & Adaptation \\
\cline{1-10}
&  & \cite{bib66},(2023) & & \checkmark & & & \\
& Pre-training for & \cite{bib73},(2023) & \checkmark & 
 \checkmark & & & \\
& Downstream & \cite{bib70},(2024) & & \checkmark & & & \\
& Tasks & \cite{bib74},(2024) & \checkmark & \checkmark & & & \\
& & MAML \cite{bib110},(2017) & & & & & & & \checkmark \\ 
\cline{2-10}
& Self-Supervised & \cite{bib67},(2023) & & \checkmark \\
Representation & Learning &&&\\
\cline{2-10}
& & \cite{bib86},(2019) & & &\checkmark \\
& & \cite{bib77},(2022) & & &\checkmark \\
& Continual & TADIL \cite{bib76},(2023) & & &\checkmark \\
& Pre-training & \cite{bib93},(2022) & & & & \checkmark & & \\
& & AFEC \cite{bib103},(2021) & & & & \checkmark & & \\
& & CFN \cite{bib72},(2020) & & & & & & \checkmark & \\ 
\cline{1-10}
& & GPM \cite{bib63},(2021) & \checkmark & & & & \\
& & \cite{bib65},(2022) &  & \checkmark & & & \\
& & \cite{bib78},(2022) &  & \checkmark & & & & & \\
& & \cite{bib82},(2018) &  & & \checkmark & & & & \\
& Weight  & \cite{bib79},(2023) & & \checkmark  & & & \\
& Regularization & OGD \cite{bib94},(2020) & & \checkmark  & & & \\
Regularization  & & TRGP \cite{bib95},(2022) & & & & \checkmark & \\
& & OWM \cite{bib96},(2019) & \checkmark & & & & \\
& & ASP \cite{bib108},(2017) & & & & & & \checkmark & &\\
& & UCL \cite{bib109}, (2019) & & & & & & \checkmark & &\\
\cline{2-10}
& Function & HCL \cite{bib85},(2021) & &  & \checkmark & & \\
& Regularization & SGP \cite{bib97},(2023) & & \checkmark & \\ \cline{2-10}
& Distillation & \cite{bib75},(2023) & & \checkmark & \\
\cline{1-10}
& & \cite{bib92},(2022) & & & & \checkmark & & \\
& Experience & \cite{bib98},(2019) & & & + & \checkmark & \checkmark & & \\
Replay & Replay & MER \cite{bib99},(2019) & & & + & \checkmark & \checkmark & \\
& & GEM \cite{bib100},(2017) & & & & \checkmark & \checkmark & \\
\cline{2-10}
& Generative & TAME \cite{bib81},(2024) & & & \checkmark & \\
\cline{1-10}
& & CN-DPM \cite{bib84},(2020) & & & \checkmark & \\
& Model & DSSAE \cite{bib87},(2023) & & & \checkmark & \\
Architecture & Decomposition & P\&C \cite{bib62},(2018) & \checkmark & & & \checkmark & & \checkmark & & \\
& & CLNP \cite{bib107},(2019) & & & & & & \checkmark & &  \\
\cline{2-10}
& Allocation & AFAF \cite{bib91},(2022) & & & & \checkmark & \\
\cline{1-10}
\end{tabular*}
\end{sidewaystable*}

%%%%%%%%%%%%%%%%%%%%%%%%%%%%%%%%%%%%%%%%%%%%%%%%%%%%%%%%%%%%%%%%%%%%%%%%%%%%%%%%%%%%%%%
\begin{table*}[h]
\centering
\footnotesize
\caption{Summary of ANN-based CL methods, outlining their computational/memory overhead.}
\begin{tabular}{p{2cm}p{2.5cm}p{2.7cm}p{8.8cm}}
\cmidrule{1-4}%
\textbf{Approach} & \textbf{CL Method} & \textbf{Network/Model} & \textbf{Limitations} \\
\cmidrule{1-4}
Regularization-based & EWC~\cite{bib61}, \newline SI~\cite{bib122}, MAS~\cite{bib187}, \newline RWalk~\cite{bib134}, \newline R-EWC~\cite{bib189}, \newline XK-FAC~\cite{bib190}, \newline ALASSO \newline ~\cite{bib191}, \newline GD~\cite{bib202} & MLP, CNN \newline  \newline \newline LeNet,  VGG-16 \newline ResNet-18 \newline ResNet-50 \newline ResNet-18 \newline Wide Residual Network & - Linearly increasing computations with the number of parameters and tasks due to Fisher Information Matrix (FIM) calculations. \newline - High computations due to factorized rotation of the parameter space to diagonalize the FIM. \newline - High computations due to the complexity of handling batch normalization. \newline - High computations due to more accurate quadratic approximation of the loss function for improving performance. \newline - High computations due to large-scale distillation over unlabeled data.\\ 
\cmidrule{1-4}%
Replay-based & GEM~\cite{bib100}, \newline A-GEM\cite{bib275}, \newline DGR~\cite{bib64}, \newline RAR~\cite{bib217}, \newline DGM~\cite{bib123}, \newline L-VAEGAN~\cite{bib24}, \newline EEC~\cite{ayub2021eec} & MLP, ResNet-18 \newline \newline GAN, \newline MLP, ResNet-18 \newline DCGAN, ResNet-18, \newline VAE-GAN,\newline DCGAN, ResNet-18 & - High computations and memory overhead during training due to the requirement for backward passes. \newline - High computations due to adversarial training, high memory for storing and generating past data. \newline - High computations due to adversarial training with sparse attention masks. \newline - High computations due to combination of VAEs with GANs. \newline - High computations and memory overhead  due to encoding and decoding of entire past episodes.\\
\cmidrule{1-4}%
Representation-based & Co2L~\cite{bib146}, \newline DualNet~\cite{bib141} & ResNet-18, \newline Two parallel CNNs & - High computations due to contrastive learning. \newline - Memory overhead due to maintaining two networks. \\

& TwF~\cite{bib106}, \newline GAN-Memory~\cite{bib154}, CODA-Prompt~\cite{bib158}, S-Prompts~\cite{bib160}, Barlow Twins + EWC~\cite{bib178} & ResNet-18, \newline GAN, \newline Vision Transformer (ViT), \newline Transformer-based \newline ResNet-50 & - Memory overhead due to parallel networks. \newline - High computations due to generative modeling. \newline - High computations due to attention calculations of ViTs leading to quadratic complexity in terms of input size. \newline - High memory overhead as prompts grow with tasks. \newline
- High computations due to quadratic penalty in EWC. \\
\cmidrule{1-4}%
Architecture-based & PNN~\cite{bib104}, \newline PathNet~\cite{bib105}, \newline Model Zoo~\cite{bib272} & MLP, CNN \newline CNN, \newline Wide-Resnet & - Exponential memory increase due to new sub-networks added for new tasks. \newline - High computations due to multiple parallel networks. \newline - Linearly increasing memory due to a separate sub-network storage for each task, and ensemble of all sub-networks. \\
\cmidrule{1-4}%
\end{tabular}
\label{Limitations} 
\end{table*}
%%%%%%%%%%%%%%%%%%%%%%%%%%%%%%%%%%%%%%%%%%%%%%%%%%%%%%%%%%%%%%%%%%%%%%%%%%%%%%%%%%%%%%%
\section{Neuromorphic Computing} 
\label{sec3}

%%%%%%%%%%%%%%%%%%%%%%%%%%%%%%%%%%
\subsection{Overview of SNNs}
\label{sec3.1}

\begin{figure}[h]
    \centering
    \includegraphics[width=0.46\textwidth]{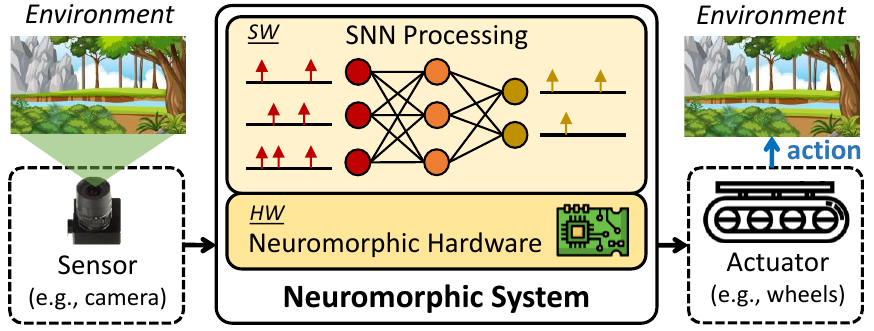} 
    \caption{An overview of a neuromorphic system, encompassing hardware (HW) and software (SW) parts.}
    \label{Fig_NeuromorphicSystem}
\end{figure}

A neuromorphic computing system encompasses SNN processing (including spike-based operations and training) in the software part, and a neuromorphic processor in the hardware part. 
The overview of the system is illustrated in Fig.~\ref{Fig_NeuromorphicSystem}, and described in the following, answering key question Q4.

\begin{figure}[t]
    \centering
    \includegraphics[width=0.46\textwidth]{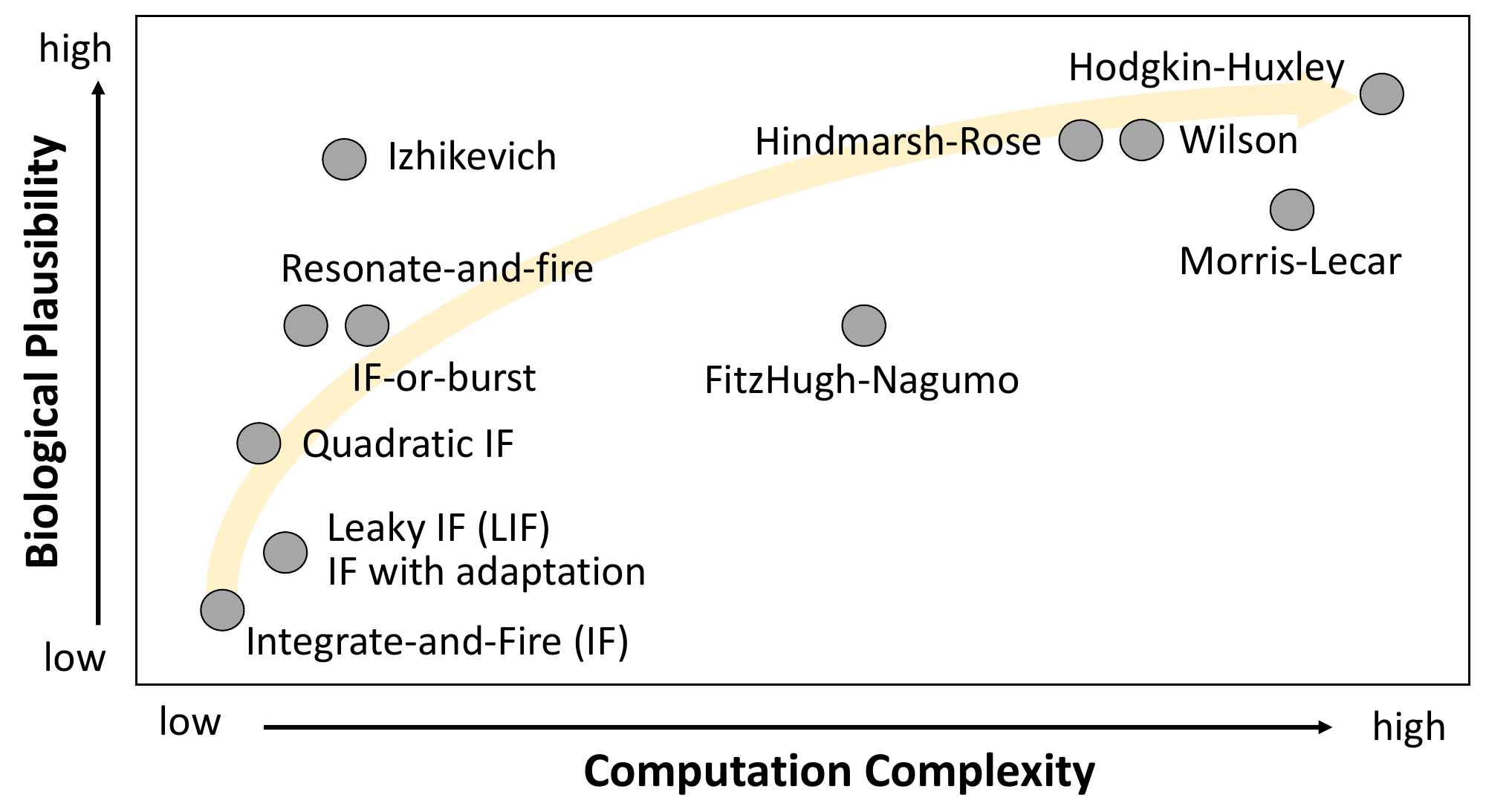} 
    \caption{Trade-off between bio-plausibility and computation complexity for different neuron models; adapted from studies in~\cite{Ref_Izhikevich_WhichNeuron_TNN04}.} 
    \label{Fig_NeuronModels}
\end{figure}

\textbf{SNN Models and Operations:} 
SNNs mimic the brains' functionality through the utilization of spikes for transmitting information~\cite{bib55, Ref_Putra_FSpiNN_TCAD20, Ref_Putra_EmbeddedSNNs_Springer24}. 
Therefore, SNNs encode information into discrete spike trains. 
Popular encoding techniques include \textit{rate coding}~\cite{Ref_Diehl_SNN_FNCOM15} (e.g., spike count, spike density, population activity~\cite{Ref_ElSayed_PhDthesis_SorbonneUniv21}), and \textit{temporal coding} (e.g., burst~\cite{Ref_Park_BurstCoding_DAC19}, time-to-first spike (TTFS)~\cite{Ref_Park_T2FSNN_DAC20}, phase~\cite{Ref_Guo_NeuralCoding_FNINS21}, and rank-order~\cite{Ref_Thorpe_RankOrder_CompNeuro98}). 
Each spiking neuron processes the input spikes, and its internal behavior (i.e., \textit{neuronal dynamics}) depends on the spiking neuron model, such as \textit{Hodgkin–Huxley (HH)}~\cite{Ref_Hodgkin_HHmodel_Physio52}, \textit{Leaky Integrate and Fire (LIF)}~\cite{Ref_Falez_PhDthesis_UnivLille19}, \textit{Izhikevich Model}~\cite{Ref_Izhikevich_WhichNeuron_TNN04}, and \textit{Adaptive Exponential Integrate-and-Fire (AdExIF)}~\cite{Ref_Brette_AdExIF_Neurophys05}.  
Neuron model selection typically considers the expected neuronal dynamics and computational complexity~\cite{Ref_Gerstner_NeuronalDynamics_Cambridge14, yamazaki2022spiking}.
The illustration of neuronal dynamics and computational complexity trade-off for different neuron models is shown in Fig.~\ref{Fig_NeuronModels}.
The output spikes are generated only when neurons’ membrane potential reaches the threshold, and transmitted through synapses, enabling ultra-low power/energy consumption~\cite{bib45, Ref_Koppula_EDEN_MICRO19, Ref_Putra_ReSpawn_ICCAD21, Ref_Putra_SparkXD_DAC21, Ref_Putra_EnforceSNN_FNINS22}.

Note, spike-based operations can also be applied to emerging network models like Transformers, i.e., so-called \textit{Spike-based Transformers}, by employing spike data representation and spiking neuron model in Transformer network architectures, as demonstrated in recent research works~\cite{Ref_Zhang_SpikeTrans_CVPR22, Ref_Zhou_Spikformer_arXiv22, Ref_Yao_SpikeTrans_NeurIPS24, Ref_Yao_Ref_Yao_SpikeTrans2_arXiv24, Putra_QSViT_arXiv25}.
These works show that Spike-based Transformers have comparable scalability and performance over the conventional Transformers, while enabling higher energy efficiency when executed on neuromorphic hardware due to their sparse spike-based operations.

\textbf{SNN Training:}
The training process can be performed using \textit{bio-plausible} or \textit{analytical learning rules}. 
The bio-plausible ones are typically characterized by local learning mechanism, like \textit{Hebbian}~\cite{Ref_Ruf_Hebbian_IWANN97}, \textit{Spike-Timing-Dependent Plasticity (STDP)}~\cite{Ref_Bi_STDP_Neuroscience98}, \textit{Spike-Driven Synaptic Plasticity (SDSP)}~\cite{Ref_Fusi_SDSP_NeCo00, Ref_Brader_SDSP_NeCo07}, and \textit{Reward-Modulated STDP (R-STDP)}~\cite{Ref_Mozafari_RSTDP_Patrec19, bib339}. 
Meanwhile, the analytical ones encompass \textit{DNN-to-SNN conversion}~\cite{Ref_Rueckauer_DNNs2SNNs_FNINS17}, \textit{Surrogate Gradient Learning}~\cite{Ref_Neftci_SurrogateSNNs_MSP19, Putra_SpiKernel_ESL24, Putra_SpikeNAS_arXiv24} (e.g., \textit{Backpropagation Through Time (BPTT)}~\cite{Ref_Neftci_SurrogateSNNs_MSP19, bib302, Ref_Shrestha_SLAYER_NeurIPS18}, \textit{Spatio-Temporal Backpropagation (STBP)}~\cite{Ref_Wu_STBP_FNINS18}, and \textit{Deep Continuous Local Learning (DECOLLE)}~\cite{Ref_Kaiser_DECOLLE_FNINS20}), \textit{Spike-Triggered Local Representation Alignment (ST-LRA)} ~\cite{bib342}, and \textit{Bayesian Learning}~\cite{bib299}. 
To facilitate SNN training, software frameworks play a vital role. Some notable ones include \textit{SpikingJelly}~\cite{Ref_Fang_SpikingJelly_Science23}, \textit{BindsNet}~\cite{hazan2018bindsnet}, \textit{SNNtorch}~\cite{eshraghian2023training}, TinySpiking~\cite{liu2025tinyspiking}, Norse~\cite{pehle2021norse}, SpyTorch~\cite{spytorch}, SINABS~\cite{sinabs}, SpykeTorch~\cite{bib341}, Brian \cite{bib303}, Lava~\cite{bib322}, BrainCog~\cite{bib309}, PySNN~\cite{pysnn}, and PyNCS~\cite{stefanini2014pyncs}. These frameworks are still evolving to support CL features. Meanwhile, standard DL frameworks such as \textit{Tensorflow}~\cite{abadi2016tensorflow} and \textit{PyTorch}~\cite{paszke2019pytorch}, and CL frameworks such as \textit{PyCIL}~\cite{zhou2023pycil}, \textit{FACIL}~\cite{bib39} and \textit{Avalanche}~\cite{JMLR:v24:23-0130} stand out for their user-friendly interfaces, flexibility, and support for complex applications. Table \ref{Frameworks} addresses the key question Q5 by comparing neuromorphic, DL and CL frameworks, outlining that the SNN frameworks with local STDP learning and neuromorphic hardware are more computationally efficient. 

\begin{table*}[h]
\centering
\footnotesize
\caption{Comparison of neuromorphic, standard DL and CL frameworks.}
\begin{tabular}{p{2cm}p{1.8cm}p{1.9cm}p{1.1cm}p{1.2cm}p{1.4cm}p{1.3cm}p{1.3cm}p{1.5cm}}
\cmidrule{1-9}%
\textbf{Frameworks} & \multicolumn{3}{c}{\textbf{Learning Mechanism}} & \textbf{CL Support} & \textbf{Hardware Support} & \textbf{Flexibility and Usability} & \textbf{PyTorch-based} &  \textbf{Computational Efficiency} \\
\cmidrule{2-4}
& \textbf{Unsupervised} & \textbf{Supervised} & \textbf{Hybrid}\\
\cmidrule{1-9}%
& & & \multicolumn{3}{c}{\textbf{Neuromorphic Frameworks}} \\
\cmidrule{1-9}%
SpikingJelly~\cite{Ref_Fang_SpikingJelly_Science23} & STDP, Hebbian Learning & BPTT, SG & RL & \checkmark & CPU, GPU, Neuromorphic chips & High & \checkmark & High\\
BindsNet~\cite{hazan2018bindsnet} & STDP, Hebbian Learning & $\times$ & RL & $\times$ & CPU, GPU & Medium & \checkmark & Moderate\\
SNNtorch~\cite{eshraghian2023training} & STDP & BPTT, SG & $\times$ & $\times$ & CPU, GPU &	High & \checkmark & High\\
TinySpiking~\cite{liu2025tinyspiking} & STDP & $\times$ & $\times$ & $\times$ &  CPU, MCU & High & Python-Based & High\\
Norse~\cite{pehle2021norse} & STDP, Hebbian Learning & BPTT, SG & RL & $\times$ & CPU, GPU & High & \checkmark & High \\
SpyTorch~\cite{spytorch} & STDP & $\times$ & $\times$ & $\times$ & CPU, GPU & Medium & \checkmark & Moderate\\
SINABS~\cite{sinabs} & STDP & BP & $\times$ & $\times$ & Neuromorphic chips & Medium & \checkmark & High\\
SpykeTorch~\cite{bib341} & STDP, R-STDP & $\times$ & $\times$ & $\times$ & CPU, GPU & Medium & Python-Based & Moderate \\
Brian\cite{bib303} & STDP, Hebbian Learning & $\times$ & $\times$ & $\times$ & CPU & Medium & Python-Based & Moderate \\
Lava~\cite{bib322} & STDP, Hebbian Learning & $\times$ & RL & $\times$ & Neuromorphic chips & Medium & Python-Based & High\\
BrainCog~\cite{bib309} & STDP, Hebbian Learning & BPTT, SG & RL & \checkmark & CPU, GPU, Neuromorphic chips &	High & \checkmark & High\\
PySNN\cite{pysnn} & STDP & SG & $\times$ & $\times$ & CPU, GPU & Medium & \checkmark & High \\
PyNCS~\cite{stefanini2014pyncs} & STDP, Hebbian Learning & $\times$ & $\times$ &  $\times$ & Neuromorphic chips & Medium & Python-Based & Moderate \\
\cmidrule{1-9}%
& & & \multicolumn{3}{c}{\textbf{Standard DL Frameworks}} \\
\cmidrule{1-9}%
TensorFlow~\cite{abadi2016tensorflow} & $\times$ & BP & RL & \checkmark & CPU, GPU, TPU & Very High & Python-Based & Relatively low\\
PyTorch~\cite{paszke2019pytorch} & $\times$ & BP & RL & \checkmark & CPU, GPU & Very High & \checkmark & Relatively low\\
\cmidrule{1-9}%
& & & \multicolumn{3}{c}{\textbf{CL Frameworks}} \\
\cmidrule{1-9}%
PyCIL~\cite{zhou2023pycil} &$\times$& BP & RL & \checkmark & CPU, GPU & High & \checkmark & Relatively low\\
FACIL\cite{bib39} & $\times$ & BP, SG & $\times$ & \checkmark & CPU, GPU & High & \checkmark & Relatively low\\
Avalanche~\cite{JMLR:v24:23-0130} & $\times$ & BP & RL & \checkmark & CPU, GPU & High & \checkmark & Relatively low\\
\cmidrule{1-9}%
\end{tabular}
\label{Frameworks} 
\end{table*}

\begin{figure}[h]
    \centering
    \includegraphics[width=0.48\textwidth]{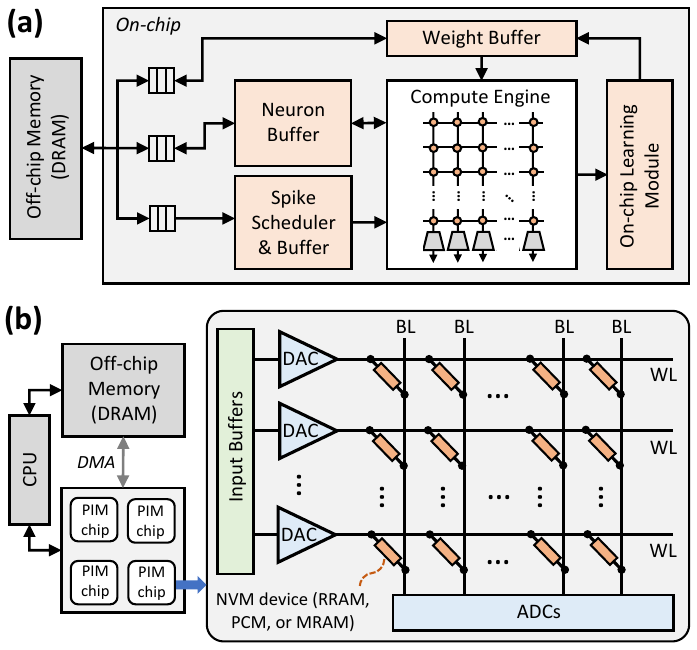} 
    \caption{Neuromorphic processors: (a) CMOS-based HW architecture, and (b) NVM-based HW architecture; adapted from studies in~\cite{Ref_Putra_EmbodiedNAI_ICARCV24}.} 
    \label{Fig_HWplatforms}
\end{figure}

\textbf{Neuromorphic Hardware:}
SNN processing demands a suitable computing hardware to maximize its potentials in accuracy, latency, and power/energy efficiency. 
Conventional von-Neumann architecture-based hardware platforms (e.g., CPU and GPUs) have been widely used to perform Python-based SNN processing using generic arithmetic units ~\cite{Ref_Stimberg_Brian2_eLife19, hazan2018bindsnet, Ref_Fang_SpikingJelly_Science23,Ref_Putra_EmbodiedNAI_ICARCV24}, thereby leading to sub-optimal efficiency gains. 
To address this, CMOS-based neuromorphic hardware accelerators facilitate efficient spike transmission and computation~\cite{Ref_Basu_SNNchips_CICC22}, allowing their implementation in the Field-Programmable Gate Array (FPGA) or Application-Specific Integrated Circuit (ASIC). 
Popular accelerators include \textit{Neurogrid}~\cite{Ref_Neurogrid_JPROC14}, \textit{ROLLS}~\cite{Ref_Qiao_ROLLS_FNINS15}, \textit{TrueNorth}~\cite{Ref_Akopyan_TrueNorth_TCAD15}, \textit{Loihi}~\cite{Ref_Davies_Loihi_MM18}, \textit{DYNAP}~\cite{Ref_SynSense_DYNAP}, and \textit{Akida}~\cite{Ref_BrainChip_Akida}\cite{Ref_Posey_Akida}. 
Their typical hardware architecture is shown in Fig.~\ref{Fig_HWplatforms}(a).
Beyond CMOS-based technologies, the \textit{processing-in-memory (PIM)} or \textit{compute-in-memory (CIM)} paradigm has been explored to further reduce latency and energy in data transfer between memory and compute units~\cite{Ref_Moitra_SpikeSim_TCAD23, Ref_Putra_HASNAS_arXiv24} by leveraging \textit{non-volatile memory (NVM)} technologies, such as \textit{Resistive Random Access Memory (RRAM)}, \textit{Magnetic RAM (MRAM)}, and \textit{Phase Change Memory (PCM)}~\cite{Ref_Asifuzzaman_SurveyPIM_Memori23}.
Their typical hardware architecture is shown in Fig.~\ref{Fig_HWplatforms}(b).

%%%%%%%%%%%%%%%%%%%%%%%%%%%%%%%%%%%%%%
\subsection{Framework, Datasets, Benchmarks, and Evaluation Metrics} 
\label{ncl_dbm}

The discussion in this section addresses the key questions Q4 and Q8. 

\textbf{Framework:}
Recently, the SNN community has been building a framework for benchmarking neuromorphic algorithms and systems, so-called \textit{NeuroBench}~\cite{Ref_Yik_NeuroBench_Nature25}, encompassing datasets and metrics. 
Its algorithm-level metrics include correctness (e.g., mean average precision (mAP)) and complexity (e.g., memory footprint, synaptic operations, as well as model and activation sparsity).
Meanwhile, its system-level metrics include task-specific correctness, timing, and efficiency.
Despite these advancements, NeuroBench framework is still under continuous development, and has not provided comprehensive benchmarks for NCL considering different CL settings and scenarios (see Section~\ref{CLsettings} and Section~\ref{CLscenarios}). 
Therefore, \textit{a standardized framework for comprehensively benchmarking NCL is still missing, and requires extensive developments}.
Toward this, we propose to consider the following additional datasets, benchmarks, and evaluation metrics.

\textbf{Datasets \& Benchmarks:} 
Neuromorphic datasets play a crucial role in advancing the NC field, particularly in the training and evaluation of SNN models.
These datasets are derived or generated to leverage the unique characteristics of event-based data. 
Some datasets are obtained using neuromorphic sensors, and some others are derived from existing conventional datasets.
Neuromorphic datasets can be divided into three categories based on their acquisition methods, as the following.
\begin{enumerate}
    \item \textit{Real-World Scene Datasets}: 
    They are collected directly from real-world environments using event-based sensors, such as Dynamic Vision Sensor (DVS) cameras. 
    Consequently, these datasets are unlabeled until the labeling process is performed. 
    For instance, the DVS-128 Gesture dataset~\cite{Ref_Amir_DVSGesture_CVPR17} for gesture recognition is obtained using a DVS camera.
    \item \textit{Transformation Datasets}: 
    They are obtained by transforming the existing (labeled) datasets from the ANN domain through recordings with event-based sensors. 
    An example for this category is the CIFAR10-DVS dataset~\cite{li2017cifar10}.
    These datasets are expected to leverage event-based data characteristics for SNNs, while having direct counterparts in the ANN domain for comparison, thereby making them popular for SNN evaluation.
    \item \textit{Algorithmically Generated Datasets}: 
    They are synthesized using algorithms that emulate the behavior of event-based sensors. 
    They convert existing labeled image, video, or speech data into neuromorphic datasets using difference-based algorithms. 
    An example for this category is the Spike-TIDIGITS~\cite{pan2020efficient}. 
\end{enumerate}

Table~\ref{datasets} provides an elaborate comparison of the prominent neuromorphic datasets, including their publication year, characteristics, and tasks. 
Currently, they also serve as benchmarks for evaluating and comparing SNN models, including the NCL methods.
However, these datasets are mainly for classification tasks, which limit the applicability of neuromorphic systems. 
Therefore, more diverse datasets are required including the types of tasks (e.g., regression and generative), diversity in conditions during data collection (e.g., weather, lighting, fog, etc.), and even diversity of application use-cases (e.g., vision, sound, tactile, and olfactory datasets).
Moreover, benchmarks for NCL should also consider different possible CL settings (e.g., supervised, unsupervised, and RL) and scenarios (e.g., task-IL, domain-IL, and class-IL).

\textbf{Evaluation Metrics:}
Section~\ref{eval_metrics} highlights the standard evaluation metrics including average accuracy, BWT and FWT, forgetting measure, model size, sample size, compute resources (e.g., FLOPS and time complexity), and power/energy efficiency.
Therefore, they cover limited metrics for evaluating the functionality of different CL tasks. 
Moreover, the conventional metrics serve as indicators of a model’s ability to retain prior knowledge while learning new tasks sequentially. However, when applied to neuromorphic systems, these metrics must be reinterpreted and complemented due to the distinct computational paradigms inherent in such architectures, such as event-driven computation, quantized activations, low-precision operations, and temporal spike-based encoding.
To address this, we reference works that propose complementary performance indicators and suggest
new metrics as follows.
\begin{itemize}
    \item \textit{Performance:} 
    In spike-based systems, processing time often relies on the neural coding (e.g., rate, TTFS, etc.). 
    Therefore, \textit{latency} (measured in time steps until a decision is made) and \textit{throughput} becomes a critical indicator of performance of low-power or real-time applications~\cite{goltz2021fast}.
    \item \textit{Efficiency:} 
    \textit{Power and energy consumption} during CL training and inference are important to indicate the energy efficiency of NCL.
    For estimation, \textit{spike count} and \textit{membrane potential dynamics} may also provide crucial insights into the energy efficiency~[42].
    Furthermore, \textit{utilization} of memory bandwidth and compute module, as well as \textit{synaptic operation rates} (SOPS) are also relevant for indicating efficiency when running networks on neuromorphic chips. 
    NCL systems that maintain accuracy with high efficiency are preferred for embedded AI deployments. 
    \item \textit{Robustness:} 
    Neuromorphic systems may exhibit variability due to low-bit precision, noisy spikes, and device-level process variations. 
    Therefore, robustness (defined as accuracy maintained under noise, perturbations, or quantization) emerges as a meaningful metric besides accuracy.
    \item \textit{Task-specific functionality metrics:} 
    They depend on the type of task. 
    For instance, mean absolute error (MAE) and mean square error (MSE) can be used for regression-based tasks.
    \item \textit{Memory ratio:} It represents how much memory overhead compared to the original model size, that is needed to execute the implemented CL method. 
    \item \textit{Adaptability scores:} 
    They represent how effective is the implemented CL method over time considering different CL scenarios i.e., task-IL, class-IL, domain-IL, TA etc.  
\end{itemize}

\begin{table*}
\centering
\footnotesize
\caption{Comparison of widely used neuromorphic datasets for SNNs with their publication year, description, key features and tasks.}
\begin{tabular*}{\textwidth}{@{\extracolsep\fill} lllll}
\cmidrule{1-5}%
\textbf{Dataset} & \textbf{Year} & \textbf{Description} & \textbf{Key Features} & \textbf{Tasks} \\
\cmidrule{1-5}%
N-MNIST	& 2015 & An event-based version of the MNIST & Classes: 10 & Digit Classification \\
\cite{bib310} & & dataset captured using Asynchronous & Resolution: 28x28  &  \\ 
& & Time-based Image Sensor (ATIS) & Training Samples: 60,000 &  \\ 
& & with each digit recorded over 300 ms & Testing Samples: 10,000 & \\
\cmidrule{1-5}%
DVS-128 Gesture & 2017 & An event-driven data of hand and arm & Classes: 11 & Gesture Recognition \\ 
\cite{Ref_Amir_DVSGesture_CVPR17} & & gestures captured using DVS128 camers, & Subjects: 29  &
\\ 
& & focusing on dynamic movement with & Illumination conditions: 3 & \\ % 
& & 6s/sample duration & Resolution: 128x128 & \\ % 
\cmidrule{1-5}%
CIFAR10-DVS & 2017 & An event-based version of CIFAR-10 & Classes: 10 \& Event streams: 10,000& Image Classification \\ % 
\cite{li2017cifar10} & & dataset with 300ms/sample duration & Resolution: 128x128 & \\ % 
\cmidrule{1-5}%
N-Caltech101 & 2015 & A spiking version of the original & Samples: 8709 & Object Recognition \\
\cite{bib310} & & frame-based Caltech101 dataset & Classes: 100 objects, and & \\ % 
& & with 300–500ms/sample duration & a background class\\
\cmidrule{1-5}%
N-CARS & 2018 & A large real-world event-based dataset & Classes: 2 & Car \& Background \\
\cite{Ref_Sironi_HATS_CVPR18} & & captured using ATIS camera with &  Samples of Car: 12,336 \& Background: 11,693 & Recognition \\ % 
& & 100ms/sample duration & Training Samples: 7940 (car), 7482 (bg) & \\
& & & Testing Samples: 4396 (car), 4211 (bg) &  \\
\cmidrule{1-5}%
N-Omniglot & 2022 & A large-scale neuromorphic dataset & Classes: 1,623 & Few-Shot Learning\\
\cite{li2022n} & & of handwritten characters, representing & Samples per class: 20 & in SNNs \\
& & 50 different languages captured by DVS & Samples: 32,460 &  \\ 
\cmidrule{1-5}%
Spiking Heidelberg & 2022 & An audio-based dataset of 10k high- & Classes: 10 & Speech
Classification\\
Digits (SHD) \cite{bib312} & & quality recordings of spoken digits ranging &  Digit Count: 10,420 & and Keyword Spotting \\
 & & from zero to nine in English and German & Speakers: 12 \\
\cmidrule{1-5}%
ES-ImageNet & 2021 & A large-scale event-stream dataset & Classes: 1000 & Image Classification\\
\cite{Ref_lin2021imagenet} & & converted from the ImageNet dataset \cite{russakovsky2015imagenet} & Training Samples: 1,257,035 & \\
& & using a software-based event generation & Testing Samples: 49,881 \\
& & Omnidirectional Discrete Gradient (ODG)  & Total Samples: 1.3 M \\
& & algorithm with 29.47ms/sample duration & Resolution: 224×224*\\
\cmidrule{1-5}%
SpikeBALL & 2023 & A neuromorphic dataset capturing 10 & Class: 1 & Object Tracking\\
\cite{guerrero2023spikeball} & & different trajectories of a ball in a table & Events: 80,000 to 100,000\\
& & football game & per trajectory\\
\cmidrule{1-5}%
NE15-MNIST & 2016 & A spiking version of MNIST dataset & Classes: 10 & Visual Recognition \\
\cite{liu2016benchmarking} & & with four sub-datasets encoded using: & Training Samples: 60,000 & \\
& & Poissonian, Rank-Order, DVS recorded & Testing Samples: 10,000 \\
& & flashing and DVS recorded moving \\
\cmidrule{1-5}%
ASL-DVS & 2019 & Event-driven data of 24 letters (A-Y, excl. J) & Classes: 24 & Sign Language \\
\cite{bi2019graph} & & from American sign language recorded & Samples: 100,800 / Samples per-class: 4,200 & Recognition  \\
& &  using DAVIS240c NVS camera & Training Samples: 80,640 \\
& &  with 100ms/sample duration & Testing Samples: 20,160 \\
\cmidrule{1-5}%
N-TIDIGITS18 & 2018 & The spike recordings from a Dynamic & Classes: 11 & Spoken Digit \\
\cite{anumula2018feature} & & Audio Sensor (DAS) in response to the & Training Samples: 8,623 & Recognition \cite{park2023advancing}\\
& & TIDIGITS audio dataset &  Testing Samples: 8,700 \\
\cmidrule{1-5}%
Spike-TIDIGITS & 2020 & A Spike-based version of TIDIGITS dataset & Classes: 11 & Speech Recognition \\
\cite{pan2020efficient} & & created using the Biologically Plausible & Training Samples: 2464 & \\
& & Auditory Encoding (BAE) algorithm & Testing Samples: 2486 \\
\cmidrule{1-5}%
Spike-TIMIT \cite{pan2020efficient} & 2020 & An event-based version of TIMIT & Training Samples: 4621 & Audio Classification \\
 & & dataset \cite{garofolo1993timit}  & Testing Samples:
1679 & \\
\cmidrule{1-5}%
DVS Lane Extraction & 2019 & A high-resolution event-based dataset & Classes: 5, (0 for bg and & Lane Extraction \\
(DET) \cite{cheng2019det} & &  of complex traffic scenes and various lane & 1,2,3,4 for four lane types) & \\
& & types annotated with multi-class & Images: 5,424\\
& & segmentation captured by CeleX-V DVS &  Resolution: 1280×800\\
\cmidrule{1-5}%
KUL-UAVSAFE~\cite{safa2021fail} & 2021 & A joint DVS and RGB dataset in an indoor & Subjects: 6 \& Variations: Clothes color, &  People Detection\\
 & & environment with one walking human &  shape \& walking style \\
\cmidrule{1-5}%
\end{tabular*}
\label{datasets} 
\begin{tablenotes} \footnotesize
\item \textit{*}: Events are generated within a 256x256-pixel range, but only those in the central 224x224-pixel region are meaningful, the rest are noise caused by edge motion; \textit{bg}: background.
\end{tablenotes}
\vspace{-0.2cm}
\end{table*}

%%%%%%%%%%%%%%%%%%%%%%%%%%%%%%%%%%%%%%
\subsection{Neuromorphic Continual Learning (NCL)}
\label{sec3.2}

The integration of CL with neuromorphic computing (i.e., NCL) is the central focus of the paper.
Specifically, how NCL facilitates continual adaptation in diverse operational environments by leveraging key characteristics of neuromorphic systems that align with CL requirements, as highlighted in the following.
\begin{itemize}
    \item \textit{Neural plasticity:} 
     It is established using bio-plausible learning rules, such as Hebbian rule and Spike-Timing Dependent Plasticity (STDP), which enable SNNs to flexibly learn new information while preserving prior knowledge (e.g., through weight potentiation and depression), which is useful for addressing CF problem. 
    \item \textit{Localized learning in each synapse:} 
    Bio-plausible learning rules in SNNs can leverage spiking activities for updating weight locally in each synapse. 
    It is useful for learning information without explicit labels, which is required for enabling online training in updating systems' knowledge and adapting to dynamic environments. 
    \item \textit{Event-based processing:} 
    The event-driven operations in SNNs consider both spatial and temporal information in the event-based data streams, which aligns with the nature of dynamic data streams in CL. 
    This potentially streamlines the execution of CL algorithms.
    \item \textit{Energy-efficient computation:} 
    The event-driven operations in SNNs and neuromorphic hardware enable ultra-low power/energy computation, which is required for enabling CL execution in tightly-constrained systems.
\end{itemize}

\begin{figure*}[t]
    \centering
    \includegraphics[width=\textwidth]{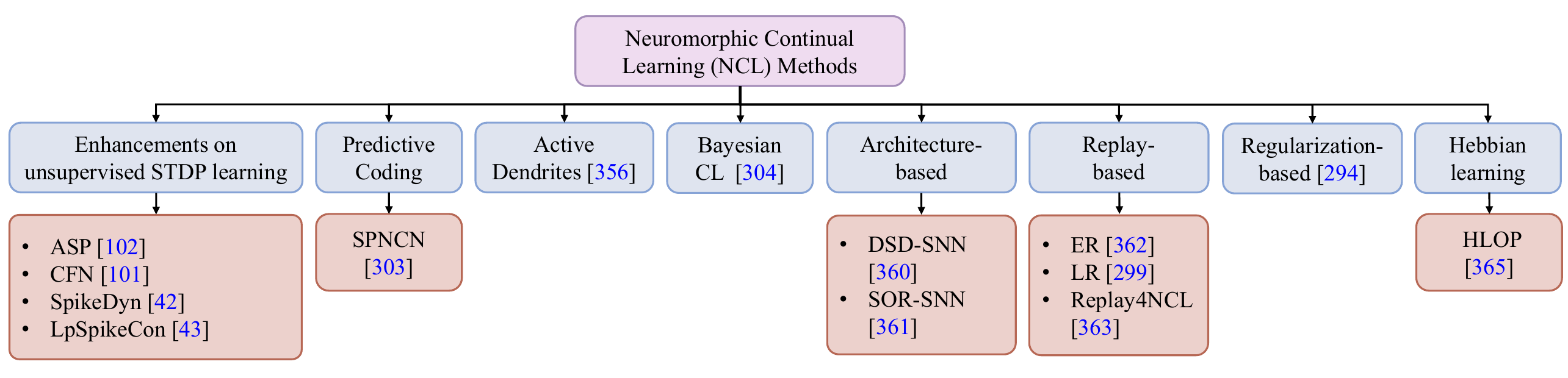} 
    \caption{A state-of-the-art taxonomy of NCL methods covered in this survey. We have highlighted the main categories (blue blocks), with each of their works shown (red blocks).}
    \label{NCL_taxonomy}
\end{figure*}

%%%%%%%%%%%%%%%%%%%%%%%%%%%%%%%%%%%%%%
\subsection{NCL Methods}

In this sub-section, we provide an in-depth review of state-of-the-art NCL methods, including their key ideas and techniques, whose details are summarized in Table~\ref{table_8}. Fig.~\ref{NCL_taxonomy} presents the taxonomy that systematically categorizes these methods based on their specific techniques.
This discussion also addresses the key question Q6.
%

%%%%%%------
\subsubsection{Enhancements on Unsupervised STDP Learning} 
\label{sec3.2.1}

While conventional STDP enables efficient online unsupervised learning, employing STDP alone may still suffer from CF~\cite{bib296, bib297}. 
To address these challenges, several techniques have been proposed in the literature, as discussed in the following.
\smallskip
\\
\textbf{Weight Decay:}
It prevents overfitting in NNs by penalizing large weight values by adding a regularization term, which encourages the model to keep the weights small. 
For instance, Adaptive Synaptic Plasticity (ASP)~\cite{bib108}, SpikeDyn~\cite{bib296} and lpSpikeCon~\cite{bib297} leak the weights to gradually remove old and insignificant information and ensure the weights do not grow too large. 
\smallskip
\\    
\textbf{Adaptive Threshold Potential:}
It refers to the dynamic adjustment of the neurons' threshold based on its spiking activity. 
It allows neurons to adapt their sensitivity to incoming spikes, making the network more flexible/robust in processing information. 
A neuron typically fires when its membrane potential exceeds a pre-defined threshold.
However, in an adaptive threshold model (as employed by SpikeDyn~\cite{bib296} and lpSpikeCon~\cite{bib297}), the threshold changes based on the neurons' recent activity, thus helping the neuron provide the following features.
\begin{enumerate}
    \item \textit{Regulating the firing rates} to prevent excessive firing and maintain a stable firing rate over time to avoid overactive neurons (\textit{homeostasis})~\cite{Ref_Diehl_SNN_FNCOM15, Ref_Putra_FSpiNN_TCAD20}. 
    \item \textit{Leveraging the temporal information} to improve the neurons' capability to respond to varying input patterns.
    \item \textit{Improving the signal-to-noise ratio (SNR)}, by separating the significant input patterns from the noise.
\end{enumerate}
\textbf{Adaptive Learning Rate:}
It dynamically adjusts the learning rate in training to improve convergence. 
The learning rate controls how much the models' weights are updated. 
ASP~\cite{bib108}, SpikeDyn~\cite{bib296} and lpSpikeCon~\cite{bib297} employ adaptive learning rates to determine the potentiation and depression factors in the STDP-based learning based on the spiking activities.
By prioritizing the adjustment of highly active synapses, it enhances learning efficiency and reduces unnecessary energy expenditure on less active connections. 
This selective learning process mimics biological synaptic behavior.

These techniques are often employed together in NCL methods as discussed below.

\begin{figure}[h]
    \vspace{-0.2cm}
    \centering
    \includegraphics[width=0.5\textwidth]{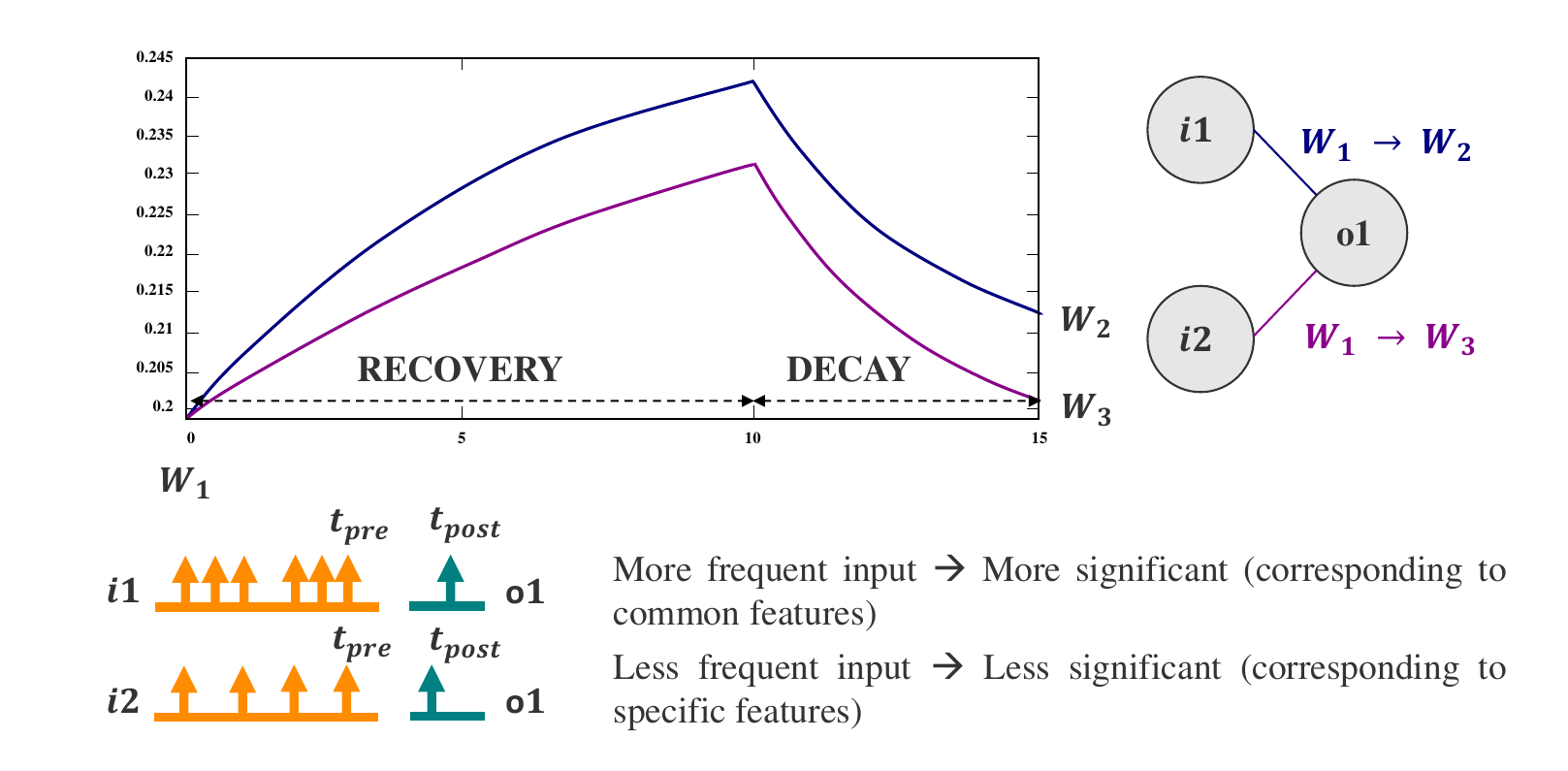} 
    \caption{Weight update process of ASP; adapted from studies in~\cite{bib108}. More frequent input spikes represent common features between old and new input patterns, and will experience a greater weight update compared to less frequent spikes, that correspond to unique features of a specific input.}
    \label{asp}
\end{figure}

\smallskip
\textbf{Adaptive Synaptic Plasticity (ASP)}~\cite{bib108}: 
It integrates weight decay with the STDP-based weight updates to balance forgetting and learning, while leveraging time-dependent learning rate. 
ASP uses a two-phase weight update process (i.e., \textit{recovery} and \textit{decay}), allowing weight updates based on spiking activities and gradually leak toward a baseline value; see Fig.~\ref{asp}. 
The SNN was implemented in Brian~\cite{bib303}, (a lightweight SNN simulator for rapid prototyping and experimentation with a flexible Python interface), to perform dynamic digit recognition with the MNIST dataset. The digit categories '0' to '9' were presented sequentially, without intermixing at any point during the training phase (i.e., no data reinforcement). For a 6400 neuron SNN, ASP achieved avg. accuracy of 94.2\% outperforming the standard STDP. A summary is presented in Table~\ref{table_8} and the empirical results are shown in Table~\ref{comparison-SSN-DNN} and Table~\ref{table11}, providing a comparative analysis in terms of network architecture, size and key performance metrics such as accuracy, memory footprint, and power/energy consumption of the reviewed works.
Despite its strengths, ASP requires larger quantities of input samples from earlier distributions than later ones, hence requiring the knowledge of task changes~\cite{bib72}, making it unsuitable for OCL scenario. 

\begin{figure}[t]
    \centering
    \includegraphics[width=0.45\textwidth]{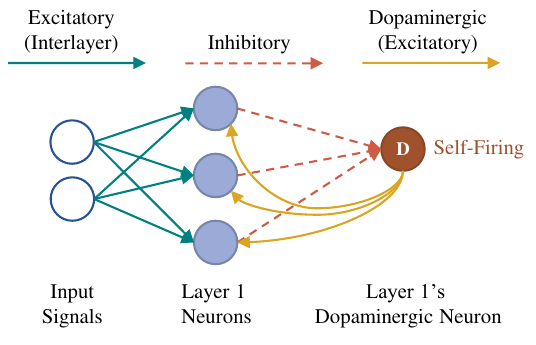} 
    \caption{Single-layer CFN architecture; adapted from studies in~\cite{bib72}. The dopaminergic neuron activates when other neurons in its layer are inactive, indicating new information in the input. This activation temporarily enhances plasticity in other neurons. Dopaminergic signals are weighted to provide targeted stimulation.}
    \label{CFN}
\end{figure}

\smallskip
\textbf{Controlled Forgetting Network (CFN)}~\cite{bib72}: 
It is an SNN architecture that exploits dopaminergic neurons to modulate the synaptic plasticity. 
Its idea is to temporarily make the weights of some neurons more plastic (easier to change) and keep the weights of other neurons. 
The modulation is triggered by dopaminergic neurons, which fire when there are no or only a small number of incoming inhibitory spikes (indicating that new information is encountered), and then stimulate a temporary boost of learning rate for other neurons (Fig.~\ref{CFN}). For a 6,400 neuron CFN, the evaluations on MNIST dataset in a fully disjoint scenario, achieved on average 95.24\% classification accuracy across all digits (see summary in Table~\ref{table_8} and the comparative empirical analysis in Table~\ref{comparison-SSN-DNN} and Table~\ref{table11}).
Despite its benefits, CFN requires additional components (i.e., dopaminergic neurons and their connections) and considers a conventional CL scenario (i.e., temporally separated tasks), which makes it challenging for deployments in an environment of changing priorities, i.e., OCL scenario. 

\begin{figure}[h]
\vspace{-0.2cm}
\centering
\includegraphics[width=0.5\textwidth]{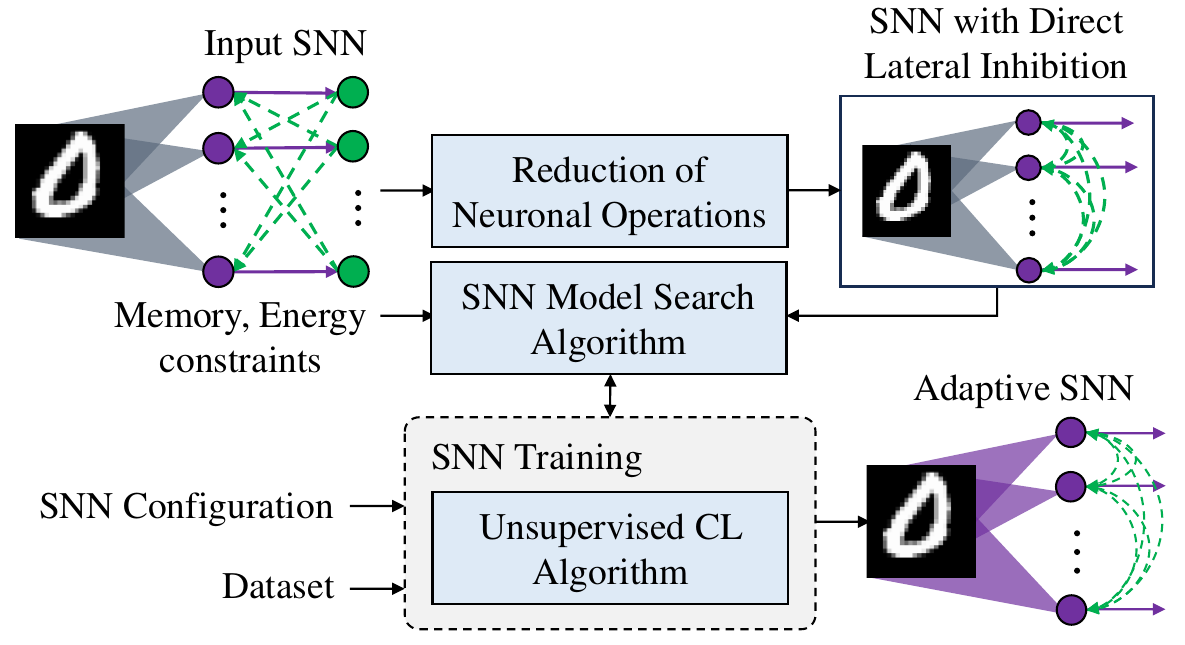} 
\caption{SpikeDyn framework for adaptive and energy-efficient unsupervised CL for SNNs; adapted from studies in~\cite{bib296}.}
\label{SpikeDyn}
\vspace{-0.2cm}
\end{figure}

\smallskip
\textbf{SpikeDyn}~\cite{bib296}: 
It focuses on enhancing STDP for enabling unsupervised CL while minimizing energy consumption in both training and inference phases (Fig.~\ref{SpikeDyn}). 
Its key ideas include: \textit{reduction of neuronal operations} by substituting the inhibitory neurons with direct lateral inhibitions to reduce memory and energy requirements; \textit{enhancing unsupervised CL algorithm} by employing adaptive learning rates, weight decay, adaptive threshold potential, and reduction of spurious weight updates; and \textit{SNN model size search} by leveraging analytical models to estimate the memory and energy requirements, and selecting a Pareto-optimal model that meets the resource constraints. The SNN was implemented using Bindsnet, a ML-oriented SNN library in Python~\cite{hazan2018bindsnet}. Evaluations on MNIST dataset in a dynamic environment (where tasks are fed sequentially, without re-feeding earlier ones, and each task contains an equal number of samples), for a 200 neuron SNN, showed avg. 23\% and 4\% improved accuracy than ASP~\cite{bib108} when learning a new task and when retaining the old task, respectively. It also reduced energy consumption by up to avg. 57\% and 51\% for training and inference, respectively (see summary in Table~\ref{table_8} and the comparative empirical analysis in Table~\ref{comparison-SSN-DNN} and Table~\ref{table11}).
SpikeDyn demonstrated that optimization techniques can be exploited to minimize memory and energy requirements of CL systems, but they should be supported with an enhanced unsupervised CL algorithm to maintain the performance. 

\begin{figure} [t]
    \centering
    \includegraphics[width=0.5\textwidth]{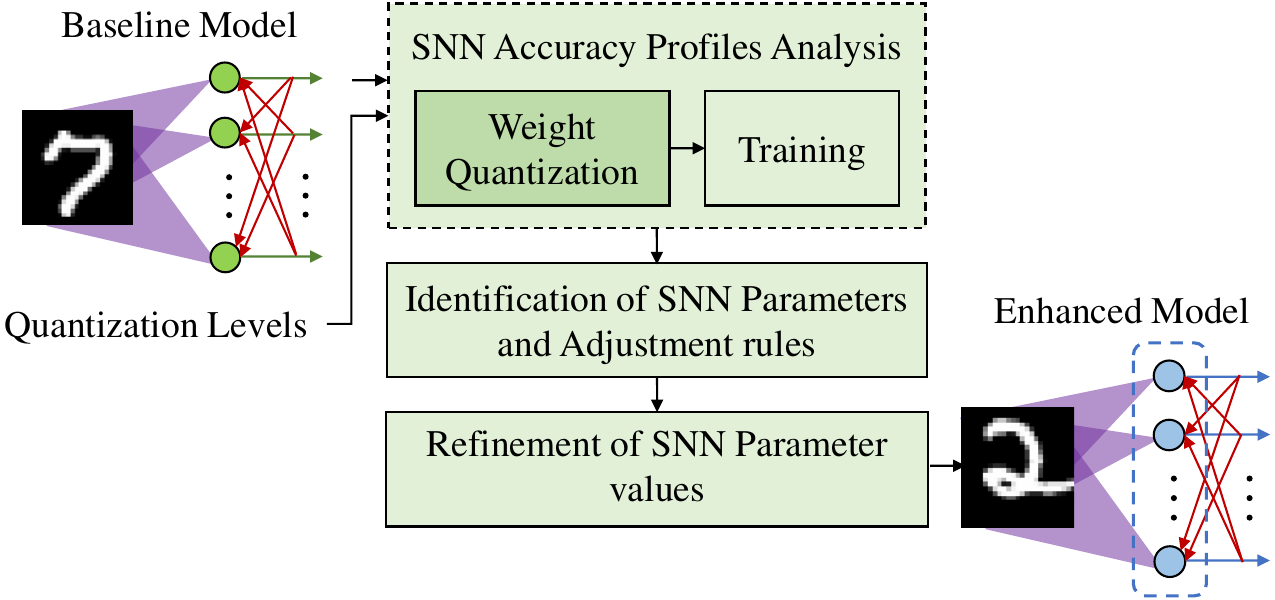} 
    \caption{Key steps of lpSpikeCon methodology for low-precision unsupervised CL for SNNs; adapted from studies in~\cite{bib297}.}
    \label{Lpspikecon}
\end{figure}

\smallskip
\textbf{lpSpikeCon}~\cite{bib297}: 
It extends the studies in SpikeDyn~\cite{bib296}, by enabling STDP-based unsupervised CL under low-precision settings for embedded AI systems (e.g., robots).  
lpSpikeCon employs weight quantization and compensates the loss of information by identifying and adjusting SNN parameters that significantly impact accuracy (Fig.~\ref{Lpspikecon}). The SNN was implemented using Bindsnet~\cite{hazan2018bindsnet}, see Table~\ref{Frameworks}. It performed a case study in a dynamic CL scenario using the MNIST dataset. The network was trained sequentially on digit classes from 0 to 9, receiving one class at a time. After each training phase, the model was evaluated on the test samples of all classes learned so far, simulating a real-time CL process. The 4-bit weight quantized SNN with key parameter adjustments showed no accuracy loss in the inference while reducing the weight memory by 8x compared to 32-bit non-quantized SNN (see summary in Table~\ref{table_8} and comparative empirical analysis in Table~\ref{comparison-SSN-DNN} and Table~\ref{table11}). 
This study found that the key SNN parameters to adjust include the adaptive threshold potential and weight decay rate. 
This study demonstrated that low-precision settings can be exploited for substantially reducing memory and energy requirements, and adjustments of other parameters are crucial to maintain the performance.
Despite their promising results, SpikeDyn and lpSpikeCon have so far focused on the MNIST dataset and conventional CL scenario. Expanding these methods to more complex scenarios for OCL is a potential future research direction.

\subsubsection{Predictive Coding}

\begin{figure}[h]
    \centering
    \includegraphics[width=0.5\textwidth]{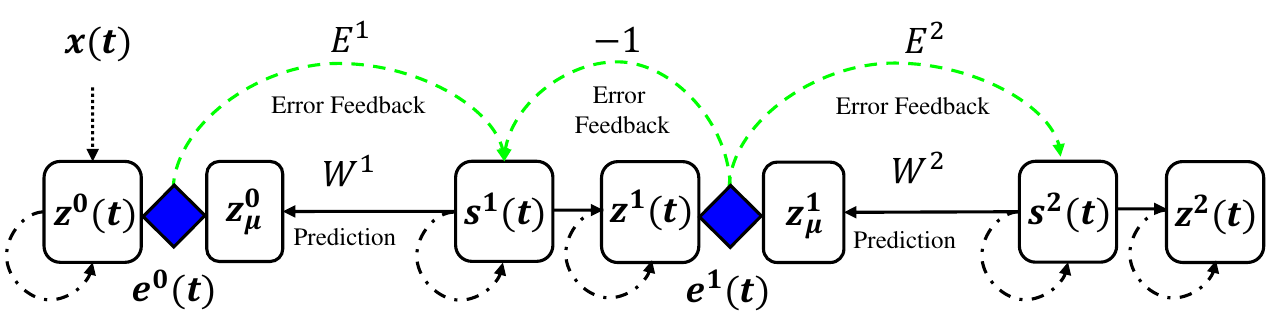} 
    \caption{A 2-layer SpNCN with error units (blue diamonds) that measure the difference between predictions ($z^0_{\mu}$, $z^1_{\mu}$) and target signals ($z^0(t), z^1(t)$). Variables $s^0(t), s^1(t)$ are the binary spike outputs of neuron groups at time $t$. Black dash-dotted arrows show the repeated transmission of the last known values. Mismatch signals, shown by green dashed arcs, adjust the spiking neuron's action potentials through synapses. Solid black lines show predictive synapses, and black dotted lines show direct information transfer; adapted from studies in~\cite{bib342}.}
    \label{spncn}
\end{figure}

This method is used by the \textbf{Spiking Neural Coding Network (SpNCN)}~\cite{bib342} to predict incoming data and then correct the prediction based on the actual inputs~\cite{bib335, bib345}; see Fig.~\ref{spncn}. 
This iterative process of \textit{``guess-and-check''} allows the network to adjust its weights continually and learn from data streams without repeated exposure to same data. 
Key ideas of SpNCN are as follows.

\begin{itemize}
    \item Prediction of neuron activity and error correction using local synaptic updates.
    \item Weight adaptation using a coordinated ST-LRA, which adjusts weights based on the mismatch between predictions and actual activity. 
    It is also combined with STDP as regularizer.
    \item A memory module and task/context-modulated lateral inhibition to enhance memory retention across tasks.
\end{itemize}
%. 

A case study was conducted using adapted versions of standard CL benchmarks, i.e., Split MNIST, Split NotMNIST, and Split Fashion MNIST (FMNIST), in a spike-train continuous-time setting. Each dataset was partitioned into five sequential tasks, where each task involved classifying two object categories (e.g., digit pairs in Split MNIST, letters A–J in NotMNIST, or clothing types in FMNIST). The tasks were presented in a fixed sequence, simulating a dynamic CL scenario where the learner must adapt to new tasks while retaining prior knowledge, despite changes in the label distribution that create cross-task interference.
Experimental results showed normalized accuracy of 0.9653 on MNIST, 0.9120 on Not-MNIST and 0.9995 on FMNIST (see summary in Table~\ref{table_8} and comparative empirical analysis in Table~\ref{comparison-SSN-DNN} and Table~\ref{table11}).
This study showed that combination of learning rules is potential for enabling CL.
However, this complex combination of multiple learning rules make it challenging for deployments under OCL scenario.

\subsubsection{Active Dendrites} 

\begin{figure}[h]
    \centering
    \includegraphics[width=0.5\textwidth]{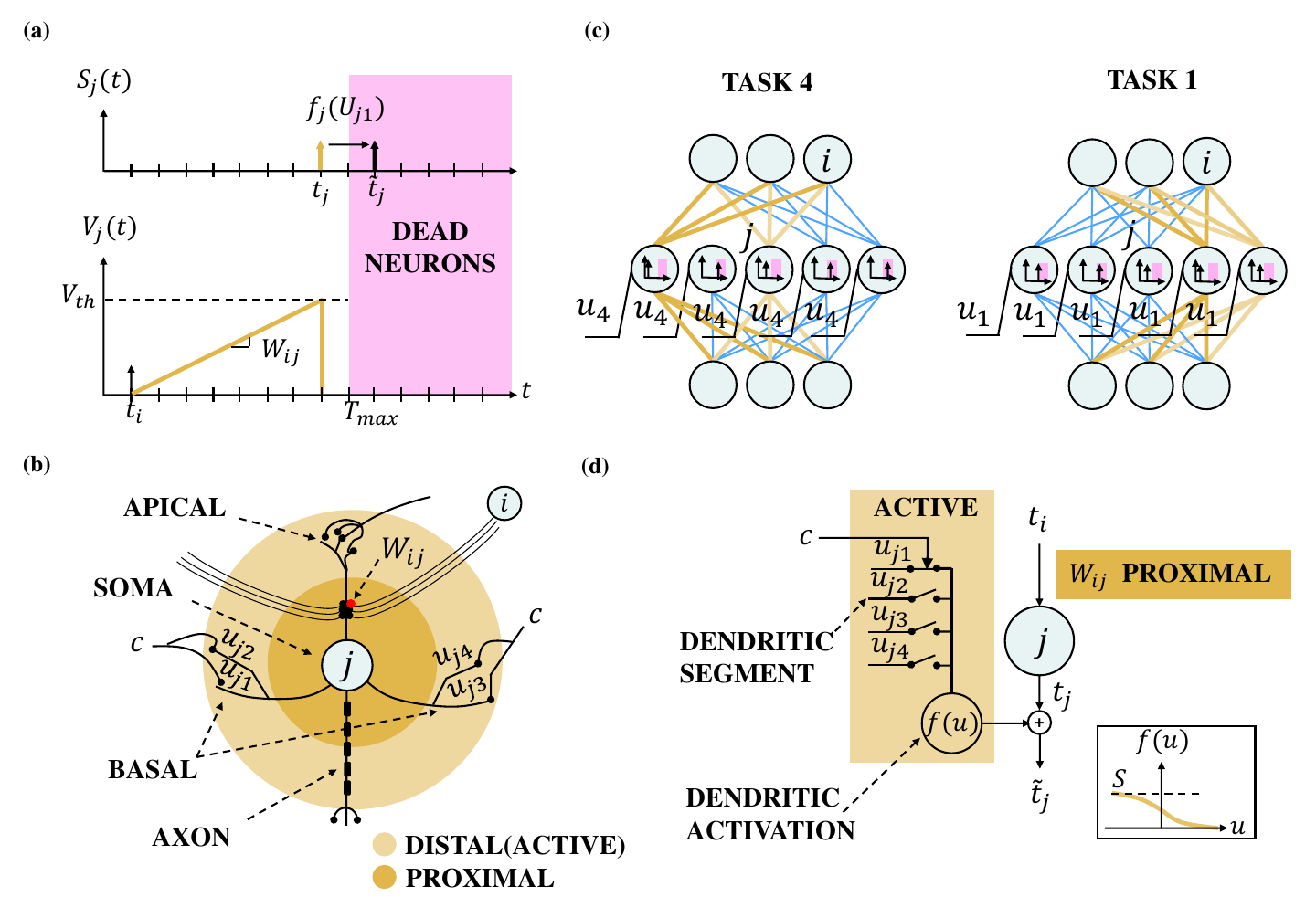} 
    \caption{Neuron model and network architecture with active dendrites; adapted from studies in~\cite{bib317}. (a) Linear integration of synaptic strength $W_{ij}$ following a pre-synaptic spike at $t_i$; Top figure shows the modulation of spike timing by dendritic processes. (b) Illustration of a pyramidal neuron. (c) Selection of different sub-networks for various tasks based on dendritic segment activity. (d) Proposed neuron model and dendritic activation function.}
    \label{AD}
\end{figure}

Studies in~\cite{bib317} proposed an SNN model leveraging \textit{active dendrites}~\cite{bib318} to facilitate task-specific sub-network formation (Fig.~\ref{AD}).
The spike time $t_j$ of a neuron is modulated by a function of the selected dendritic segment $u_{jn}$ for the current task. 
The dendritic activation function $f(u)$ modulates the spike time dynamically, allowing the model to adapt its behavior based on the task context. 
It also leverages TTFS encoding to introduce a gating mechanism, that enables selection of sub-networks for various tasks.
Experimental results on the Split MNIST dataset for sequential CL tasks demonstrated an end-of-training accuracy of 88.3\%. Moreover,
the FPGA implementation matched the quantized software model with an average inference time of 37.3ms and an accuracy of 80.0\%, highlighting the  potential for applications in resource-constrained environments (see summary in Table~\ref{table_8} and comparative empirical analysis in Table~\ref{comparison-SSN-DNN} and Table~\ref{table11}).
However, this method requires labeled data or supervisory signals, hence making it challenging for deployments under the OCL scenario.

\subsubsection{Bayesian Continual Learning} 
\label{sec3.2.4}

\begin{figure}[t]
    \centering
    \includegraphics[width=0.5\textwidth]{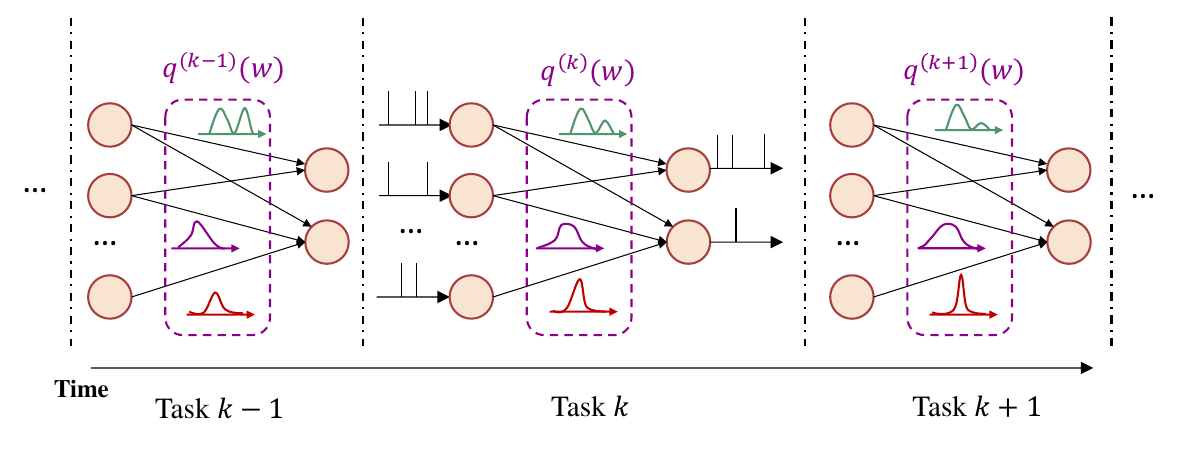} 
    \caption{In the Bayesian continual learning, the system is sequentially presented with similar yet distinct tasks; adapted from studies in~\cite{bib299}.}
    \label{BayCL}
\end{figure}

This method represents weights with parameters that quantify the epistemic uncertainty based on prior knowledge and observed data, and employs Bayesian methods for handling uncertainty over time by determining which knowledge to retain and which to forget~\cite{bib299} (Fig.~\ref{BayCL}).
For real-valued synapses, it uses a \textit{Gaussian} variational distribution to adjust the values; while for binary synapses, it uses a \textit{Bernoulli} variational distribution with \textit{Gumbel-softmax}~\cite{jang2016categorical}. The work was implemented using Intel's Lava platform~\cite{bib322}, enabling Bayesian CL in SNNs, see Table~\ref{Frameworks}.
Experimental results achieved avg. test accuracy of $85.44\pm0.16\%$ with 5× memory reduction on Split-MNIST and 74\% for MNIST-DVS (see summary in Table~\ref{table_8} and comparative empirical analysis in Table~\ref{comparison-SSN-DNN} and Table~\ref{table11}).
Although, this method provides better-calibrated decisions and better detection compared to conventional \textit{frequentist} approaches \cite{kreutzer2020natural}, the uncertainty quantification incurs high computational complexity, thereby making it challenging for deployments in OCL scenario. 

%%%%%%------
\subsubsection{Architecture-based Approach}

\begin{figure}[h]
    \centering \includegraphics[width=0.5\textwidth]{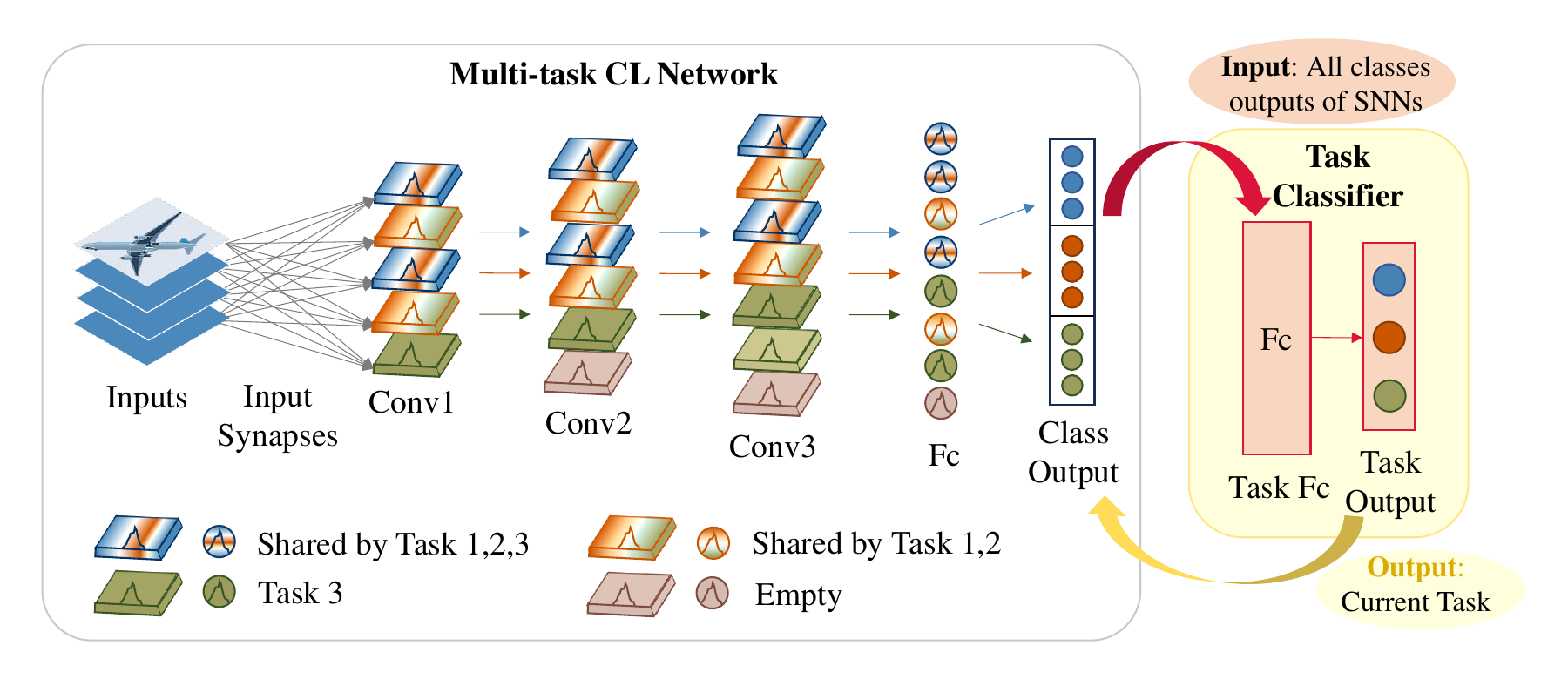} 
    \caption{Deep SNN architecture with dynamic structure development; adapted from studies in~\cite{bib304}.}
    \label{Dsdsnn}
\end{figure}

\textbf{Dynamic Structure Development of Spiking Neural Networks (DSD-SNN)} enhances the SNN structure by growing new neurons for new tasks and pruning redundant neurons~\cite{bib304} (Fig.~\ref{Dsdsnn}). 
It employs a deep SNN architecture comprising of multiple convolutional (CONV) and fully-connected (FC) layers, which is equipped with \textit{random growth}, \textit{adaptive pruning}, and \textit{freezing of neuron} mechanisms. This SNN was implemented using the Brain-inspired Cognitive intelligence
engine (BrainCog)~\cite{bib309}, see Table~\ref{Frameworks} for details.
Experimental results on the MNIST dataset in Task-IL scenario, demonstrated an accuracy
of $97.30\pm0.09\%$ with a network parameter compression rate of 34.38\%, outperforming the EWC, GEM and DEN DNN-based CL methods. Furthermore, an accuracy of $96.94\pm0.05\%$ was reported for the N-MNIST and $77.92\%\pm0.29$ in Task-IL and 60.47\% (10 steps) in Class-IL scenario for the CIFAR100 dataset (see summary in Table~\ref{table_8} and comparative empirical analysis in Table~\ref{comparison-SSN-DNN} and Table~\ref{table11}).
This study advanced multi-task learning while enhancing memory capacity and efficiency. 
However, the development of dynamic structure adds complexity to SNN implementation, which is challenging for OCL scenario.

\begin{figure}[t]
    \centering
    \includegraphics[width=0.5\textwidth]{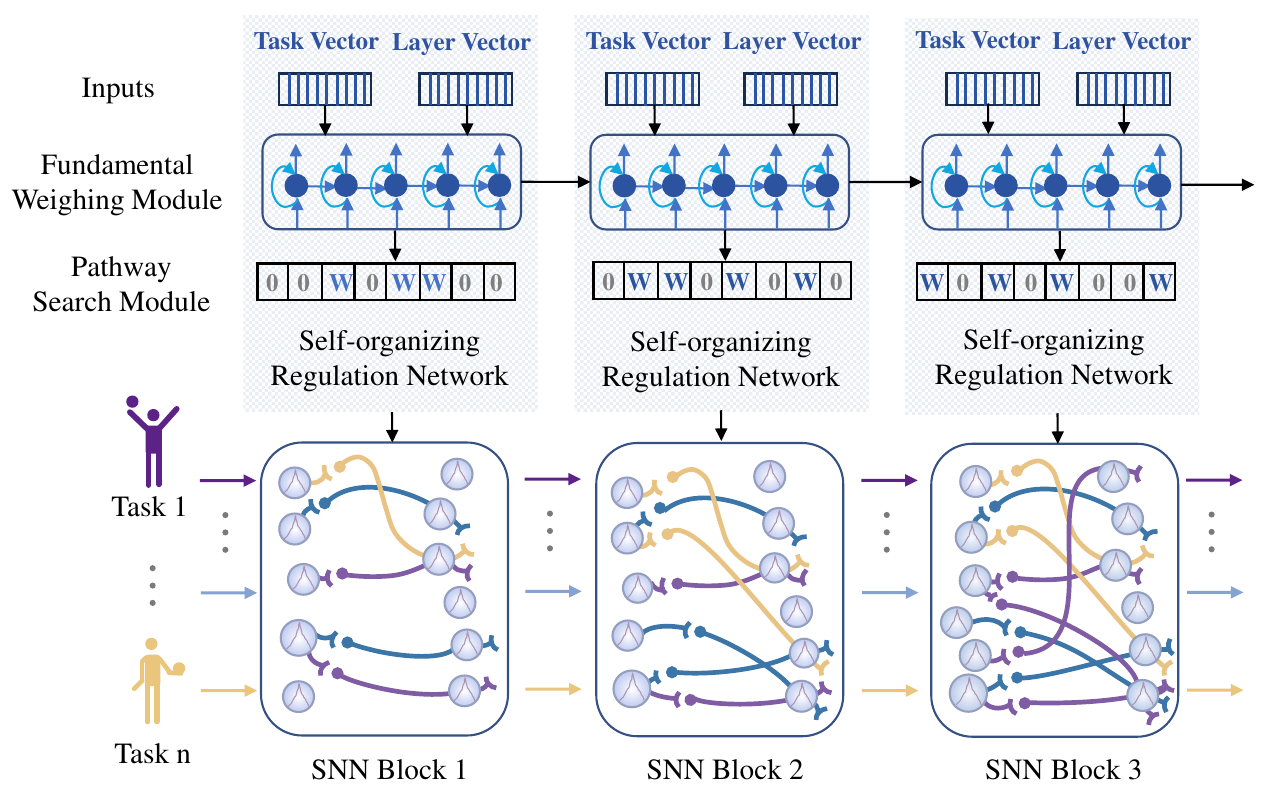} 
    \caption{The process of SOR-SNN model; adapted from the studies in~\cite{bib300}. 
    Each SNN block includes a self-organizing regulation network which selectively activates task-specific sparse pathways. 
    For instance, the purple connections represent the pathway for Task 1. The numerous combinations of connections enable the limited SNN to incrementally learn a larger number of tasks.}
    \label{sor2}
\end{figure}

\textbf{Self-Organized Regulation SNN (SOR-SNN)} employs a \textit{pathway search module} to adaptively activate task-specific sparse neural pathways based on fundamental weights $W_t$~\cite{bib300} (Fig.~\ref{sor2}). 
Each synapse has two states (i.e., active and inactive), and is determined by learnable synaptic selection parameters. 
The model decides whether to activate or inhibit each weight by comparing the learnable parameter $A_s$ with the threshold $\tilde A_s$, where activation is preferred if $A_s > \tilde A_s$. The experimental results on the CIFAR100 dataset in Class-IL scenario demonstrated an average accuracy of $86.65\%\pm0.20$ for 20 steps, where each task contains 5 classes and an accuracy improvement of 2.20\% compared to the DSD-SNN. Moreover, on the Mini-ImageNet dataset, it reported $>55\%$ accuracy, higher than the existing studies (see summary in Table~\ref{table_8} and comparative empirical analysis in Table~\ref{comparison-SSN-DNN} and Table~\ref{table11}).
This study showed that SOR-SNN enables adaptive reorganization, BWT, and self-repair capabilities.
Despite its benefits, this method requires complex dynamic connectivity capabilities, hence making it challenging for deployments under OCL scenario.

%%%%%%------
\subsubsection{Rehearsal/Replay-based Approach} 

\begin{figure}[t]
    \centering
    \includegraphics[width=0.5\textwidth]{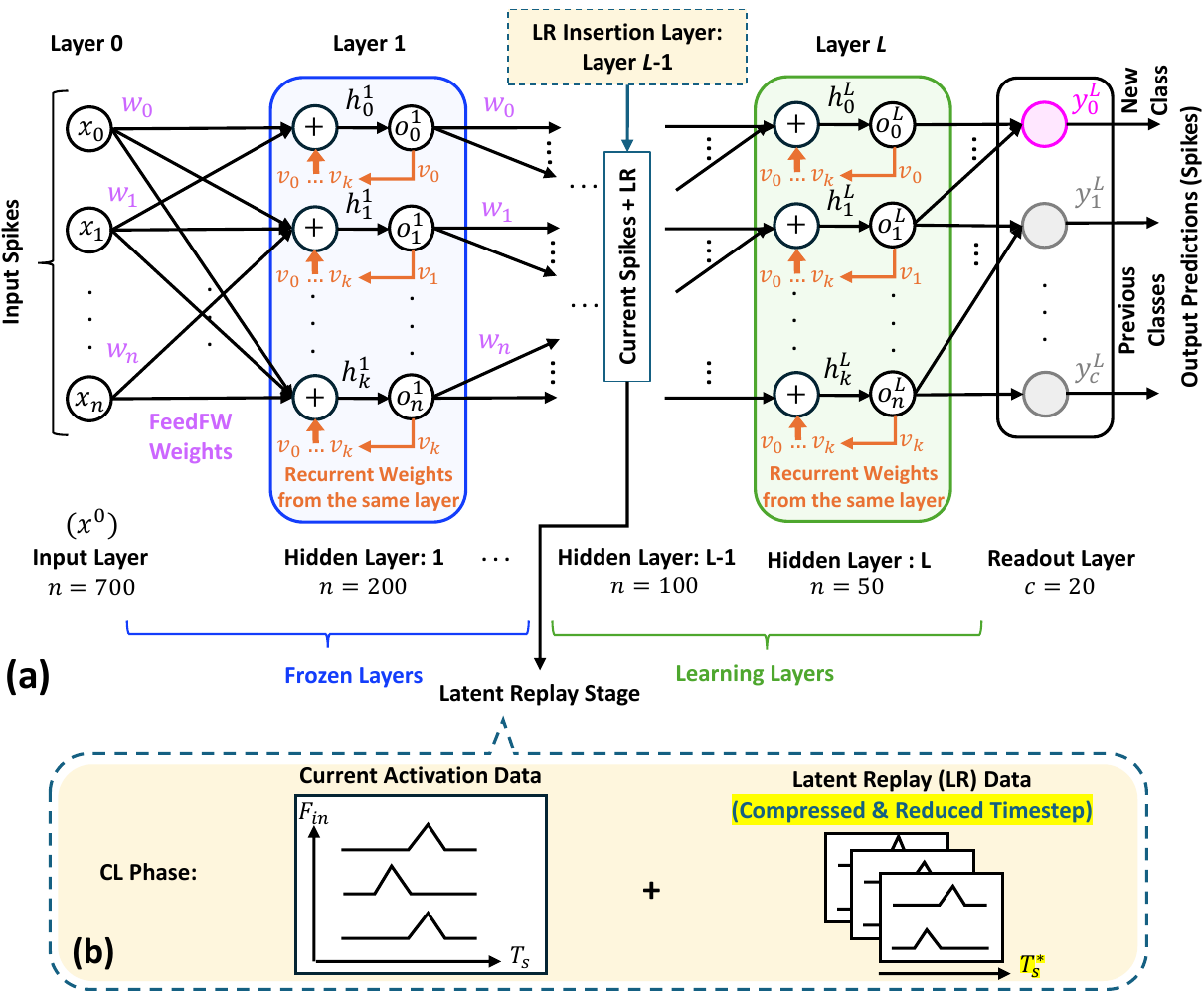} 
    \caption{(a) SNN architecture with recurrent neurons and latent replay (LR) and (b) configuration of current activation data and LR data~\cite{minhas2025replay4ncl}.}
    \label{SNNarch}
\end{figure}

A memory replay approach using ER method has been developed, where a CSNN is trained to learn in Class-IL and TA scenarios~\cite{bib301}. 
For resource-constrained devices, the memory-efficient \textbf{Latent Replay (LR)}-based method has been developed, which stores compressed data representations~\cite{bib302}. 
The Latent Replay training involves a pre-training phase, where SNN is divided into two parts i.e., \textit{frozen and learning layers}. 
The network stores latent replays for memory, and only trains the learning layers on new data.
Here, the main challenges are related to the efficient store-load mechanisms for replay data under OCL scenario. The implementations are based on the Python framework. Experimental results on SHD dataset showcased a Top-1 accuracy of 92\% in the CLass-IL scenario and a memory-accuracy trade-off with only 4\% accuracy drop due to compression.
Unlike prior method~\cite{bib302} that rely on long timesteps and compression-decompression for accuracy, \textbf{Replay4NCL}~\cite{minhas2025replay4ncl} advances NCL for embedded AI systems by significantly reducing latency and energy consumption. It compressed latent data and replayed it with fewer timesteps during NCL training. Parameter adjustments (e.g., neuron thresholds, learning rate) compensated for reduced spikes. The SNN (Fig.~\ref{SNNarch}) was implemented in Python in a Class-IL scenario with 19 tasks for pre-training and a 20th task for CL. Evaluations on the SHD dataset demonstrated 90.43\% accuracy compared to 86.22\% of the baseline~\cite{bib302} at layer 3, with 4.88$\times$ speed-up, 20\% latent memory saving, and 36.43\% energy savings (see summary in Table~\ref{table_8} and comparative empirical analysis in Table~\ref{comparison-SSN-DNN} and Table~\ref{table11}).
%

%%%%----
\subsubsection{Regularization-based Approach}

\begin{figure}[h]
    \centering
    \includegraphics[width=0.3\textwidth]{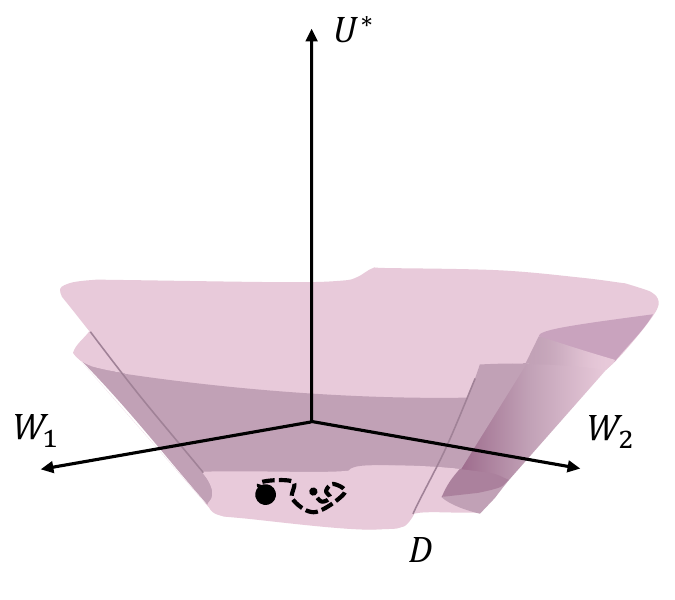} 
    \caption{The effective potential \textit{force field} ($U^*$) created by the R-STDP mechanism, which prevents the Brownian motion (represented by the dashed line) from leaving the optimal weight domain $D$; adapted from studies in~\cite{bib339}.}
    \label{ld}
    \vspace{-0.2cm}
\end{figure}

This approach employs regularization terms to balance the old and new tasks, such as \textit{Noise Regularization}~\cite{bib344}, \textit{Freezing Large Weights}, and \textit{Stochastic Langevin Dynamics}~\cite{bib339}.
For instance, Langevin Dynamics exploit the fact that each weight $w_i$ can vary without impacting the accuracy in the domain $D_i$ if constrained to a specific space when learning new tasks; see Fig.~\ref{ld}. The SNN was implemented in SpykeTorch~\cite{bib341}, an open-source PyTorch-based framework supporting non-leaky IF neurons and local learning rules like STDP and R-STDP, see Table~\ref{Frameworks}. The experiments were performed in a Task-IL scenario using the MNIST dataset (i.e., Task1) for initial training in a layer-by-layer training approach, where layer-S1 and layer-S2 were trained using STDP, while the final layer-S3 was trained with R-STDP. The Extended MNIST (EMNIST) dataset containing both letters and digits was then used for subsequent training (i.e., Task 2), assuming MNIST data is not available. The classification accuracy for digits reached $92.0\pm0.1\%$ and for letters it reached $79.7\pm0.5\%$ (see summary in Table~\ref{table_8} and comparative empirical analysis in Table~\ref{comparison-SSN-DNN} and Table~\ref{table11}).
This study highlighted that Langevin dynamics can exhibit very similar performance as the replay-based methods while being less memory-intensive.
However, its compute requirements are relatively high, thus requiring further studies for OCL scenario. 

\subsubsection{Hebbian Learning}

\begin{figure} [h]
    \centering
    \includegraphics[width=0.5\textwidth]{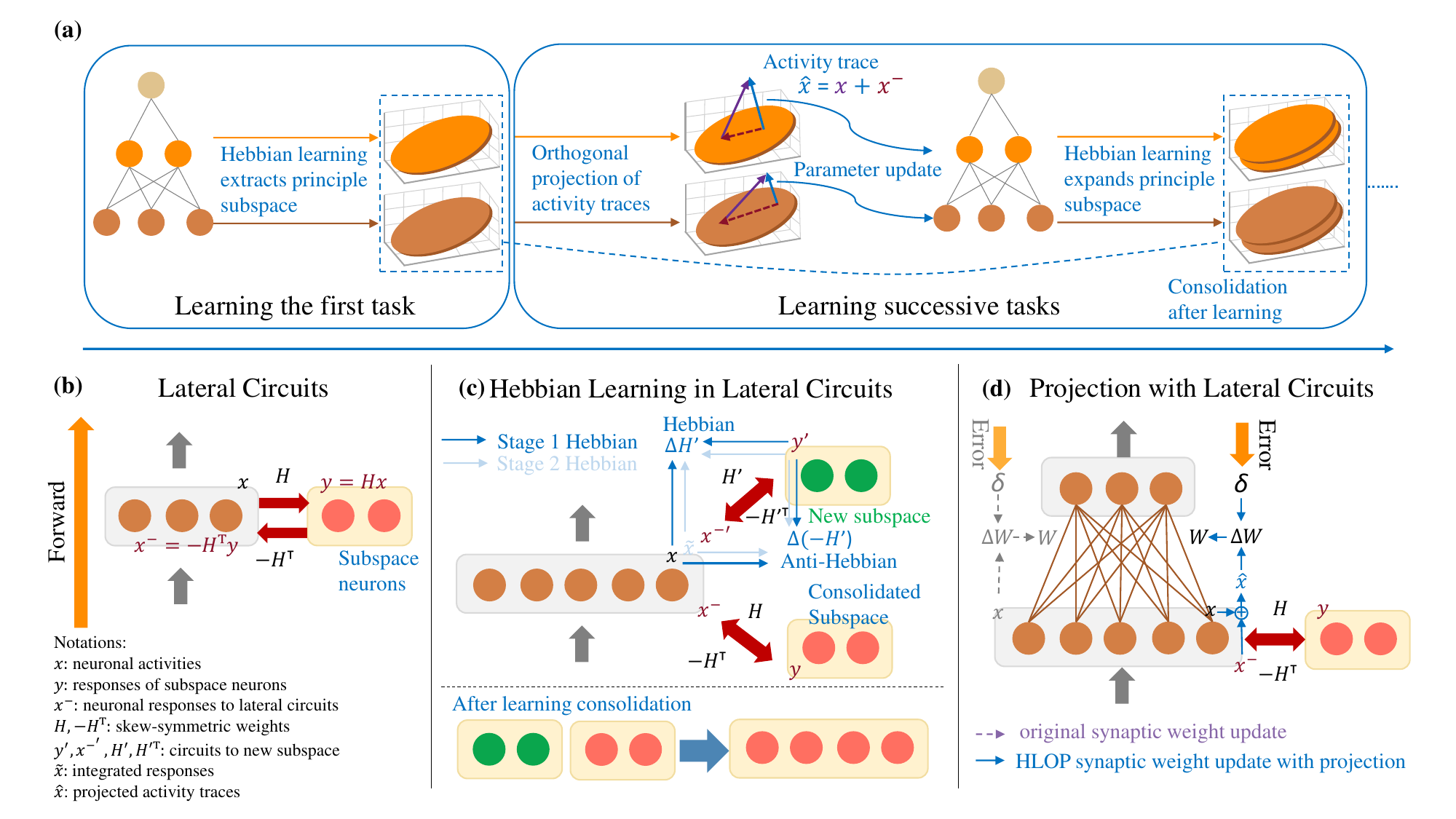} 
    \caption{An illustration of the HLOP; adapted from studies in~\cite{bib350}. (a) Overview of the method. (b) Skew-symmetric connection weights in lateral circuits. (c) Construction of new subspaces for new tasks with Hebbian learning in lateral circuits. (d) Recurrent lateral connections based orthogonal projection.}
    \label{hlop}
\end{figure}

\textbf{Hebbian Learning-based Orthogonal Projection (HLOP)} method leverages lateral connections and Hebbian learning to achieve CL~\cite{bib350}.
It employs orthogonal gradient projection to modify pre-synaptic activity traces (Fig.~\ref{hlop}) to ensure that weight updates for new tasks do not interfere with the weights associated with the old tasks.
Hebbian learning is useful to extract the principal subspace of neural activities, and updates synaptic weights based on the correlation between pre- and post-synaptic neuron activities, while the activity traces are updated using lateral signals. The implementations are based on the PyTorch framework~\cite{paszke2019pytorch}. Experimental results on various datasets and scenarios, including Permuted MNIST (PMNIST), CIFAR-100, and a combination of datasets like CIFAR-10, MNIST, SVHN, Fashion MNIST, and notM
NIST, demonstrated higher accuracy such as 95.15\% for PMNIST and lower forgetting (i.e., BWT) across different training methods like Dynamic Spike Representation (DSR), BPTT with SG, and Online Training Through Time (OTTT). HLOP also showcased superior performance compared to DNN-based methods such as EWC, HAT, and GPM highlighting its potential for robust NCL (see summary in Table~\ref{table_8} and comparative empirical analysis in Table~\ref{comparison-SSN-DNN} and Table~\ref{table11}).

\begin{sidewaystable*}
\caption{Summary of the state-of-the-art works for energy-efficient CL employing bio-plausible SNNs (i.e., NCL).}
\label{table_8}
\centering
\footnotesize
\begin{tabular*}{\textheight}{@{\extracolsep\fill}cccccccccccc}
\cline{1-10} 
\\
\textbf{Work,} & \textbf{CL} & \textbf{Learning} & \textbf{Neuron} &  \textbf{Neural} & \textbf{Learning} & \textbf{Learning} & \textbf{Optimization} & \textbf{Dataset} & \textbf{Software}
\\ %
\textbf{Year} & \textbf{Scenario} & \textbf{Setting} & \textbf{Model} & \textbf{Coding} & \textbf{Rule} & \textbf{Rate} & \textbf{Technique} & &  &  \\ %
\\
\cline{1-10} 
\\
ASP~\cite{bib108}, & Class-IL & Unsupervised & LIF & Rate & STDP & Adaptive & Weight Decay & MNIST & Brian  & \\
(2017) & & & & & & & & & \cite{bib303} \\
& & & & & & & & & \\
SpikeDyn~\cite{bib296}, & Class-IL & Unsupervised & LIF & Rate & STDP & Adaptive & Weight Decay & MNIST & Bindsnet & \\
(2021) & & & & & & & & & ~\cite{hazan2018bindsnet} \\
& & & & & & & & & \\
lpSpikeCon~\cite{bib297}, & Class-IL & Unsupervised & LIF & Rate & STDP & Adaptive & Weight & MNIST & Bindsnet \\
(2022) & & & & & & & Quantization & & ~\cite{hazan2018bindsnet}\\
& & & & & & & & & \\
CFN \cite{bib72}, & Class-IL & Unsupervised & LIF & Rate & STDP & Adaptive & - & MNIST & - \\
(2020) & & & & & & \\
& & & & & & & & & \\
SpNCN~\cite{bib342}, & Task-IL & Supervised & LIF & Rate & ST-LRA, & - & - & MNIST~\cite{bib305} & - \\
(2023) & & & & & ST-LRA+ & & & FMNIST~\cite{bib349} & & \\
& & & & & STDP & & & Not-MNIST~\cite{notMNIST_dataset} & & \\
& & & & & & & & & \\
\cite{bib317},(2024) & Task-IL & Supervised & Enhanced & TTFS & Back- & Fixed & Weight and Delay & Split-MNIST & FPGA, & \\ %
& & & model of & & Propagation  & & Quantization & & Python & \\ %
& & & \cite{bib319} & & & &  & & \\ %
& & & & & & & & & \\
\cite{bib299},(2022) & Class-IL & Supervised & SRM & Rate & Bayesian & - & - & Split-MNIST, & Intel's  \\
& & & & & Learning & & & MNIST-DVS & Lava \\
& & & & & & & & \cite{bib323} & \cite{bib322} \\
& & & & & & & & & \\
DSD-SNN & Task-IL, & Supervised & LIF & - &  Back- & - & - & MNIST & BrainCog & \\
\cite{bib304},(2023) & Class-IL & & & & Propagation & & &  N-MNIST~\cite{bib310}, & \cite{bib309} \\
& & & & & & & & Split-CIFAR100 & \\
& & & & & & & & & \\
SOR-SNN & Class-IL & Supervised & LIF & - & Back- & Adaptive & - & CIFAR100~\cite{bib306}, & - \\
\cite{bib300},(2023) & & & & & Propagation & & & ImageNet~\cite{bib34} \\
& & & & & & & & & \\
\cite{bib301},(2023) & Class-IL, & Supervised & LIF & - & BPTT & Fixed & - & MNIST & PyTorch \\
& TA & & & & & Fixed & & & \cite{paszke2019pytorch} \\
& & & & & & & & & \\
\cite{bib302},(2023) & Sample-IL, & & LIF & - & BPTT & - & Time  & SHD & Python \\
& Class-IL & Supervised & & & & & Compression & \cite{bib312} & \\
& & & & & & & & & \\
Replay4NCL & Class-IL & Supervised & LIF & - & BPTT & Adaptive & Timestep & SHD & Python \\
~\cite{minhas2025replay4ncl},(2025)  & & & & & & & Reduction & \cite{bib312} & \\
& & & & & & & & & \\
\cite{bib339},(2022) & Task-IL & Unsupervised & IF & Rank- & STDP, &  $a^+< 0.15$, &  - & MNIST, & SpykeTorch\\
& & & & Order & R-STDP & $a^->-0.1125$ & & EMNIST~\cite{bib340} & \cite{bib341} \\
& & & & & & & & & \\
HLOP-SNN & Task-IL, & Supervised & LIF & Rate & Hebbian & Fixed & - & PMNIST \cite{bib305}, & Python & \\
\cite{bib350},(2024) & DIL & & & & & & & CIFAR100\\
\\
\cline{1-10} 

\end{tabular*}
\end{sidewaystable*}
\subsection{Hybrid Learning Paradigms} 
\label{sec3.3}
\begin{figure}[t]
    \centering
    \includegraphics[width=0.5\textwidth]{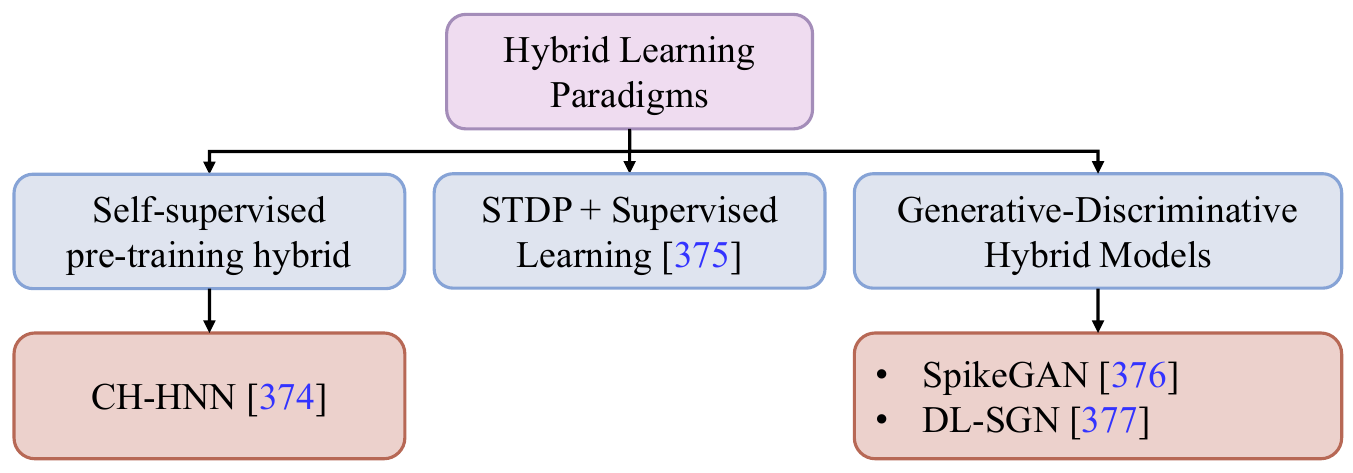} 
    \caption{A taxonomy of hybrid approaches combining supervised and unsupervised learning paradigms. We have highlighted the main categories (blue blocks), with their works shown (red blocks).}
    \label{ha_taxonomy}
\end{figure}

The scalability of most NCL methods to complicated problems with many tasks or complex inputs is challenging. In this section, we explore hybrid approaches that combine supervised and unsupervised learning paradigms to address CF and improve CL performance. We categorize these approaches into three main classes: 1) self-supervised pre-training hybrids, 2) STDP + supervised learning hybrids, and 3) generative-discriminative hybrid models; see Fig.~\ref{ha_taxonomy}.

\subsubsection{Self-supervised Pre-training Hybrid}

\textbf{Cortico-Hippocampal Hybrid Networks (CH-HNN) }~\cite{shi2025hybrid} 
integrated ANNs for generalization and SNNs for specific memory encoding, inspired by dual-memory corticohippocampal circuits. It employed ANN-guided episode inference and metaplasticity with SG training to dynamically regulate synaptic plasticity.
CH-HNN was validated in two robotic case studies: a Unitree GO1 robot performed real-time MNIST digit recognition using ANN-SNN inference, and a Unitree Z1 arm achieved 82$\pm7.25\%$ accuracy in sCIFAR-100 object grasping using YOLO. Across benchmarks like split MNIST, pMNIST, sCIFAR-100, sTiny-ImageNet (70.72\% accuracy), and DVS Gesture, CH-HNN outperformed EWC, SI, XdG, iCaRL, and FOSTER in stability-plasticity balance, disparity, and efficiency. For neuromorphic deployment, int8 quantization incurred minimal loss, with SNNs yielding 60.82\% power savings over ANNs. Although CH-HNN supports various spiking neuron models, their selection involves trade-offs between computational cost, biological fidelity, and deployment efficiency.

\subsubsection{STDP + Supervised Learning}

The study in~\cite{rathi2020enabling} integrated ANN-to-SNN conversion with STDB to converge to optimal accuracy in fewer epochs compared to training from scratch. Evaluations on image classification tasks using CIFAR-10, CIFAR-100, and ImageNet for VGG and ResNet architectures showed that SNNs trained with the hybrid method required significantly fewer time steps ($10\times - 25\times$ fewer) to achieve competitive accuracy compared to purely converted SNNs. For instance, the model achieved a top-1 accuracy of 65.19\% on the ImageNet dataset using only 250 time steps. Although, the study showed promise, the inherent challenges of converting SNNs, such as their non-differentiable nature, still pose difficulties that may affect broader applicability. 

\subsubsection{Generative-Discriminative Hybrid Models}

\textbf{SpikeGAN}~\cite{rosenfeld2022spiking}
is a hybrid generative model that combined an SNN generator with an ANN discriminator to learn and generate temporal spiking data distributions. 
The SNN generator captured spatio-temporal patterns, while the ANN discriminator enabled adversarial training akin to standard GANs. Bayesian learning was applied to the generator's weights, and a continual meta-learning framework supported adaptation to multiple real-world distributions. Evaluated on handwritten digit generation, SpikeGAN outperformed ANN classifier accuracy by 20\% and enabled SNN classifiers trained on synthetic data to match the performance of those trained on real rate-encoded inputs. The model leveraged the complementary strengths of ANNs and SNNs; however, it may suffer from sample space coverage issues, as indicated by Train-on-Synthetic-Test-on-Real (TSTR) errors, pointing to potential limitations in robustness when generalizing to unseen real data.

\textbf{Dynamic Lifelong learning with Spiking Generative Networks (DL-SGN)}~\cite{zhang2024spiking} is a hybrid SNN-ANN lifelong learning framework designed to enable energy-efficient image generation and classification on edge devices while addressing CF. It integrated ANN-based dynamic expert modules that grow through Dynamic Knowledge Adversarial Fusion (DKAF), an SNN-based student module combining a VAE and classifier, and an ANN-based assistant discriminator trained adversarially to support generalization. Implemented in PyTorch, DL-SGN achieved substantial improvements in FID scores e.g., MNIST (56.26 vs. 95.26) and CIFAR10 (92.46 vs. 192.2) and surpassed ANN-based replay methods such as LTS and LGM in classification accuracy across complex benchmarks like SVHN-CIFAR10-ImageNet10 task (average: 52.43\%). In semi-supervised lifelong learning, it reduces MNIST classification error to 2.47\%, significantly outperforming DGR (7.27\%). However, the framework incurs notable computational overhead due to expert module expansion and adversarial training. Future directions include optimizing expansion strategies, developing SNN-native generative models, and unifying supervised and unsupervised objectives for improved adaptability in dynamic environments.

Our analysis reveals that hybrid architectures consistently outperform single-paradigm approaches by leveraging complementary strengths of different learning mechanisms. Key observations include the effectiveness of biologically-inspired designs CH-HNN, the scalability advantages of generative replay methods, and the computational efficiency gains from hybrid training protocols. These approaches demonstrate significant improvements in both Task-IL and Class-IL scenarios across standard benchmarks including MNIST, CIFAR-100, and ImageNet variants.

%%%%%%%%%%%%%%%%%%%%%%%%%%%%%%%%%%%%%%
\subsection{Efficiency Enhancement Methods for NCL} 
\label{sec3.4}

To further reduce the memory footprint and energy consumption of NCL methods, several optimization methods proposed in literature are discussed below, addressing key question Q6.

%%%%----
\subsubsection{Reduction of SNN Operations} 
 
\begin{figure}[h]
    \centering
    \includegraphics[width=0.5\textwidth]{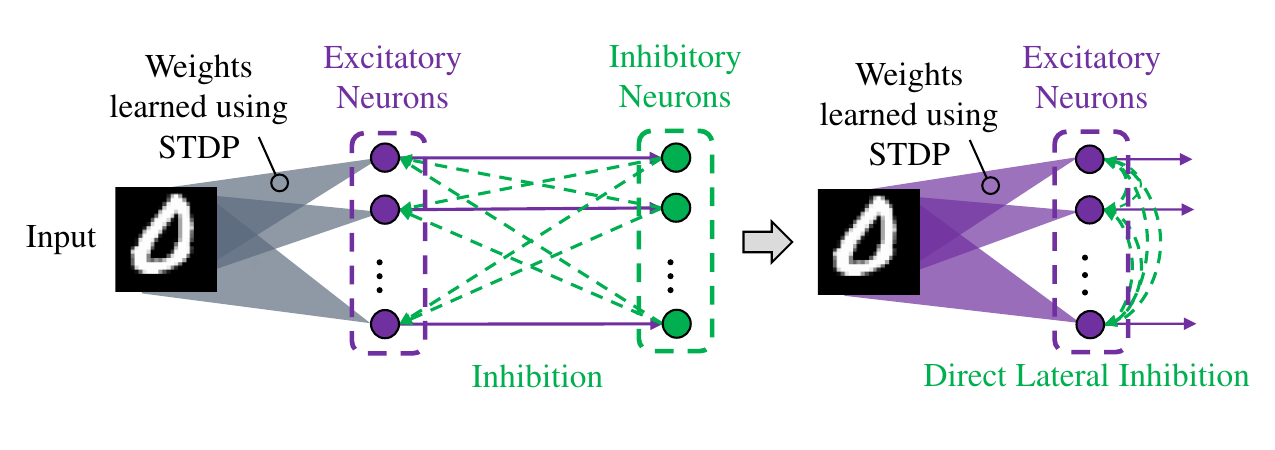} 
    \caption{Replacement of the inhibitory neurons with direct lateral inhibitions; adapted from studies in~\cite{bib296}.}
    \label{rno}
\end{figure}

Real-time resource-constrained applications require fast and efficient processing SNNs. Their operations can be reduced through several network optimizations by:
\begin{itemize}
    \item Leveraging \textit{sparse neurons and sparse synapses} to reduce the number of neuron operations and connections/weights, respectively.
    \item Utilizing \textit{temporal encoding} uses timing of spikes to encode information in a way that can reduce computational complexity compared to \textit{rate encoding}.
    \item Employing \textit{simple neuron models} like LIF instead of complex models to reduce the computational complexity required per neuron.
\end{itemize}
\smallskip
SpikeDyn~\cite{bib296} replaced the inhibitory neurons with the direct lateral inhibitions in the SNN model used in ASP~\cite{bib108}, which consisted of input, excitatory, and inhibitory layers (Fig. \ref{rno}), thereby eliminating the operations in the inhibitory layer.

%%%%----
\subsubsection{Weight Quantization} 

This method enhances SNNs efficiency by representing network weights with lower precision instead of using full-precision floating-point numbers~\cite{Ref_Putra_QSpiNN_IJCNN21, Ref_Putra_ReSpawn_ICCAD21, Ref_Putra_SparkXD_DAC21, Ref_Putra_EnforceSNN_FNINS22, Ref_Putra_RescueSNN_FNINS23, Ref_Putra_SoftSNN_DAC22, Ref_Putra_SNN4Agents_FROBT24}.
Weights can be quantized to a smaller number of selected discrete levels such as 8-bit, 4-bit, or even binary, using \textit{uniform} or \textit{non-uniform quantization}~\cite{bib314}. For instance, lpSpikeCon~\cite{bib297} reduced SNN weights using uniform quantization with \textit{truncation}~\cite{bib313,bib314}. Similarly, to reduce the required on-chip memory, the synaptic weights and dendritic delays are quantized in \cite{bib318}. Thus, keeping the user-defined number of bits and removing the remaining bits from the fractional part, reduces the storage size, speeds up the training and inference processes, and enables more efficient hardware implementations~\cite{Ref_Putra_FSpiNN_TCAD20}.  

%%%%%%%%%%%--------------

\subsubsection{Knowledge Distillation}
To enhance the efficiency of SNNs in resource-constrained environments, knowledge distillation has been widely explored as an optimization technique~\cite{lee2021energy,hong2023lasnn,xu2023constructing,qiu2024self}. In particular, teacher-student knowledge distillation enables smaller models to retain the performance of larger models while reducing memory and computational overhead.
Shaw et al. ~\cite{shaw2023teacher} demonstrated the successful application of teacher-student knowledge distillation for radar perception on embedded accelerators, optimizing neural networks for efficient real-time processing. Similar techniques could be leveraged in NCL to improve computational efficiency and reduce model complexity for neuromorphic edge applications. 

%%%%%%%%%%%%%%%%%%%%%%
%%%%%%%%%%%%%%%%%%%%%%%%%%%%%%%%%%%%%%%%

\subsection{Compatibility with Neuromorphic Hardware}
\label{sec_NC_compatibility}

The requirements for implementing a CL method on neuromorphic hardware include: (1) computational units/modules that can execute the CL algorithm, and (2) sufficient size of computational and memory resources. 
Specifically, the neuromorphic hardware should have compute units that can execute the given CL algorithm on-chip.
For instance, if the CL algorithm employs a bio-plausible STDP learning rule, then the neuromorphic hardware should have a learning unit that can execute it.
Meanwhile, the size of computational and memory modules from the neuromorphic hardware defines how much workload can be executed at one time, thereby computational and memory constraints will affect the performance in terms of processing latency, throughput, and energy efficiency.

The compatibility of reviewed methods on the existing neuromorphic hardware can be analyzed based on the capability of the hardware to perform on-chip learning. The summary of existing neuromorphic hardware and their on-chip learning capabilities is provided in Table~\ref{Tab_onchipLearning}. 
This table shows that, most of neuromorphic hardware platforms do not support on-chip learning, or only support specific on-chip learning mechanisms (e.g., bio-plausible STDP). 
Therefore, most of reviewed methods may not be executed on the existing neuromorphic hardware, rather they can be executed on the host processing unit (CPU/GPU), which is capable of executing more types of operations, addressing the key question Q7. 

\begin{table}[h]
\caption{Summary of the existing neuromorphic hardware platforms with their on-chip learning capabilities.}
\centering
\begin{tabular}{|c|c|l|}
\hline
\textbf{Neuromorphic} & \textbf{On-chip} & \multicolumn{1}{c|}{\textbf{On-chip Learning}}   \\ 
\textbf{Processor} & \textbf{Learning} & \multicolumn{1}{c|}{\textbf{ Mechanism}}   \\ \hline
\hline
NeuroGrid~\cite{Ref_Neurogrid_JPROC14} & No & - \\ \hline
ROLLS~\cite{Ref_Akopyan_TrueNorth_TCAD15} & Yes & SDSP \\ \hline
TrueNorth~\cite{Ref_Akopyan_TrueNorth_TCAD15} & No & - \\ \hline
SpiNNaker~\cite{furber2014spinnaker} & Yes & STDP \\ \hline
BrainscaleS-2~\cite{pehle2022brainscales} & Yes & STDP, R-STDP, SG \\ \hline
Loihi~\cite{Ref_Davies_Loihi_MM18} & Yes & STDP, Surrogate Gradient Learning \\ \hline
Tianjic~\cite{Ref_Pei_Tianjic_Nature19} & No & - \\ \hline
MorphIC~\cite{Ref_Frenkel_MorphIC_TBCAS19} & Yes & SDSP \\ \hline
DYNAP~\cite{Ref_SynSense_DYNAP} & No & - \\ \hline
Akida~\cite{Ref_BrainChip_Akida} & Yes & Last-layer Learning \\ \hline
PAIBoard~\cite{chen2024paiboard} & No & -\\ \hline
PAICORE~\cite{zhong2024paicore} & Yes & STDP \\
\hline
\end{tabular}
\label{Tab_onchipLearning}
\end{table}

\begin{table*}[h!]
\centering
\scriptsize
\caption{Comparative quantitative analysis of the surveyed NCL methods evaluated on P-MNIST, Split CIFAR-100, Split MNIST, and ImageNet datasets with relevant state-of-the-art DNN-based CL methods addressing energy-efficiency problem with various settings and scenarios.}
\setlength{\tabcolsep}{1pt}
\begin{tabular*}{\textwidth}{@{\extracolsep{\fill}} p{1.7cm}  p{2.1cm} p{2.3cm} p{1.4cm} p{1.5cm} p{1.5cm} p{1.7cm} p{1.1cm}}
\toprule
\toprule
\multicolumn{8}{c}{\textbf{P-MNIST}} \\
\midrule
\midrule
\textbf{Methods} & \textbf{Network/} & \textbf{Accuracy} & \textbf{CL} & \textbf{CL} & \textbf{Memory} & \textbf{Latency} & \textbf{Energy} \\
& \textbf{Model} &  & \textbf{Setting} & \textbf{Scenario} &  \\
\midrule
\cellcolor{DNNColor}GPM~\cite{bib63} & FC & $93.91\%\pm0.16$ & Supervised & Task-IL, \newline Class-IL   & n/a  & 245s (train) & n/a \\
\midrule
\cellcolor{DNNColor}~\cite{bib276} & FC & $84.3\%\pm0.3$ & Supervised & Task-IL & n/a & n/a & n/a\\
\midrule
\cellcolor{DNNColor}A-GEM~\cite{bib275} & FC &  $\approx$	90\% & Supervised & Task-IL & n/a & n/a & n/a \\
\midrule
\cellcolor{DNNColor}GSS~\cite{bib125} & MLP &	$77.3\%\pm0.5$ & Supervised	& Task-IL  & n/a & n/a & n/a  \\
\midrule
\cellcolor{SNNColor}HLOP-SNN~\cite{bib350} & FC &	$95.15\%$ & Supervised &	Task-IL, \newline Domain-IL & n/a & n/a  & n/a \\
\midrule
\midrule
\multicolumn{8}{c}{\textbf{CIFAR-100}} \\
\midrule
\midrule
\cellcolor{DNNColor}GPM~\cite{bib63} & 5-layer \newline AlexNet & $72.48\%\pm0.40$ \newline (10-split)  & Supervised & Task-IL, \newline Class-IL & 5.84M \newline parameters & 770s (train) & n/a \\
\midrule
\cellcolor{DNNColor}Adam-NSCL~\cite{bib279} & ResNet-18 &   75.95\% (20-split) & Supervised & Class-IL & n/a & n/a & n/a \\
\midrule
\cellcolor{DNNColor}\cite{bib276} & ResNet18 &  $71.0\%\pm0.3$ & Supervised & Task-IL & n/a & n/a  & n/a	\\
\midrule
\cellcolor{DNNColor}GDumb~\cite{bib232} & ResNet32 & $60.3\%\pm0.85$ & OCL & Task-IL & $1105^1$ & 60s (train) & n/a \\
\midrule
\cellcolor{DNNColor}CoPE~\cite{de2021continual} & Resnet18 & 21.62\%±0.69 & Supervised & - & $5000^1$	\\
\midrule
\cellcolor{DNNColor}X-DER~\cite{boschini2022class} & ResNet18 & 49.93\% & Supervised & Class-IL & $\approx$ 50MB & 5$\times$ more runtime \newline than DER++ & n/a	\\
\midrule
\cellcolor{DNNColor}ACAE-REMIND~\cite{wang2021acae} & Resnet-18 \& \newline Resnet-32 & 62.30\% AOC \newline (50 steps) & OCL	& Class-IL, \newline TA & 12.8 MB & n/a & n/a	\\
\midrule
\cellcolor{DNNColor}A-GEM~\cite{bib275} & ResNet18 & $\approx$ 62\% & Supervised	& Task-IL & 10 times lower than EWC	& 100 times faster than EWC & n/a \\	
\midrule
\cellcolor{DNNColor}DVC~\cite{gu2022not}	& Resnet18 & $24.1\%\pm0.8$	& OCL &	Class-IL & $5000^1$ &	n/a &	n/a \\	
\midrule
\cellcolor{DNNColor}SDAF~\cite{yu2022mitigating} & ResNet18 \&  MLP & $39.0\%\pm0.3$ & OCL &	Class-IL &	$5000^1$ & n/a & n/a \\		
\midrule
\cellcolor{DNNColor}SCR~\cite{mai2021supervised} & ResNet18 \& MLP & $37.8\%\pm0.3$ & Supervised& Class-IL & $5000^1$ & $\approx$ 250s & n/a \\
\midrule
\cellcolor{SNNColor}DSD-SNN~\cite{bib304} & Multi-CONV \& \newline 1 FC layers	& $77.92\%\pm0.29$, \newline 60.47\% (10 steps) & Supervised & Task-IL, \newline Class-IL &	37.48\% network compression & n/a &	n/a	\\
\midrule
\cellcolor{SNNColor}SOR-SNN~\cite{bib300} &	ResNet-18 &  $86.65\%\pm0.20$ \newline (20 steps)	& Supervised & Class-IL & 0.32M \newline parameters & n/a & 	n/a	\\
\midrule
\cellcolor{SNNColor}HLOP-SNN~\cite{bib350} &	ResNet-18 & 78.58\% & Supervised	& Task-IL, \newline Domain-IL & n/a	& 	n/a & 	n/a	\\
\midrule
\midrule
\multicolumn{8}{c}{\textbf{MNIST}} \\
\midrule
\midrule
\cellcolor{DNNColor}DER++~\cite{bib231} & FC &	$92.77\%\pm 1.05$ &  Supervised& Domain-IL & $500^1$ & n/a	& n/a \\	
\midrule
\cellcolor{DNNColor}EEC~\cite{ayub2021eec} & 3-layer CONV \newline autoencoder & 97.83\% (10 classes) & Supervised &	Class-IL & 0.2MB disk space & n/a &n/a	\\	
\midrule
\cellcolor{DNNColor}CoPE~\cite{de2021continual} &MLP&  $93.94\%\pm0.20$ & Supervised & - & $2000^1$ & n/a & n/a	\\	
\midrule
\cellcolor{DNNColor}GDumb~\cite{bib232}	& MLP & $88.9\%\pm0.6$	& OCL & Class-IL  & $300^1$ & 60s (train)	& n/a	\\
\midrule
\cellcolor{SNNColor}\cite{bib339} & 3 CONV layers &  $92.0\%\pm0.1$	& Unsupervised & Task-IL &  n/a & $2.5min^2$ & n/a	\\	
\midrule
\cellcolor{SNNColor}DSD-SNN~\cite{bib304} & Multi-CONV & $97.30\%\pm0.09$ & Supervised &	Task-IL & 34.38\% network compression	& n/a &	n/a \\
\midrule
\cellcolor{SNNColor}Bayesian CL~\cite{bib299}	& Multi-FC layers & 	 $85.44\%\pm0.16$ & Supervised & Class-IL & 5× reduction	 & n/a & n/a \\	
\midrule
\cellcolor{SNNColor}SpNCN~\cite{bib342} & 4 FC layers & 96.50\%	& Supervised & Task-IL & n/a	& n/a & n/a \\	
\midrule
\cellcolor{SNNColor}ASP~\cite{bib108} & 2 FC layers & 94.20\% & Unsupervised &	Class-IL &	n/a	 & n/a &	n/a\\	
\midrule
\cellcolor{SNNColor}SpikeDyn\cite{bib296} & 2 FC layers & Improved 23\% (new task), 4\% (old task) than ASP	& Unsupervised & Class-IL & <3000kB & $0.2s^3$ & 57\% (train), 51\%(inference) lower than ASP  \\
\midrule
\end{tabular*}
\label{comparison-SSN-DNN} 
\vspace{-5pt}
\end{table*}

\addtocounter{table}{-1}
\begin{table*}
\label{comparison-SSN-DNN}
\centering
\scriptsize
\caption{Continued.}
\setlength{\tabcolsep}{1pt}
\begin{tabular*}{\textwidth}{@{\extracolsep{\fill}} p{1.4cm}  p{1.5cm} p{2.2cm} p{1.4cm} p{1.5cm} p{2cm} p{1.1cm} p{1.6cm}}
\toprule
\cellcolor{SNNColor}lpspikecon\newline ~\cite{bib297}& 2 FC layers	& 68\% (6-bit quantized weights)	& Unsupervised & Class-IL & 8x weight memory reduction with 4-bit than 32-bit &	n/a	& n/a	\\	
\midrule
\cellcolor{SNNColor}\cite{bib317}& 4 FC layers &	80.0\% on FPGA & Supervised & Task-IL &  35.3\% flip-flops, 29.3\% BRAMs	& n/a & n/a \\			
\midrule
\midrule
\multicolumn{8}{c}{\textbf{ImageNet}} \\
\midrule
\midrule
\cellcolor{DNNColor}X-DER\cite{boschini2022class} &	EfficientNet-B2 & 28.19\% & Supervised & Class-IL & $2000^1$ & n/a & n/a \\
\midrule
\cellcolor{DNNColor}SDAF~\cite{yu2022mitigating} & ResNet18 \& MLP & $33.2\%\pm 0.5$ &	OCL & Class-IL & $5000^1$ & n/a & n/a \\
\midrule
\cellcolor{DNNColor}DVC~\cite{gu2022not} & Resnet18 & $19.1\%\pm0.9$ & OCL & Class-IL & $5000^1$ & n/a & n/a \\
\midrule
\cellcolor{DNNColor}\cite{bib276} &	ResNet18 & $39.5\%\pm0.3$ & Supervised & Task-IL &  n/a	& n/a & n/a \\	\midrule	
\cellcolor{DNNColor}SCR~\cite{mai2021supervised} & ResNet18 \& MLP& $35.4\%\pm0.5$ & OCL & Class-IL & $5000^1$ & n/a & n/a \\
\midrule
\cellcolor{SNNColor}SOR-SNN~\cite{bib300} & ResNet-18& >55\% & Supervised & Class-IL & 0.32M \newline parameters & n/a & n/a\\		
\midrule
\end{tabular*}
\begin{tablenotes} \footnotesize
\item \textsuperscript{1} Memory buffer size; \textsuperscript{2} Approximate time one training epoch took for 24k training patterns; \textsuperscript{3} Inference time of an image; \textbf{Note:} Rows with \colorbox{SNNColor}{\strut\hspace{0.5em}} indicate \textbf{NCL} methods and \colorbox{DNNColor}{\strut \hspace{0.5em}} indicate \textbf{DNN-based CL} methods.
\end{tablenotes}
\label{comparison-SSN-DNN} 
\vspace{-5pt}
\end{table*}

%%%%%%%%%%%%%%%%%%%%%%%%%%%%%%%%%%%%%%%%
\subsection{Trade-Off Analysis}
\label{Sec_NCLtradeoff}

The discussion in this section addresses the key question Q9. Hardware implementation of NCL typically aims at improving performance (speed-up) and computational efficiency (e.g., in terms of power/energy consumption), while offering high accuracy. 
However, achieving high accuracy with high speed-up and high efficiency in NCL is a challenging task due to the conflicting nature of their requirements.
For instance, in order to achieve high accuracy, the network model typically requires more resources (e.g., more neurons and weights) to provide more memory for continually storing new knowledge from learning new tasks without forgetting, thus significantly increasing the model size.

\textit{Energy-Accuracy Trade-offs:}
SpikeDyn~\cite{bib296} demonstrated substantial energy savings with 51\% reduction in training energy and 37\% in inference energy for 400 neuron network, (see Table~\ref{table11}), while achieving 21\% accuracy improvement over baseline method~\cite{bib108} for new tasks and 8\% improvement for previously learned tasks. For smaller 200 neuron network, the trade-offs were even more favorable, showing 57\% training and 51\% inference energy reductions with 23\% accuracy improvements. These results indicate that careful architectural optimizations can yield significant energy benefits without accuracy losses.

\textit{Memory-Accuracy Trade-offs:}
Compressed latent replay methods~\cite{bib302, minhas2025replay4ncl} demonstrated memory-accuracy trade-offs, achieving 140× memory reduction (from 22.4 MB to 160 KB) with only 4\% accuracy degradation in Sample-IL scenario~\cite{bib302}. In Class-IL scenario, this method achieved $\approx$86\% accuracy on layer 3 in case of a 1:5 ratio with 320 kB memory usage, compared to naive rehearsal (without compression) requiring 22.4 MB for 89.55\% accuracy, representing a 7× memory efficiency improvement with $\approx$3\% accuracy drop~\cite{bib302}. Replay4NCL~\cite{minhas2025replay4ncl} revealed that storing compressed latent replays at layer 3 and replaying them with fewer timesteps (i.e., 40 instead of 100) with critical parameters adjustment achieved 90.43\% accuracy compared to 86.22\% of the baseline~\cite{bib302}, with 4.88x speed-up, 20\% latent memory saving, and 36.43\% energy saving. 
lpSpikeCon~\cite{bib297} showed that lowering weight precision from 32-bit to 4-bit degrades task-specific accuracy due to information loss, while achieving 8× memory reduction.

\textit{Hardware Platform Energy-Latency Trade-offs}:
Platform-specific performance showed significant latency variations in~\cite{bib296}: 1.71s per inference on Jetson Nano (10W) versus 0.2s on RTX 2080 Ti (250W), indicating that energy-constrained platforms require 8.5× longer processing time but consume 25× less power, providing clear energy-latency trade-off quantification for different deployment scenarios.

Moreover, maintaining high accuracy often requires the model to employ high precision data format, which may not be supported in commodity neuromorphic hardware~\cite{Ref_BrainChip_Akida}, thereby making it even more difficult to reduce the network size. 
On the other hand, neuromorphic hardware platforms typically have limited compute and memory resources, which constrains how the network model will be implemented (i.e., mapped and executed) on the hardware fabric.
Therefore, a larger model usually requires longer processing latency, and hence higher power/energy consumption.
The DSD-SNN model achieved an efficient trade-off between network compactness and CL by dynamically adapting its structure~\cite{bib304}. On CIFAR100, it stabilized at 37.48\% of the full network size while maintaining superior accuracy than the non-pruned variant, which rapidly exhausted memory, leading to performance collapse after six tasks. This highlights the effectiveness of pruning in reducing overhead without compromising learning performance.
In some cases, a very large model may even need to be split into several parts (e.g., layer-based partition), so that each part can be mapped and executed on the hardware fabric.
Consequently, this condition leads to higher processing latency (lower speed-up) and higher power/energy consumption (lower efficiency).

To address this, trade-off analysis is required. 
This analysis aims at identifying the appropriate network model candidates considering the given hardware constraints (e.g., compute and memory budgets), and then guides the selection on the most suitable one~\cite{bib296}. 
For instance, if we can accept a slight accuracy degradation for a targeted application, we may be able to decrease the model size significantly through some optimization techniques (such as quantization and pruning), and hence reducing processing latency (higher speed-up) and power/energy consumption (higher efficiency)~\cite{bib297}\cite{Ref_Putra_FSpiNN_TCAD20}. 

%%%%%%%%%%%%%%%%%%%%%%%%%%%%%%%%%%%%%%%%
\subsection{Summary of NCL}
\label{Sec_NCLsummary}

The state-of-the-art methods have performed initial studies for enabling NCL (Table~\ref{table_8}). We provide a comparative quantitative analysis (i.e., numerical results) considering design factors (i.e., network complexity) and key evaluation metrics (accuracy, memory footprint, latency, power/energy usage) of the reviewed NCL methods with relevant state-of-the-art DNN-based CL methods (from Section~\ref{sec2.5}), see Table~\ref{comparison-SSN-DNN}. The table is designed to align according to evaluated datasets (i.e., P-MNIST, Split CIFAR-100, Split-MNIST, and ImageNet) and includes the performance metrics (as reported in the respective studies of each method). Several reviewed NCL methods like SpNCN~\cite{bib342}, SOR-SNN~\cite{bib300} and DSD-SNN~\cite{bib304} explicitly compare their method with established DNN-CL baselines, such as EWC~\cite{bib61}, SI~\cite{bib122}, MAS~\cite{bib187}. For more detailed comparisons, we refer to the respective studies of the NCL methods.

We observe that while optimization-based CL methods in DNNs (i.e., GPM~\cite{bib63},~\cite{bib276}, A-GEM~\cite{bib275} and GSS~\cite{bib125}) report high accuracy on standard benchmarks such as P-MNIST, the NCL method HLOP-SNN\cite{bib350} achieves superior accuracy (95.15\%) with the same CL setting and scenario. 
Moreover, the NCL method SOR-SNN~\cite{bib300} achieves competitive accuracy (86.65\%) on CIFAR-100 with significantly lower memory (0.32M params) than DNN-based CL methods, highlighting the energy-efficient scalability of spiking architectures.
On the MNIST, NCL methods such as DSD-SNN~\cite{bib304} and SpikeDyn\cite{bib296} report comparable or better accuracy with drastic reductions in latency, memory, and energy (e.g., 0.2s, <3000KB), demonstrating the suitability of NCL for lightweight and resource-constrained applications.
While accuracy remains lower for both domains on ImageNet, NCL methods like SOR-SNN~\cite{bib300} show promising results in parameter efficiency and deployment feasibility.
Furthermore, the NCL methods with bio-plausible learning rules (e.g., STDP) facilitate unsupervised learning, which is suitable for OCL scenario.
These works mainly employ FC SNN architectures (see Table~\ref{table11}), consisting of input layer and a pair of excitatory-inhibitory layer.  
As the input data grows in complexity, FC SNNs struggle to capture the important hierarchical features without increasing the network size. 
Larger architectures often lead to better accuracy but require larger storage sizes, which may exceed resource constraints. 
This limits the deployable networks on practical hardware platforms, and makes their performance limited to simple datasets (e.g., MNIST) with simple CL scenarios (e.g., Split-MNIST). 
Meanwhile, NCL methods that employ supervised-based learning rules (e.g., SG-based BP) can achieve acceptable performance on complex datasets (e.g., CIFAR-100 and ImageNet).

However, they are not suitable for OCL scenario, as they require costly labeled data and power-hungry training. 
Therefore, alternative methods are required especially for enabling energy-efficient SNNs with OCL capabilities, which are beneficial for many real-world application use-cases. 
Although, we provide the comparison (Table~\ref{comparison-SSN-DNN}), we acknowledge that perfect fairness remains a challenge due to differences in CL scenarios, hyperparameters, underlying model architectures, training paradigms, hardware assumptions, encoding schemes, evaluation protocols, and CL settings (i.e., OCL, supervised or unsupervised). 
We have emphasized these limitations in our discussion (Section-\ref{sec_openchallenges}). Meanwhile, we ensured that our comparisons are as meaningful and aligned, as possible under current benchmarking constraints. Thus, providing a contextual foundation for assessing the relative strengths and limitations of approaches and the reliability of results.

Replay-based CL in SNNs differs from its ANN counterparts in terms of operation as the ANN-based replay methods like iCaRL~\cite{bib135}, DGR~\cite{bib64} rely on storing raw inputs or latent vectors and updating weights using full gradient feedback. In contrast, SNN-based replay must preserve temporally precise spike trains under event-driven constraints, making it dependent on local or SG learning and limited by encoding complexity and latency alignment.
While ANN replay offers high gradient precision and plasticity, SNN replay often relies on approximations like compressed or spike-rate encoded replay to conserve power and memory, which may reduce accuracy. Nonetheless, SNN replay methods such as latent replay or STDP-based approaches are significantly more energy-efficient and well-suited for OCL in low-power environments.
Additionally, spike noise from stochastic firing and temporal jitter, along with hardware-induced variability such as component mismatch and drift, introduces instability in synaptic updates and membrane dynamics. These factors hinder reproducibility and memory consolidation in SNNs, necessitating robustness-aware learning rules like homeostatic STDP or meta-plasticity, such challenges are absent in conventional hardware.

%%%%%%%%%%%%%%%%%%%%%%%%%%%%%%%%%%%%%%%%%%%%%%%%%%%%%%%%%%
%%%%%%%%%%%%%%%%%%%%%%%%%%%%%%%%%%%%%%%%%%%%%%%%%%%%%%%%%%
%%%%%%%%%

\begin{sidewaystable*}
\caption{Quantitative analysis of the state-of-the-art NCL in terms of network architecture, model size, performance, memory footprint, and power/energy consumption.}\label{table11}
\centering
\footnotesize
\begin{tabular*}{\textheight}{@{\extracolsep\fill}llllll}
\hline
\\
\textbf{Work} & \textbf{SNN Architecture} & \textbf{Model Size} & \textbf{Performance} & \textbf{Memory Footprint} & \textbf{Power/Energy Consumption} \\
\\
\hline 
\\
ASP~\cite{bib108} & Hierarchical
SNN & 6400 excitatory & Achieved avg. accuracy of 94.2\% outperforming & n/a & n/a 
\\
&  & neurons  & standard STDP & \\
&  & & & \\
\\
SpikeDyn & 2 FC layers (1 input & 200 and 400 & Improved accuracy by 23\% for the new task \& & $<$3000 KB & Avg. reduction of 57\% \\ 
\cite{bib296} & \& 1 excit. layers) & excit. neurons & 4\% for old tasks than ASP for 200-neuron SNN & & (training) and 51\% \\
& & & & & (inference) than ASP \\
\\

lpSpikeCon & 2 FC layers (1 input & 200 and 400 & No accuracy loss in inference compared to & 8x reduction & n/a \\
\cite{bib297} & \& 1 excit. layers) & excit. neurons & the 32-bit weights baseline (SpikeDyn) & from SpikeDyn & \\
\\

CFN & Multi-FC layers & 400-6400 & Achieved 95.24\% avg. accuracy across MNIST  & n/a & n/a \\
~\cite{bib72} & (1 excit., 1 inhib., \&  & excit. neurons &  digits in Class-IL for 6400-neuron CFN & \\
& 1 dopaminegic layers) & & & \\
\\

SpNCN & 4 FC layers & 3000 neurons- & Normalized accuracy of $0.9653$ (MNIST), & n/a & n/a \\
~\cite{bib342} & & per-layer & $0.9120$ (Not-MNIST), and $0.9995$ (FMNIST) & \\
\\ 

\cite{bib317} & 4 FC layers & 784-403-403-2 & Achieved avg. inference time of 37.3ms & 35.3\% flip-flops, & n/a \\
& & 784-400-400-2 & \& test accuracy of 80.0\% on FPGA &  29.3\% BRAMs &  \\
\\

\cite{bib299} & Multi-FC layers & 2048, 4096, 4096 & Achieved 
avg. test accuracy of 85.44±0.16\% & 5× reduction & n/a\\
& & 2048, 1024 neurons/layer &  for Split-MNIST and 74\% for MNIST-DVS &  \\
\\

DSD-SNN & Multi-CONV \& & 100 \& 500 & Achieved 97.30\%±0.09\% accuracy (34.38\% & n/a & n/a \\
\cite{bib304} & 1 FC layers  & neurons & compression rate) for MNIST in Task-IL, \\
& & & 96.94\%±0.05\% accuracy for N-MNIST, \\
& & & 60.47\% (10 steps) for CIFAR100 in Class-IL \\
\\

SOR-SNN & ResNet18 &  96 hidden & Achieved 86.65\%±0.20 accuracy, \& improved & n/a & 0.32M parameters \\
\cite{bib300} & & layer neurons &  accuracy by 2.20\% from DSD-SNN (CIFAR100) & & (9.35\% of DSD-SNN)  \\
\\

\cite{bib301} & 2 CONV layers & n/a & Achieved 51\% avg. accuracy (40 batches-per-task) & n/a & n/a \\
& & & in Class-IL \& 36\% accuracy with memory replay  \\
& & & (100 batches-per-task) in Task-Free scenario \\
\\

\cite{bib302} & 4 Recurrent-FC & 200-100-50-20 & Achieved Top-1 accuracy of 92.5\% and 92\% in & Two orders of & n/a \\
& layers & neurons & Sample and Class-IL scenarios, respectively & magnitude reduction\\
& & & and 78.4\% in multi Class-IL scenario &  from naive rehearsal \\
& & & & (max. 4\% acc. drop) \\
Replay4NCL & 4 Recurrent-FC & 200-100-50-20 & Achieved Top-1 accuracy of 90.43\% on old knowl- & 20\% reduction in & 4.88x latency improvement \\
~\cite{minhas2025replay4ncl} & layers & neurons & edge, outperforming the state-of-the-art & latent memory usage & and 36.43\% energy reduction  \\
& & & & & than \cite{bib302}\\
\\
\cite{bib339} & 3 CONV layers & 30-250-200 & Achieved $92.0\pm0.1\%$ accuracy for digits in & n/a & n/a \\
& & neurons & joint training, \& 79.7±0.5\% accuracy for letters & \\
\\

HLOP-SNN & FC, ResNet-18, \& & 784-800-10 & Achieved 95.15\% accuracy (PMNIST),  & n/a & n/a \\
\cite{bib350} & 3 CONV layer & neurons for & 78.58\% (CIFAR-100), 63.40\% (miniImageNet),  & \\
& architectures & FC SNN & \& 88.65\% (5-Datasets)\\

\hline 
\end{tabular*}
\end{sidewaystable*}
%%%%%%%%%%%%%%%%%%%%%%%%%%%%%%%%%%%%%%%%%%%%%%%
\section{Real-World Application Use-Cases}\label{sec_usecases}

\begin{figure*}[t]
    \centering
    \includegraphics[width=\textwidth]{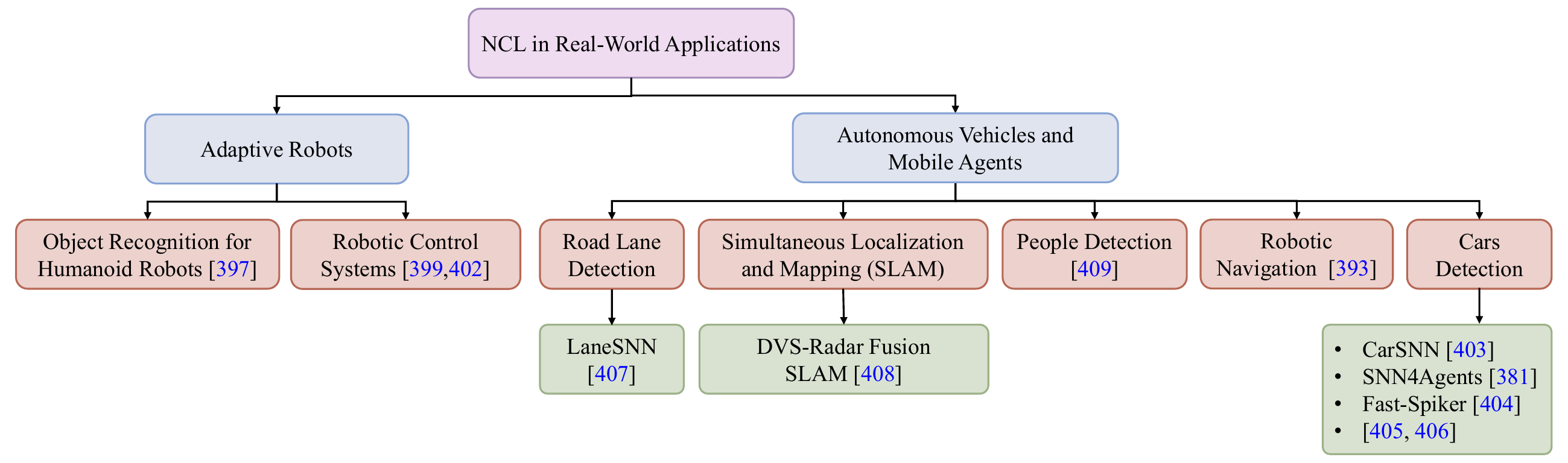} 
    \caption{A taxonomy of real-world application use-cases covered in this survey, that will benefit from bio-plausible SNNs with OCL capabilities. We have highlighted the main categories (blue blocks), with sub-categories (red blocks), and each of their works shown (green blocks).}
    \label{NCL_apps_taxonomy}
\end{figure*}

In this section, we discuss the real-world application use-cases that will benefit from \textit{bio-plausible SNNs with OCL capabilities}, taxonomy shown in Fig.~\ref{NCL_apps_taxonomy}.
Additionally, we also highlight the current state-of-the-art for emerging application use-cases, and hence addressing the key question Q10. A summary of real-world application use-cases that will benefit from SNNs with OCL, is presented in Table~\ref{table_12}. Quantitative analysis (i.e., numerical results) of the case studies covered by these works are reported in Table~\ref{table_13}.

%%%%%%%%%%%%%%%%%%%%%%%%%%%%%%%%%%%%%%%%%%%%%%
\subsection{Adaptive Robots}

%%%%----
\subsubsection{Object Recognition for Humanoid Robots}

\begin{figure}[h]
\centering
\includegraphics[width=0.5\textwidth]{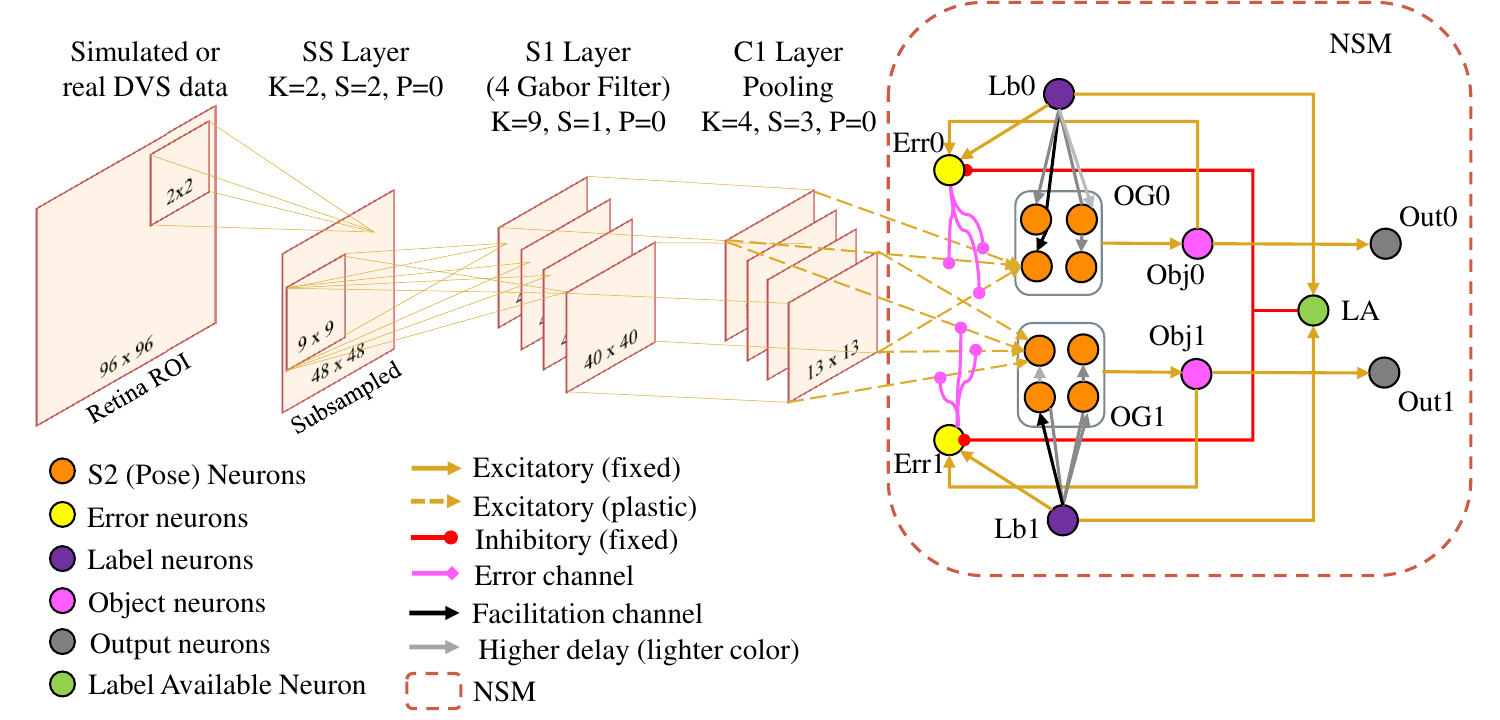}
\caption{The NSM framework: Subsampling layer (\textit{SS}), S1 and C1 perform the visual feature extraction, S2 is the plastic layer (dark blue neurons) that stores visual patterns, Error (\textit{Err}) and label neurons (\textit{Lb}) control inference decision of output neurons (Out); adapted from studies in~\cite{hajizada2022interactive}.}
\label{NSM}
\end{figure}

Adaptive Robots need to identify, classify, and locate objects within an image/video. 
For this, \textit{Neural Engineering Framework (NEF)} is explored for developing \textit{Neural State Machine (NSM)} framework leveraging SNNs, managing the allocation of new neurons (Fig.~\ref{NSM}), where data may be collected on-the-fly without any labels~\cite{hajizada2022interactive}. A case study on the interactive continual object learning of the iCub robot was presented in~\cite{hajizada2022interactive}; see Table~\ref{table_12}. Experiments in the Gazebo robotic simulator~\cite{tikhanoff2008open} mimicking event-based camera conditions, and implementations on  Intel’s Loihi neuromorphic chip~\cite{Ref_Davies_Loihi_MM18}, demonstrated $96.55\pm2.02\%$ accuracy in learning object representations within three epochs, adapting efficiently to new objects~\cite{hajizada2022interactive}; see Table~\ref{table_13}. CL assessments showed rare confusions ($<20\%$) after the first epoch. The model allocated neurons dynamically based on object complexity, optimizing resource use. The continual classifier was benchmarked against conventional online classifiers on a CPU, showing up to 300× better energy efficiency for learning and 150× for inference~\cite{hajizada2022interactive}. These findings highlight SNNs' potential for real-world deployment in embedded AI systems, with future extensions to physical robotic platforms.
However, it relies on supervised settings which require costly labeled data, power-hungry training, and long training time to update its knowledge, thus it is not suitable for OCL scenario.

%%%%----
\subsubsection{Robotic Control Systems}

\begin{figure}[h]
    \centering
    \includegraphics[width=0.5\textwidth]{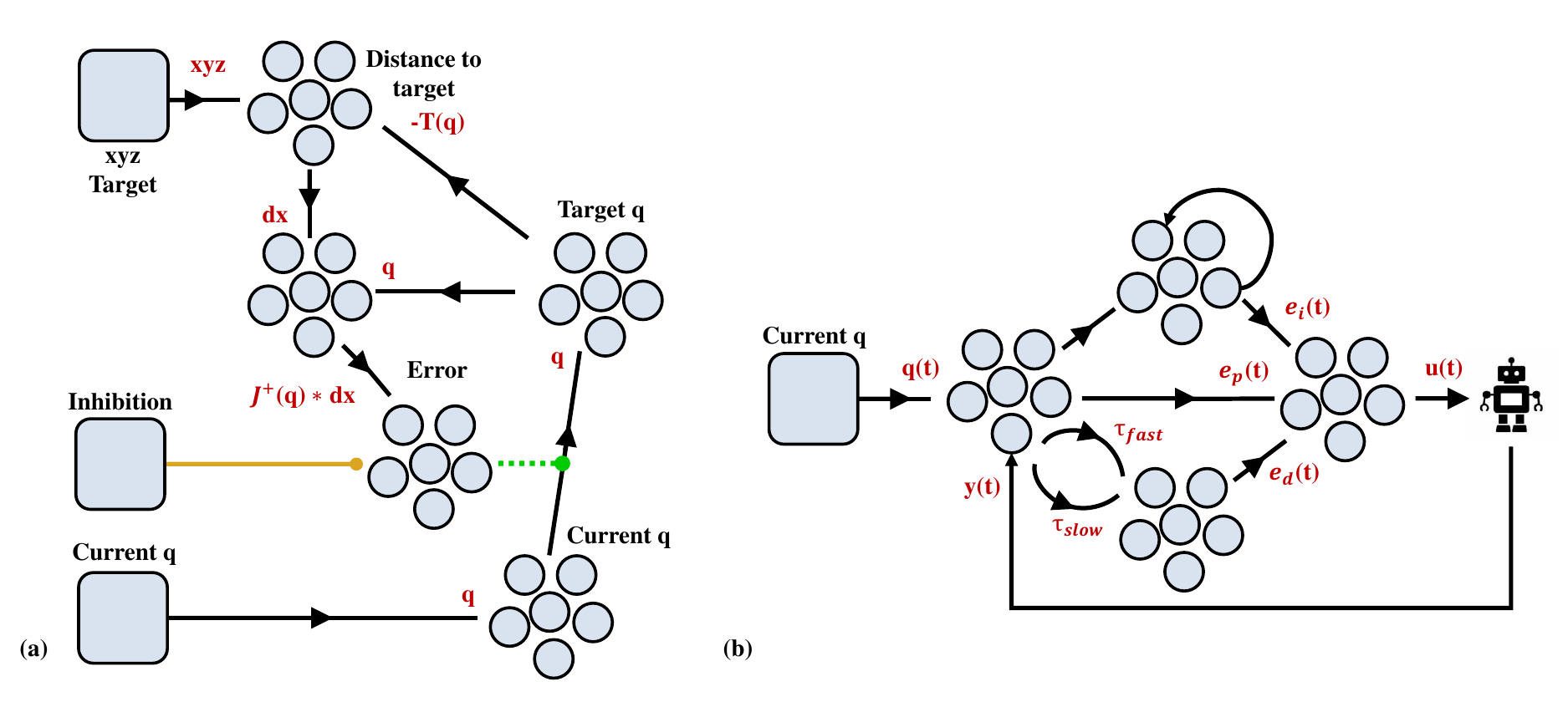}
    \caption{(a) A model for neuromorphic IK with online learning, and (b) Neuromorphic PID controller in action; adapted from studies in~\cite{zaidel2021neuromorphic}.}
    \label{IK}
\end{figure}

The study~\cite{zaidel2021neuromorphic} explored NEF for developing neuromorphic algorithms for \textit{Inverse Kinematics (IK)} and \textit{Proportional-Integral-Derivative (PID) control} systems, supporting online learning of control signals, using \textit{Prescribed Error Sensitivity (PES)}.
Fig.~\ref{IK}(a) shows the model schematic for neuromorphic IK with online learning, whose target is to efficiently compute the joint angles required to place the robots' end-effector at the desired position.
Meanwhile, the PID controller aims to ensure that the robots’ actuators accurately follow a desired trajectory.
Hence, it needs to continuously adjust the actuators to minimize the error in position, ensuring accurate motion control. 
To achieve this, the PID controller has been implemented using spiking neurons to represent the error signals and components; see Fig.~\ref{IK}(b). This work presented a case study on the control of a 6-degrees of freedom robotic arm; see Table~\ref{table_12}. The implementation was based on Nengo~\cite{bekolay2014nengo,lin2018programming} a Python
based neural compiler, that translates high-level descriptions to low-level neural models and Intel Loihi neuromorphic chip~\cite{Ref_Davies_Loihi_MM18}, offering
high performing and energy-efficient neuromorphic control; see Table~\ref{table_13}. However, it relies on supervised settings which require costly labeled data, power-hungry training, and long training time to update its knowledge and thus is not suitable for the OCL scenario.
Similarly, a recent work in~\cite{marrero2024novel} used NEF within the Nengo simulator and MATLAB R2022b to implement a SNN-based PID controller for robotic arm trajectory tracking. The controller was evaluated on a case study of simulated 3-DoF robotic arm; see Table~\ref{table_12}. It reported  achieving a 6\% improvement in the Integral of Time-weighted Absolute Error (ITAE) and a 30\% reduction in Root Mean Square Error (RMSE) compared to the conventional PID and fuzzy controllers; see Table~\ref{table_13}.

%%%%%%%%%%%%%%%%%%%%%%%%%%%%%%%%%%%%%%%
\subsection{Autonomous Vehicles and Mobile Agents}

CL in autonomous vehicles and mobile agents is crucial to enable continuous adaptation to the changes in environments (e.g., road and weather) and personalize the preferences. 
Hence, they are typically equipped with sensors and cameras for continuous data collection during operation. 
Moreover, they usually rely on limited battery, and often face unlabeled data and constrained resources. 
Therefore, their use-cases can benefit from SNNs with OCL capabilities.

%%%%%%%%
\subsubsection{Cars Detection}

\begin{figure}[h]
\centering
\includegraphics[width=0.5\textwidth]{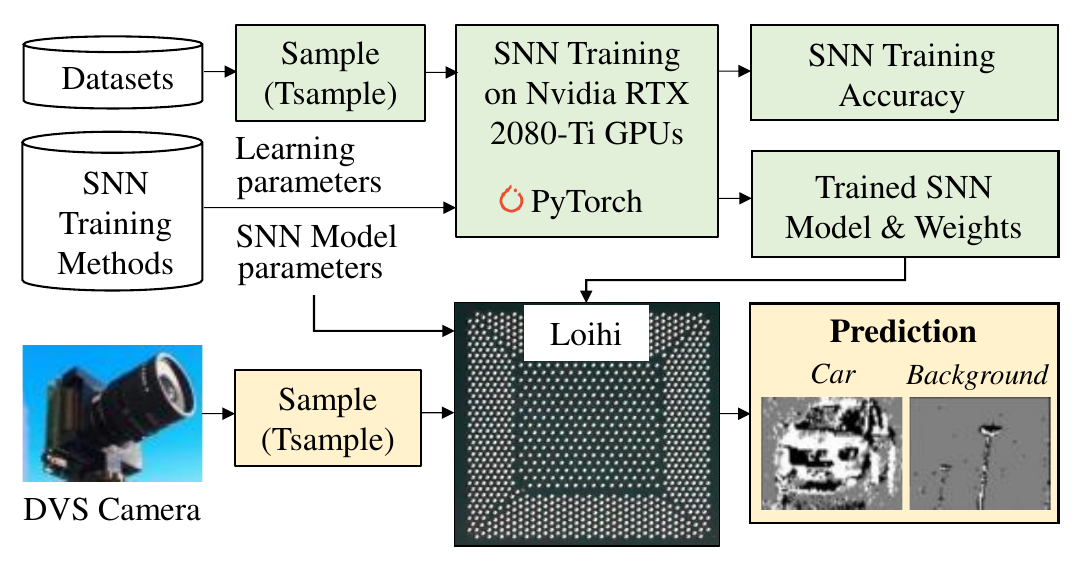}
\caption{Setup and tool-flow of CarSNN; adapted from studies in~\cite{viale2021carsnn}.}
\label{carsnn}
\end{figure}

Several works have been developed for low-power autonomous vehicles using SNNs, including studies of cars detection in~\textbf{CarSNN}~\cite{viale2021carsnn}, \textbf{SNN4Agents}~\cite{Ref_Putra_SNN4Agents_FROBT24}, \textbf{FastSpiker}~\cite{bano2024fastspiker} and \cite{cordone2022object, bano2024methodology}. 
CarSNN and SNN4Agents used an \textit{attention window mechanism} (i.e., focusing on regions with the highest event density) to process event-based input samples from the N-CARS dataset~\cite{Ref_Sironi_HATS_CVPR18}; see Fig.~\ref{carsnn}. For the cars vs. background classification the SNN model was implemented on the Intel Loihi neuromorphic chip~\cite{Ref_Davies_Loihi_MM18} (see Table~\ref{table_12}). The neuromorphic hardware implementation had maximum 0.72 ms of latency for every sample, and consumed only 310 mW power with an accuracy of 83\%~\cite{viale2021carsnn} (see Table~\ref{table_13}).
However, the state-of-the-art~\cite{viale2021carsnn, Ref_Putra_SNN4Agents_FROBT24} require a long training time~\cite{bano2024fastspiker}. 
Therefore, \textbf{FastSpiker} proposed a methodology to accelerate SNN training while maintaining accuracy through \textit{learning rate enhancements}~\cite{bano2024fastspiker}. 
However, these works considered supervised settings which require costly labeled data, power-hungry training, and long training time to update its knowledge, hence they are not suitable for OCL scenario.

%%%%%%%%
\subsubsection{Road Lane Detection} 

\begin{figure}[h]
\centering
\includegraphics[width=0.5\textwidth]{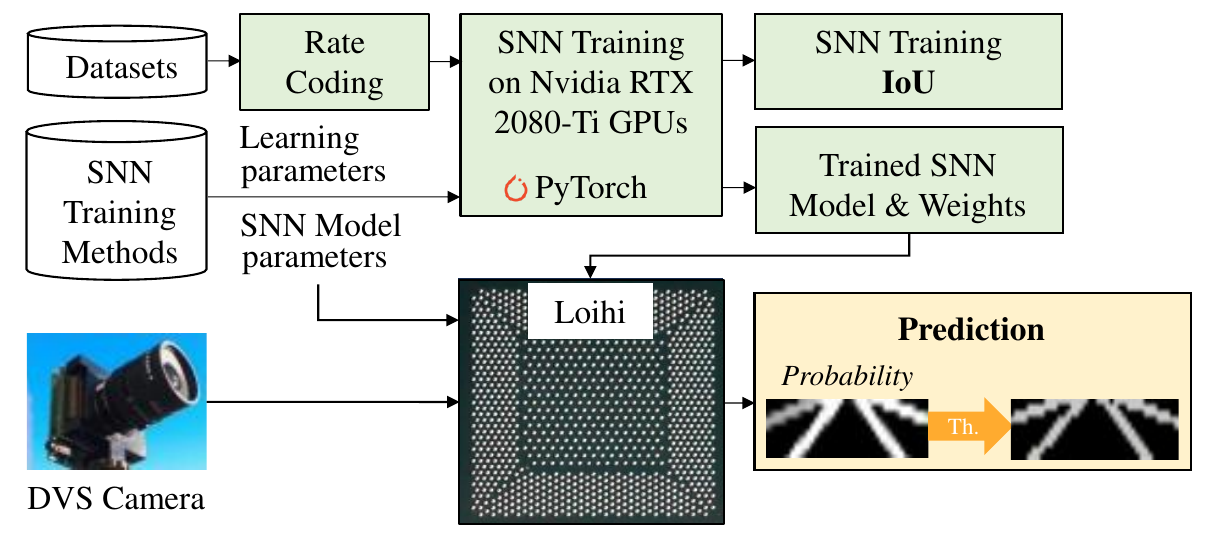}
\caption{Setup and tool-flow of LaneSNNs; adapted from studies in~\cite{viale2022lanesnns}.}
\label{LaneSNN}
\end{figure}

Other SNN works for low-power autonomous vehicles include the road line detection.
For instance, \textbf{LaneSNN}~\cite{viale2022lanesnns} addressed the imbalance between lane and background classes by employing a novel loss function combining \textit{Weighted Binary Cross Entropy (WCE)} and MSE; see Fig.~\ref{LaneSNN}. For detecting the lanes marked on the streets using the event-based camera input, the SNNs training and implementation was based on PyTorch library~\cite{paszke2019pytorch} and the Intel Loihi neuromorphic chip~\cite{Ref_Davies_Loihi_MM18}; see Table~\ref{table_12}. Evaluations demonstrated maximum latency of less than 8 ms, power consumption of about 1 W during the classification of a single image, and online IoU equal to 0.623, thereby making it superior to the state-of-the-art techniques like LaneNet and RefineNet in terms of performance and power efficiency; see Table~\ref{table_13}.
However, it still considered supervised settings which require costly labeled data, power-hungry training, and long training time to update its knowledge, hence it is not suitable for OCL scenario.

%%%%%%%%
\subsubsection{Simultaneous Localization and Mapping (SLAM)}

\begin{figure}[t]
\centering
\includegraphics[width=0.44\textwidth]{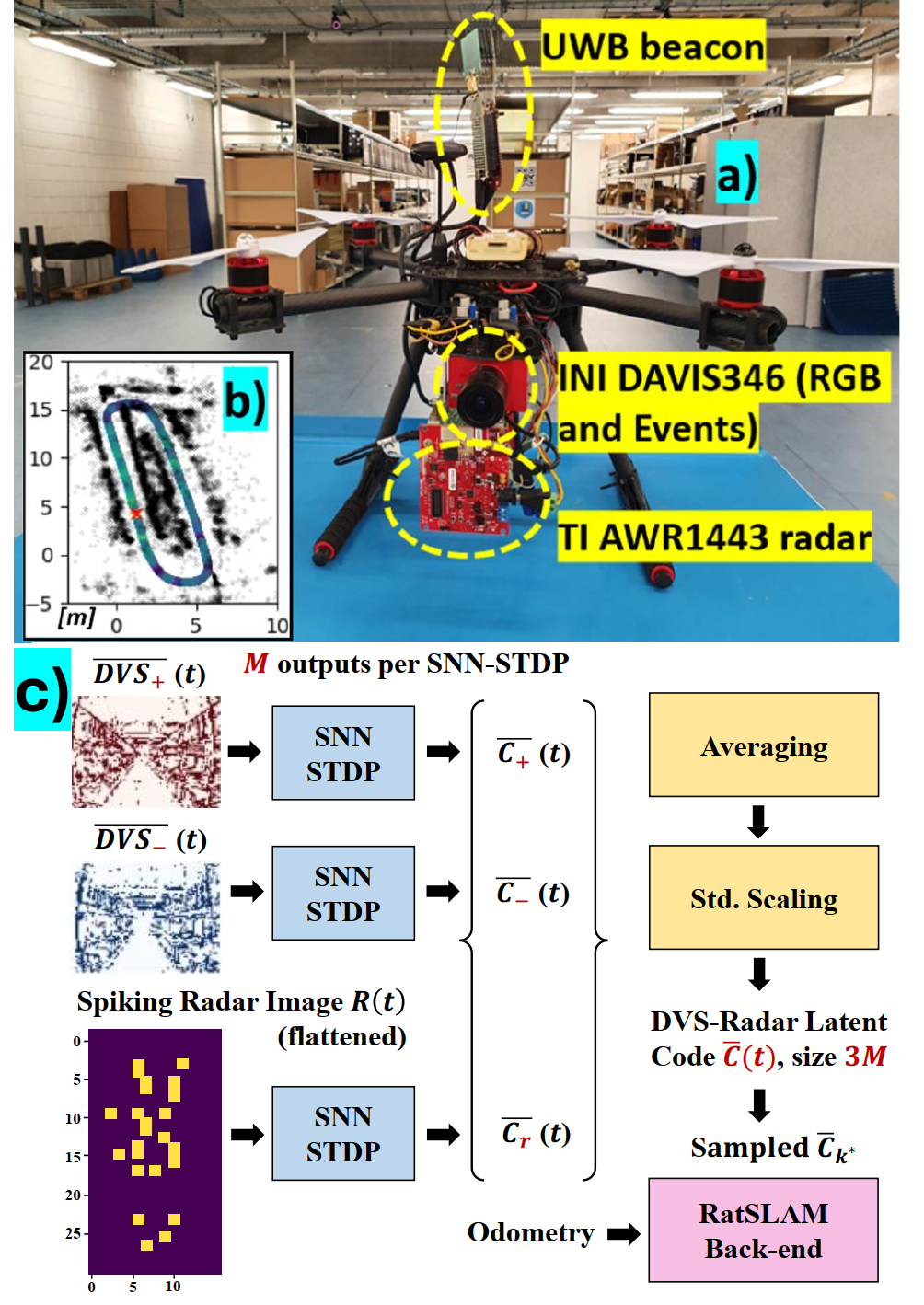}
\caption{a) Setup on a drone in a UWB-equipped warehouse for ground truth localization, b) Event and radar data are processed by an STDP-trained SNN for loop closure detection in SLAM, with radar also modeling obstacles (black dots), and c) DVS-Radar Fusion for SNN-STDP based CL SLAM system~\cite{bib354}.}
\label{dvs-radar}
\end{figure}

SLAM is important for an autonomous vehicle/agent for constructing a map while simultaneously keeping track of its location in an unknown environment. 
The \textbf{DVS-Radar Fusion SLAM}~\cite{bib354} is the first SNN-based method for enabling CL-based SLAM using drones (Fig.~\ref{dvs-radar}), integrating DVS and a \textit{Frequency Modulated Continuous Wave (FMCW)} radar to encode sensory data on-the-fly into feature descriptors. 
To provide drones’ velocity information and aid in obstacle detection, a \textit{radar-gyroscope odometry} method is used for accurate navigation and mapping. 
SNN output is utilized as feature descriptors for \textit{loop closure detection} that are fed to a RatSLAM back-end. 
This process helps in identifying when the drone revisits a previously mapped area. 
The obstacle modelling is achieved using radar data by detecting objects and integrating this information into its SLAM framework. 
DVS-Radar fusion setup outperformed RGB-based methods achieving lower $MAE_L$ and $MAE_M$ values of 0.51 \& 0.17 for drone flight sequence 1, and 0.81 \& 0.45 for sequence 2; see Table~\ref{table_12} and~\ref{table_13}. This work showed the potential of implementing OCL on SNN-based mobile agents.

\begin{table*}[h!]
\caption{Summary of recent progress in real-world NCL application use-cases that will benefit from SNNs with OCL capabilities, with their implementation details and case studies.} 
\label{table_12}
\centering
\footnotesize
\setlength{\tabcolsep}{1pt}
\begin{tabular*}{\textwidth}{@{\extracolsep\fill}llllllllccc}
% \hline %
\cmidrule{1-10}
\textbf{Application} & \textbf{Use-} & \textbf{Work} &  \textbf{Case} & \textbf{CL} & \textbf{Dataset} & \textbf{CL} & \textbf{Learning} & \textbf{Neuron} & \textbf{HW} \\
\textbf{Domains} & \textbf{cases} & \textbf{(Year)} &  \textbf{Study} & \textbf{Setting} & & \textbf{Approach} & \textbf{Rule} & \textbf{Model} &  \\
\cmidrule{1-10}
& Object & \cite{hajizada2022interactive} & Interactive & Supervised & Custom & Representation- & Modified & LIF & Loihi \\
& Recognition & (2022) & continual object & & DVS dataset & based & Hebbian & & & \\
Adaptive & & & learning in iCub robot \\
% & &&&&&&&& \\  
\cline{2-10}  
Robots & & & Control of  & & Custom & & & & \\
& Robotic Arm & \cite{zaidel2021neuromorphic} (2021) & 6-DoF, &  Supervised & dataset from & Representation- & PES & LIF & Loihi \\
& Control &  \cite{marrero2024novel} (2024) & 3-DoF trajectory & & robotic arms' & based & & & \\
& & & tracking & & operations & & & & \\
\cmidrule{1-10}
& Cars & \cite{viale2021carsnn} (2021) & Cars vs. & & & STBP & & LIF & Loihi\\ 
& Detection & \cite{cordone2022object} (2022) & background  & Supervised & N-CARS & - & SG & LIF & GPU \\
& & \cite{Ref_Putra_SNN4Agents_FROBT24} (2024) & classification & & & STBP & & LIF & GPU \\
\cmidrule{2-10}
& Road Lane & \cite{viale2022lanesnns} & Detection of  &  &  &  & & &   \\
Autonomous & Detection & (2022) & lanes marked & Supervised & DET \cite{cheng2019det}  & - & STBP & LIF & Loihi\\
Vehicles/ & & & on the streets \\ 
\cline{2-10}
Agents & & \cite{bib354} & CL SLAM & & Real-time & Representation- &  &  &  \\
&  SLAM & (2023) & fusing DVS \& & Unsupervised  & data from & based & STDP & LIF & n/a  \\ 
& & & radar on drone & & sensors  \\
\cmidrule{2-10}
& People & \cite{bib357} & CL of people & & KUL-&  &  &  & ReckOn \\
& Detection & (2024) & detection from & Semi-supervised &  UAVSAFE & Representation- & STDP & LIF & $chip^*$ \\
& & & event-camera & &  \cite{safa2021fail} & based &  & &  \cite{frenkel2022reckon} \\
&&& mounted on drone &&&&&& \\
\cmidrule{2-10}
& Robotic  & ~\cite{chen2024paiboard} & Tracking \& & & CIFAR-10, \\
& Navigation & (2024) & obstacle avoidance & Hybrid &  self-built & - & - & LIF & PAIBoard  \\
& & & in robot dog~\cite{unitree_aliengo} & & dataset \\
% \hline
\cmidrule{1-10}
\end{tabular*}
\begin{tablenotes} \footnotesize
\item \textit{*} Potential future work.
\end{tablenotes}
\end{table*}
%%%

\begin{table*}[h!]
\caption{Quantitative analysis of case studies covered by the reviewed works in Section IV.}
\label{table_13}
\centering
\footnotesize
\setlength{\tabcolsep}{1pt}
\begin{tabular*}{\textwidth}{@{\extracolsep\fill}clcccc}

\cmidrule{1-6}
\textbf{Work} & \textbf{Performance} & \textbf{OCL} & \textbf{Latency} & \textbf{Memory} & \textbf{Power/Energy} \\
& & \textbf{Capability} & &  \textbf{Footprint} & \textbf{Savings} \\
% \hline
\cmidrule{1-6}
\cite{hajizada2022interactive} & 96.55±2.02\% new object accuracy within  & $\times$ & n/a & n/a & 300× better for training \\
& three epochs, <20\% confusions after first epoch &  & & & \& 150× for inference\\
\cmidrule{1-6}
\cite{zaidel2021neuromorphic} & Loihi accuracy
 outperformed the simulation  & $\times$ & Loihi implementation converged faster & n/a & Improved \\
& across different learning rates & & to the target than the simulated model \\
\cmidrule{1-6}
~\cite{marrero2024novel} & 6\% improvement in ITAE and 30\% reduction  & $\times$ & n/a & n/a & n/a \\
& in RMSE compared to traditional PID controller \\
\cmidrule{1-6}
~\cite{viale2021carsnn} & 83\% accuracy & $\times$ & 0.72 ms & n/a &  310 mW \\
\cmidrule{1-6}
~\cite{viale2022lanesnns} & online IoU 0.623 &  $\times$  & 8 ms & n/a & $1 W^1$ \\
\cmidrule{1-6}
~\cite{bib354} & $MAE_L$ of 0.51 \& $MAE_M$ of 0.17 for drone flight &  &  \\
& sequence 1, $MAE_L$ of 0.81 \& $MAE_M$ of 0.45 for & \checkmark & n/a & n/a & n/a\\
&  sequence 2 \\
\cmidrule{1-6}
\cite{bib357} & Peak $F_1$ score 19\% higher compared to same- & \checkmark & n/a & 514 kB & n/a \\
& size CNN  \\
\cmidrule{1-6}
\cite{chen2024paiboard} & 90.2\% accuracy & \checkmark & n/a & n/a & $energy^2$ $791 FPS/W$, \\
& & & & & $power^3$ $12.8 W$ \\
\cmidrule{1-6}
\end{tabular*}
\begin{tablenotes} \footnotesize
\item \textsuperscript{1} For single image classification; \textsuperscript{2} For image classification tasks; \textsuperscript{3} For tracking and obstacle avoidance task.
\end{tablenotes}
\end{table*}

%%%%%%%%
\subsubsection{People Detection} 

\begin{figure}[t]
\vspace{-0.5cm}
\centering
\includegraphics[width=0.45\textwidth]{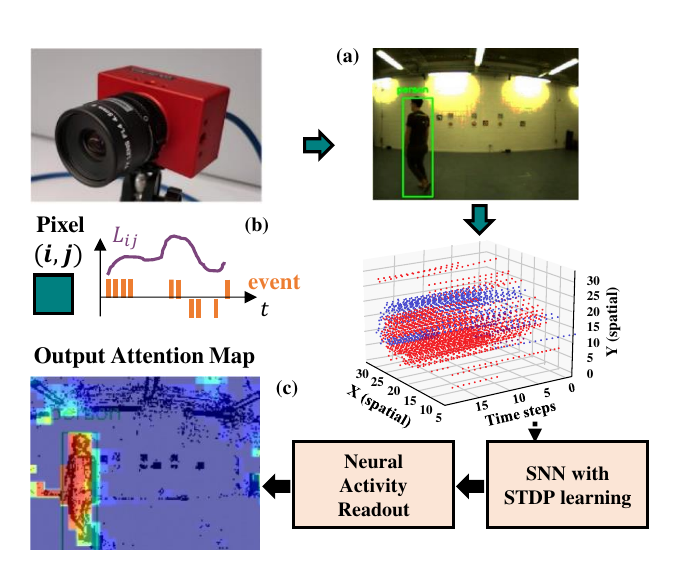}
\caption{Attention maps based CL system for people detection; adapted from studies in~\cite{bib357}. 
(a) The DVS produces a stream of spikes over time and space. (b) A spike is emitted when the change in light intensity $L_{ij}$ at pixel $(i,j)$ exceeds a certain threshold. The spike is positive if the change $\Delta L_{ij}>0$ and negative otherwise. 
(c) The SNN-STDP followed by a readout of neural activity is used to investigate the continuous development of attention-based perception.}
\label{pd}
\end{figure}

A CL-based people detection framework has been developed in~\cite{bib357}; see Fig.~\ref{pd}. 
It employs attention maps to enable network adaptation on walking-people detection in dynamic environments, thus avoiding collision.
To create attention maps from DVS data, the network is integrated with a readout mechanism. 
It also generalizes the unsupervised STDP from~\cite{safa2021new} to a semi-supervised learning case with \textit{teacher signals (TSs)}. 
To control overfitting, an anti-Hebbian rule (i.e., negative STDP) is applied when person is absent, while positive STDP is used when a person is detected. 
The empirical results indicated that this SNN-STDP system achieved a peak $F_1$ score 19\% higher than a comparable same-size CNN processing DVS frames, demonstrating its effectiveness in dynamic human detection task; see Table~\ref{table_12} and~\ref{table_13}. Future works include: a) investigating multi-object contexts with multiple walking people instead of a single person, b) implementing proposed system on neuromorphic hardware ReckOn chip~\cite{frenkel2022reckon}, which supports local learning, multilayer learning, and SNN ensembles with non-spiking readouts, making it suitable for this SNN-STDP-readout approach.
This work further showed the potential of implementing OCL on SNN-based systems.

\begin{figure}[t]
\centering
\includegraphics[width=0.5\textwidth]{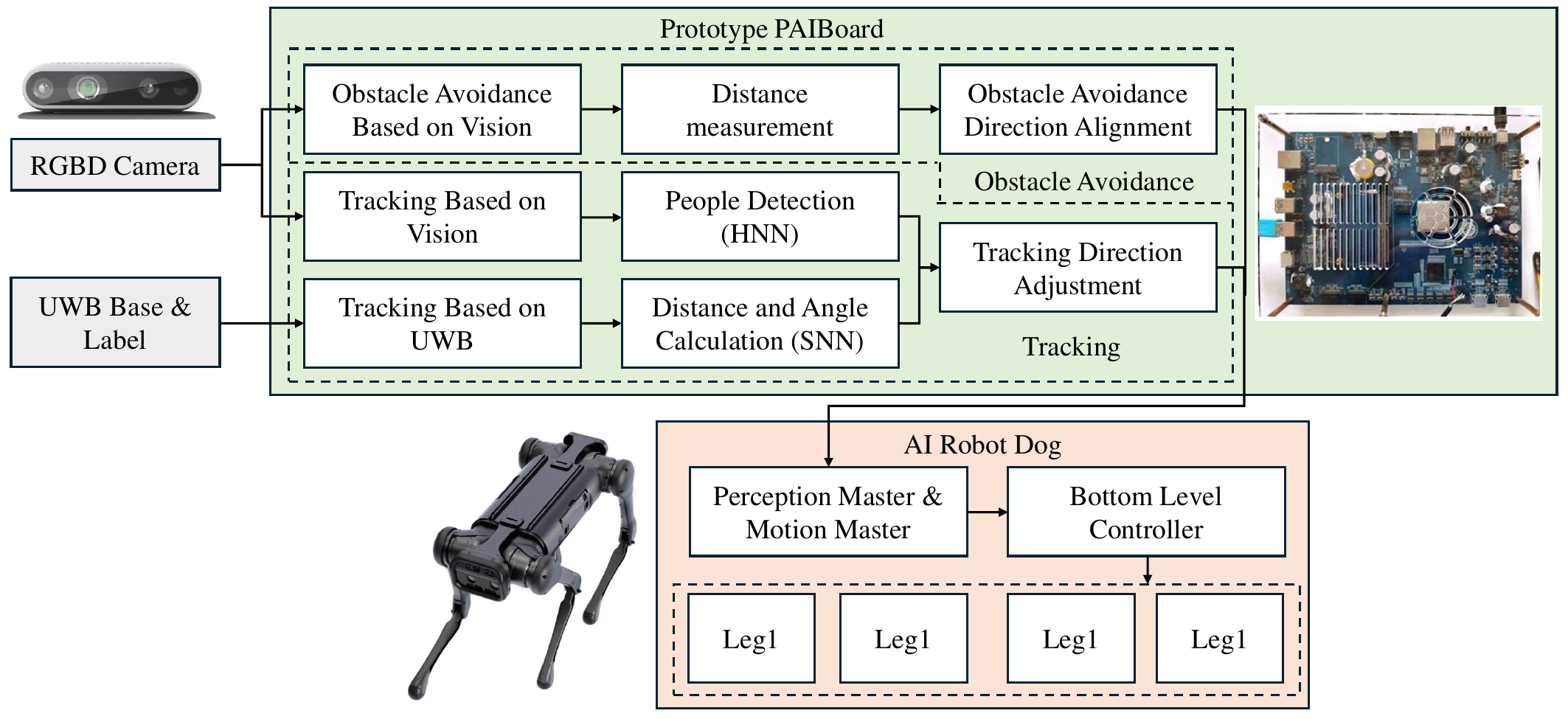}
\caption{Deployment of PAIBoard in a quadruped robot for autonomous navigation, where sensor fusion (UWB + vision) and hybrid SNNs enable robust real-time tracking and obstacle avoidance~\cite{chen2024paiboard}.}
\label{paiboard}
\end{figure}

\subsubsection{Robotic Navigation}

Recent study~\cite{chen2024paiboard} demonstrated the use of PAIBoard neuromorphic computing platform in a quadruped robot~\cite{unitree_aliengo} performing real-world navigation tasks such as tracking (i.e., via UWB and visual fusion) and obstacle avoidance (i.e., via depth perception) based on vision-based NN processing, as shown in Fig.~\ref{paiboard}. 
The system leveraged SNNs for adaptive, low-power control, underscoring the viability of hybrid SNN-ANN (HNN) platforms in autonomous sensorimotor tasks. 
PAIBoard achieved 90.2\% accuracy on CIFAR-10 using fewer neurons and cores than TrueNorth and Tianjic, highlighting efficient resource usage. 
It also demonstrated $791 
FPS/W$ energy efficiency for image classification and consumed $12.8 W$ during robot tracking and obstacle avoidance task; see Table~\ref{table_12} and~\ref{table_13}. 
The platform is designed to handle such complex tasks, implies a potential for OCL. 

%%%%%%%%%%%%%%%%%%%%%%%%%%%%%%%%%%%%%%%%%%%%%%%%%%%%%%%%%%%
%%%%%%%%%%%%%%%%%%%%%%%%%%%%%% 

\begin{table*}
\caption{Summary of open research challenges with their brief description, identified research gaps, and future directions in NCL.}
\label{table_14}
\centering
\footnotesize
\begin{tabular}{p{1.7cm}p{3.4cm}p{4cm}p{7.1cm}}
\cmidrule{1-4}%
\textbf{Challenge} & \textbf{Description} & \textbf{Identified Research Gaps} & \textbf{Future Directions}  \\
\cmidrule{1-4}%  
Adaptive knowledge retention and transfer & Includes issues like overlapping neural representations, retaining task-specific knowledge with memory efficiency. & - Insufficient exploration of NCL methods considering task-similarity for BWT. \newline - Limited exploration of advanced compression techniques in replay-based NCL methods. & - Developing methods with enhanced BWT leveraging similarities between tasks, while avoiding interference. \newline - Exploring efficient memory-augmented techniques for SNNs that preserve important synapses. \newline - Exploring dynamic allocation of dendritic segments only when needed to avoid overfitting. \newline - Designing adaptive algorithms that automatically determine the number of subspace neurons based on task complexity.\\
\cmidrule{1-4}%
Adaptation of DNN-based CL methods in SNN domain (i.e., NCL) and Hybrid SNN-DNN Approaches & The adaption of current DNN-based CL methods in SNN domain for NCL developments may lead to effective solutions. \newline - The integration of hybrid SNN-DNN models, leveraging the strengths of both architectures to improve CL performance. & - Limited adaptation of different CL-based methods (e.g., regularization) in SNNs, particularly for OCL scenarios. \newline - Limited exploration of hybrid SNN-DNN models for CL. \newline - Limited exploration of knowledge distillation techniques like teacher-student in SNN-based CL models. & - Exploring adaptions of CL-based methods (e.g., regularization) and their impact on SNNs trained with local learning rules in dynamic environments. \newline - Investigating the optimal memory size for replay-based methods in SNNs. \newline - Hybrid approaches where DNNs handle feature extraction while SNNs process temporal or event-driven information could enhance CL in real-world applications. \newline - Exploring knowledge distillation to transfer knowledge from deep networks to SNNs, enabling efficient CL while preserving performance. \\
\cmidrule{1-4}%
Adaptation to dynamic input patterns & Changes in input patterns may occur during run time, thereby requiring OCL capabilities to handle dynamic variability of input. & - SNN learning quality in OCL scenarios is still a developing area. - Limited development of adaptive learning rates that are responsive to temporal correlations in spiking patterns. & - Investigating SNN architectures that effectively perform weight updates to maintain performance (e.g., accuracy).  \newline - Investigating adaptive learning and weight decay rate strategies to improve the flexibility of neuron dynamics. \newline - Developing methods for efficient clustering and representation learning. \newline - Developing adaptive algorithms to explore various configurations of dendritic segments to optimize task-specific adaptation. \\
\cmidrule{1-4}%
Balancing CL desiderata & To meet multiple, often conflicting, requirements of CL. & Balancing CL desiderata remains a key optimization goal for NCL algorithms. & Developing methods that balance and prioritize CL desiderata trade-offs based on the specific application needs. \\
\cmidrule{1-4}%
Input noise robustness & Real-world input data is often noisy, with more complex, non-linear patterns than AWGN. & Limited exploration of diverse noise types and intensities for ensuring robustness of NCL methods against noise for OCL scenarios. & - Expanding the scope of evaluation to include diverse noise types and intensities. \newline - Designing algorithms that automatically filter irrelevant information while emphasizing critical features, thus improving overall performance. \\
\cmidrule{1-4}%
Data representation & Most of the current NCL methods use a single encoding technique, i.e., rate coding, which neglects the temporal structure of spike trains, limiting their ability to fully exploit temporal information. & - Limited exploration of NCL methods using temporal encoding with high information density. \newline  - Lack of studies for exploring encoding methods that utilize the strengths of multiple encoding methods. & 
- Incorporating time-encoded signals into SNNs to enhance their learning from temporal data (e.g., TTFS encoding) to improve the network's ability to handle time-dependent patterns. \newline - Developing SNNs that integrate hybrid encoding techniques for optimization of task performance, latency and energy efficiency. \\
\cmidrule{1-4}%
Scalability & Current CL methods incur significant memory and power/energy overheads when the networks get bigger, thereby posing scalability issue. & - Limited studies for realizing dynamic allocation of neurons without significant computational and memory overheads. \newline  - Lack of effective and biologically inspired synaptic reorganization (pruning and growing) mechanisms for SNNs. & - Investigating the integration of more complex neuron models (e.g., Hodgkin-Huxley model), allowing for a richer representation of neuronal activity and potentially improving the model's predictive capabilities and scalability. \newline - Exploring techniques to dynamically allocate neurons for new tasks while pruning redundant ones from previous tasks.\newline - Investigating hierarchical or modular approaches for SNN design to enable scalable architectures for CL. \newline - Investigating techniques for dynamic structural plasticity, inspired by the brain's ability to reorganize neural connections. \\
\cmidrule{1-4}%
Learning rule developments & The existing CL methods often rely on  non-local learning rules and primarily use a single learning paradigm, either supervised or unsupervised, limiting performance and adaptability. & - Enhancements of bio-plausible  learning rules (e.g., STDP) for OCL scenarios remain under-explored. \newline - Limited studies in combining supervised and unsupervised settings for NCL. &  Developing advanced local learning rules, such as adaptive variations of STDP, can help SNN models achieve NCL capabilities, especially in OCL scenarios. \newline - Developing hybrid approaches for NCL that combine different learning paradigms for training SNNs (i.e., supervised and unsupervised settings), considering OCL scenarios.  \\
\cmidrule{1-4}%
\end{tabular}
\end{table*}

\addtocounter{table}{-1}
\begin{table*}
\caption{Continued}
\label{table_14}
\centering
\footnotesize
\begin{tabular}{p{1.7cm}p{3.4cm}p{4cm}p{7.1cm}}
\cmidrule{1-4}%
\textbf{Challenge} & \textbf{Description} & \textbf{Identified Research Gaps} & \textbf{Future Directions}  \\
\cmidrule{1-4}%
Generalization capability & Current methods are limited in their adaptability to drastically changing environments or task-specific nuances. & Ensuring NCL methods function effectively across diverse operational settings and tasks is a developing area. & - Developing adaptive NCL methods that can dynamically adjust to diverse tasks and environmental context, such as adapting to changes from indoor to outdoor conditions. \newline - Expanding the evaluation of NCL methods to diverse datasets like DVS Gesture to improve generalization for complex tasks.\\
\cmidrule{1-4}%
Evaluation datasets and benchmarks & Datasets should be derived or generated to leverage unique characteristics of event-based data, which properly represent the original data. & - Research on neuromorphic datasets for benchmarking NCL methods is still developing. \newline - Inconsistent data pre-processing  (e.g., time resolution), making it difficult to compare results fairly. \newline - Data transformation may fail to capture temporal information, thus limiting SNN capabilities. \newline - Current datasets are mostly small-scale and narrowly focused, limiting the generalizability and benchmarking for SNNs. & - Developing standardized pre-processing rules to ensure fairness and consistency in dataset comparisons.  \newline - Measuring and optimizing energy savings for showing the NC potential in practical applications. \newline - Constructing datasets emphasizing temporal dynamics, leveraging event-driven and spike-based data generation. \newline - Creating larger-scale, spatial-temporal event-based  datasets for diverse tasks to enhance benchmarking and real-world applicability. \newline - Task-specific performance metrics are suggested, such as accuracy for classification, MAE for regression, memory ratio and power/energy overhead for scalability,  and adaptability scores for functionality.\\ 
\cmidrule{1-4}%
Hardware deployments & - Implementing NCL methods on neuromorphic processors can maximize the efficiency benefits via massively-parallel processing and inherent low-power. \newline - NCL methods involve frequently updating parameters, like neuron thresholds and decay rates along with the adjustments of membrane potentials and synaptic weights.& - The transition from algorithmic level to practical implementation on hardware remains under-explored. \newline  - Limited exploration of in-memory computing paradigms for handling frequent updates efficiently. \newline - Cross-integration of diverse paradigms (e.g., DNNs and SNNs) on neuromorphic chips remains under-explored. &  - Developing more efficient hardware designs and leverage SNN mapping for supporting NCL methods. \newline - Introducing approximations in neuron parameters and synaptic weights to meet hardware constraints. \newline- Exploring the potential of beyond-CMOS technologies (e.g., NVM) for more energy-efficient implementation. \newline - Developing CIM systems with high-performance interfaces to support quick and frequent updates, minimizing latency and energy overheads. \newline - Developing processors for supporting the combination of unsupervised local learning with supervised global learning (e.g., combining different conventional \& neuromorphic chips). \\
\cmidrule{1-4}%
\end{tabular}
\end{table*}

%%%%%%%%%%%%%%%%%%%%%%%%%%%%%%%%%%%%%%%%%%%%%%%%%%%%%%%%%%%
%%%%%%%%%%%%%%%%%%%%%%%%%%%%%%%%%%%%%%%%%%%%%%%%%%%%%%%%%%%
\section{Open Research Challenges} \label{sec_openchallenges}

Due to the huge potentials and benefits of energy-efficient SNNs with OCL capabilities, continuous advancements in the NCL field are expected in the future. 
In this section, we identify open research challenges in NCL, highlight the gaps between current research efforts, and propose future research directions; see the summary presented in Table~\ref{table_14}.
This discussion also addresses the key question Q11.

\smallskip
\textbf{Adaptive Knowledge Retention and Transfer}: 
Recent NCL methods mitigate CF through adaptive learning rates and weight decay coupled with STDP~\cite{bib296}, parameter adjustments with weight potentiation and depression~\cite{bib297}, error-corrected predictions, multi-layer architectures for maintaining distinct task representations~\cite{bib342}, dendritic-dependent spike time delays and context-dependent gating system~\cite{bib317}, as well as compressed LR of learned data~\cite{bib302}. 
Moreover, Hebbian learning and lateral connections, to project neuronal activity into an orthogonal subspace~\cite{bib350} have been utilized to preserve previous knowledge. 
Despite these advancements, we identify the existing research gaps and potential future directions, as follows.
\begin{enumerate}
    \item Exploration of task-similarity-based BWT is limited, thus the challenge of interference in overlapping classes with high feature similarity remains open. 
    To address this, improving task-similarity-based BWT by exploring advanced feature extraction techniques to enhance class separability and reduce interference. 
    \item In replay-based methods, increasing number of classes may result in substantial growth in storage. Meanwhile, lossy compression techniques can cause information loss that degrade the performance on previously learned task. 
    Toward this, advanced compression techniques that preserve temporal information with efficient memory-augmentation methods may preserve synaptic connections, thus enhancing performance. 
    \item To avoid overfitting, dendritic segments can be dynamically allocated.
    \item Finally, developing adaptive algorithms to determine subspace neuron numbers based on task complexity and optimizing computational efficiency through code enhancements or asynchronous parallel neuromorphic hardware could enable real-time adaptability for OCL.
\end{enumerate}

\smallskip
\textbf{Adaptation of Existing CL Methods in SNN Domain and Hybrid SNN-DNN approaches}: 
Exploring how existing ANN-based CL methods can be adapted to the SNN domain, to leverage the strengths of both architectures.
For instance, adapting the regularization methods and how it impacts the performance of SNNs trained with local learning rules under dynamic environments can be explored. 
Furthermore, the optimal memory size for replay-based method in SNNs that balances desiderata effectively under dynamic environments, can also be explored. 
Additionally, integration of hybrid SNN-DNN models, leveraging the strengths of both architectures, where DNNs handle feature extraction while SNNs process temporal or event-driven information to enhance CL performance in real-world applications can be explored. Moreover, knowledge distillation techniques like teacher-student to transfer knowledge from deep networks to SNNs, for efficient CL can be explored.

\smallskip
\textbf{Adaptation to Dynamic Input Patterns}: 
In dynamic environments, input data change unpredictably. 
Recent advancements leveraging STDP and active dendrites offer promising solutions. 
STDP adjusts synaptic strength based on spike timing, enabling efficient unsupervised learning and real-time adaptation without labeled data~\cite{bib296,bib297}. 
Meanwhile, active dendrites enhance adaptability by dynamically selecting task-specific sub-networks via a gating mechanism~\cite{bib318, devkota2024mtspark}. 
Despite these advancements, we identify that the existing techniques have not exploited temporal correlations in spiking patterns in inputs, hence hindering the network from achieving higher performance. 
To overcome this, there are several potential solutions, such as (1) adaptive learning with effective weight decay and weight updates, to improve the flexibility of neuron dynamics for adapting to input spikes; (2) clustering and representation learning based on the input patterns; and (3) adaptive algorithms to explore various configurations of dendritic segments in active dendrite techniques to optimize task-specific adaptation.

\smallskip
\textbf{Balancing CL Desiderata:} 
To meet multiple, often conflicting, requirements of CL is extremely challenging, requiring trade-offs to satisfy all CL desiderata (i.e., scalability, no/minimal usage of old data, task-agnostic, positive forward and backward transfer, controlled forgetting, and fast adaptation/recovery). 
For example, improving learning quality for past tasks while minimizing the use of old data is non-trivial.
Therefore, it is important to prioritize the desired characteristics based on the applications and balance them during the operational lifetime within acceptable performance.

\smallskip
\textbf{Input Noise Robustness}: 
Noise in input data, such as low contrast, background interference, or additive white Gaussian noise (AWGN), can obscure important features and significantly impair learning efficiency, particularly in dynamic environments where noise patterns vary unpredictably.  
A recent study addressed this issue by implementing a selective attention mechanism that focuses on task-relevant features, leading to enhanced performance in noisy conditions~\cite{bib108}. 
However, its reliance on specific noise types, such as AWGN, limits its generalizability to more complex, non-linear noise patterns. 
To overcome this, future research should consider various noise types and intensities, ensuring robustness in real-world scenarios. 
Adaptive algorithms capable of real-time adjustments to varying noise conditions may also enhance resilience. 
Furthermore, efficient pre-processing techniques and integration of high-resolution event-based sensors (e.g., DVS cameras) with conventional sensors (e.g., RGB camera) can also improve input data quality and robustness.

\smallskip
\textbf{Data Representation:}
Current NCL methods predominantly use rate coding to represent data, which is effective in some scenarios, but overlooks the temporal structure of spike trains. 
This limits the advantages of temporal information. 
Furthermore, studies on NCL methods with temporal coding is also under-explored. 
To address these challenges, future research may focus on incorporating temporal coding techniques with high information density (e.g., TTFS). 
Additionally, advancing hybrid encoding techniques that combine multiple encoding approaches may also improve performance, while optimizing latency and energy efficiency.

\smallskip
\textbf{Scalability:} 
Current CL methods incur significant memory and power/energy overheads when the networks get bigger, hence posing scalability issue. 
However, studies for realizing dynamic allocation of neurons without significant computational and memory overheads, as well as biologically inspired synaptic reorganization mechanisms are limited.
To address these challenges, several actions can be taken, as follows.
\begin{enumerate}
    \item Investigating the integration of more complex neuron models (e.g., Hodgkin-Huxley model) to enable richer representation of neuronal activity, thus potentially improving performance with good scalability.
    \item Exploring techniques to dynamically allocate neurons for new tasks while pruning redundant ones from previous tasks.
    \item Employing hierarchical or modular SNN design to enable scalable network architectures for CL.
    \item Investigating techniques for dynamic structural plasticity to enable reorganization of neural connections.
\end{enumerate}

\textbf{Learning Rule Developments:} 
Current NCL methods often rely on non-local learning rules. 
While variations and enhancements of STDP for OCL scenarios have shown promise, they remain under-explored. 
Additionally, current approaches primarily focus on a single learning paradigm, either supervised or unsupervised, which restricts the adaptability and overall performance of NCL methods. 
To address these limitations, future advancements should focus on developing advanced localized learning rules, such as adaptive variations of STDP, that enable SNNs to better retain past information while achieving quick learning convergence for new data in dynamic environments. 
Furthermore, hybrid learning approaches that integrate supervised and unsupervised paradigms could significantly enhance the training of SNNs, improving performance and adaptability in complex, real-world scenarios.

\textbf{Generalization Capability}: 
Ensuring that NCL methods can function effectively across diverse tasks and operational settings remains an open challenge. 
To address this, future advancements should focus on developing adaptive NCL methods capable of dynamically adjusting to varied environmental contexts, such as transitions from indoor to outdoor conditions. 
For this, expanding the evaluation of NCL methods to diverse datasets is important to improve generalization for complex tasks, thereby enhancing their robustness and versatility in real-world applications.

\begin{figure*}[t]
\centering
\includegraphics[width=\textwidth]{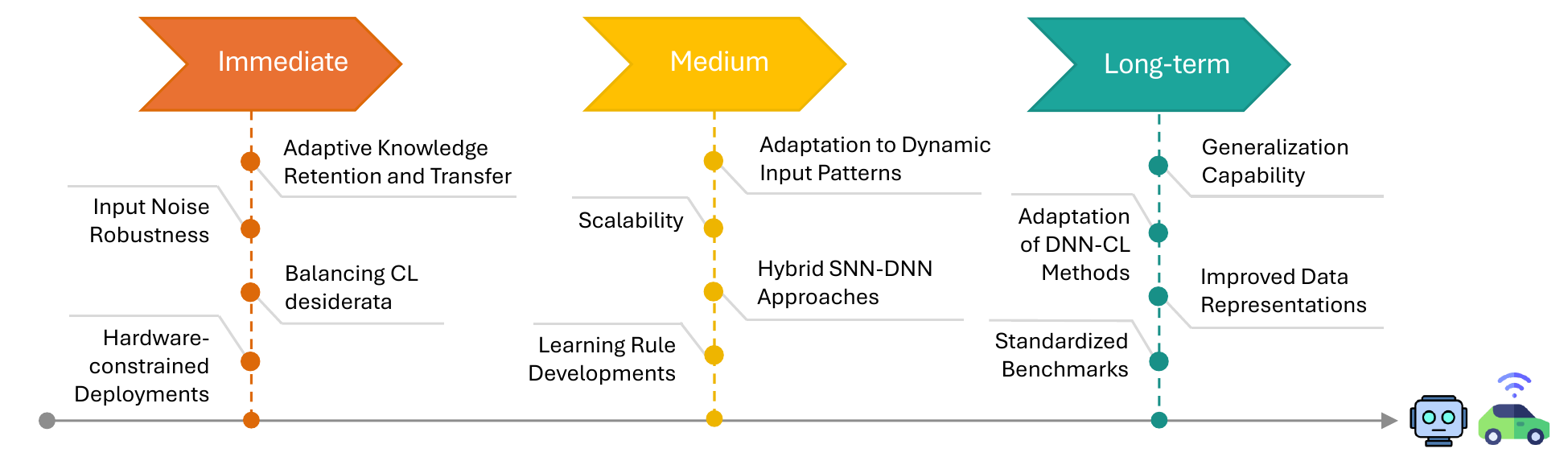}
\caption{Prioritized roadmap for NCL challenges in relation to real-world deployment.}
\label{roadmap}
\end{figure*}

\textbf{Standardized Framework, Evaluation Datasets, and Benchmarks}: 
Research on neuromorphic datasets for benchmarking NCL is still in its early stage, and the existing datasets still have limitations, such as:
\begin{enumerate}
    \item Inconsistent data pre-processing practices, e.g., varying time resolutions and image compression techniques, which impede fair comparisons across methods.
    \item Current datasets are predominantly small-scale and narrowly focused, limiting their generalizability and applicability to real-world tasks. 
    \item Data transformation from conventional to event-based data (spikes) may fail to capture temporal information, thus limiting SNN capabilities. 
    \item Most works do not specify the quantitative measurements of power/energy savings, latency and memory footprint for showing the NCL potential in practical real-world applications.
\end{enumerate}
To address these limitations, following actions can be taken.
\begin{enumerate}
    \item Developing standardized pre-processing protocols to ensure fairness and consistency in dataset comparisons.
    \item Creating large-scale, spatial-temporal event-based datasets tailored for diverse tasks, apart from the classification task, would improve benchmarking and enable the broader application of NCL methods.
    \item Constructing datasets emphasizing temporal dynamics, leveraging spike-based data generation.
    \item Task-specific performance metrics, such as accuracy for classification, MAE for regression, memory ratio and power/energy overhead for scalability, adaptability scores for functionality and robustness under noise perturbations or quantization. 
    \item Reporting and optimizing quantitative power/energy savings, latency and memory for showing the NCL potential in practical real-world applications.
\end{enumerate}

\textbf{Hardware Deployments}:
Deployments of NCL methods should leverage the inherent low-power and massively-parallel processing strengths of the neuromorphic chips, thus maximizing the efficiency benefits. 
For instance, algorithm mapping policies on the given hardware accelerators hold an important role in determining the processing latency and efficiency. 
Additionally, the frequently updating parameters such as neuron thresholds, decay rates, membrane potentials, and synaptic weights, align well with the capabilities of neuromorphic hardware. 
However, the hardware deployments remains under-explored.
Therefore, future research should focus on the following direction.
\begin{enumerate}
    \item Developing more efficient hardware designs that fully utilize SNN unique properties.
    \item Incorporating approximations in SNN parameters and weights can also help meet hardware constraints.
    \item Exploration of beyond-CMOS technologies (e.g., NVMs like RRAM, PCM, and MRAM) for energy-efficient hardware implementations is another promising direction. 
    \item Developing CIM systems with high-performance interfaces to support quick and frequent updates, minimizing latency and energy overheads.
    \item Developing processors for supporting the combination of unsupervised local learning with supervised global learning (e.g., combining different conventional \& neuromorphic chips). 
\end{enumerate}

To better align research focus with practical needs, we prioritize these challenges based on their relevance to real-world application domains such as robotics and autonomous vehicles/agents. 
Immediate ones include adaptive knowledge retention and transfer, input noise robustness, balancing CL desiderata and hardware-constrained deployments, each of which directly impacts system reliability and performance in dynamic environments.
Medium-priority challenges such as dynamic input adaptation, scalability, hybrid SNN-DNN approaches, and the development of energy-efficient learning rules will enable broader applicability. 
Finally, long-term ones such as generalization capability, improved data representations, adaptation of DNN-based CL methods, and standardized benchmarking frameworks, will be essential for building robust neuromorphic intelligence. A structured roadmap prioritizing NCL challenges in relation to real-world deployment is illustrated in Fig.~\ref{roadmap}.

%%%%%%%%%%%%%%%%%%%%%%%%%%%%%%%%%%%%%%%%%%%%%%%%%%%%%%%%%%%%%%%%%%%%%%%%
%%%%%%%%%%%%%%%%%%%%%%%%%%%%%%%%%%%%%%%%%%%%%%%%%%%%%%%%%%%%%%%%%%%%%%%%
\section{Conclusion} 
\label{sec_conclusion}

This paper presented a comprehensive survey of CL, reviewing the state-of-the-art works in both DNN and SNN-based methods. Regarding the DNN-based methods, the survey focused on the hardware deployment challenges and discussed the need for energy-efficient CL approaches. Then, it provided an extensive technical background of low-power neuromorphic systems covering its key aspects. Regarding the SNN-based CL methods (i.e., NCL), the survey categorized them and provided their comparison with relevant DNN-based CL methods, discussed and categorized hybrid approaches and focused on efficiency enhancement techniques currently used in literature. 
Moreover, it included the current progress of real-world NCL applications in adaptive robots and autonomous vehicles covering a wide range of use-cases, and provided quantitative analysis. Furthermore, it reported prominent neuromorphic datasets, metrics and benchmarks, emphasizing the need for standardized benchmarks and evaluation protocols, and suggested additional metrics for NCL. Among the reviewed methods, architecture-based like DSD-SNN and SOR-SNN achieved high accuracy on image classification benchmarks like MNIST, CIFAR100 and ImageNet. Methods incorporating compressed latent replay like Replay4NCL exhibited better performance in Class-IL scenario with significant latency and energy savings in speech classification. Similarly, more biologically inspired methods such as SpikeDyn and lpSpikeCon offered significant gains in memory efficiency and inference cost for low-power embedded systems. Additionally, hybrid SNN-ANN approaches are emerging as promising directions for enabling scalable and efficient CL. Despite the notable progress in the NCL field, there is still a pressing need for ongoing and significant innovation to fully realize the potential of low-power neuromorphic systems in the design of energy-efficient CL systems. The open challenges were discussed along with future directions at the end of the survey. Notably, CL strategies developed for DNNs, such as EWC, GEM, and ER are not directly transferable to SNNs due to differences in training dynamics, data encoding, and the lack of differentiability. Addressing this gap requires the design of SNN-compatible learning rules that preserve long-term knowledge while enabling online plasticity. Moreover, effective deployment of NCL methods on neuromorphic hardware remains constrained by architectural limitations, energy budgets, and on-chip learning capabilities for OCL. Future research should prioritize hardware-aware learning algorithms and standardized benchmarks to enable fair, reproducible evaluations across both domains.

%%%%%%%%%%%%%%%%%%%%%%%%%%%%%%%%%%%%%%%%%%%%%%%%%%%%%%%%%%%%%%%%%%%%%%%%

\section*{Acknowledgment}
This work was partially supported by United Arab Emirates University (UPAR) with Fund code 12N169. This work was also partially supported by the NYUAD Center for Artificial Intelligence and Robotics (CAIR), funded by Tamkeen under the NYUAD Research Institute Award CG010.

\appendices

\bibliographystyle{IEEEtran}
\bibliography{sn-bibliography}

\begin{IEEEbiography}[{\includegraphics[width=1in,height=1.25in,clip,keepaspectratio]{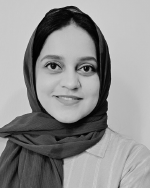}}]{Mishal Fatima Minhas} 
received the B.Sc. degree (Hons.) in computer systems engineering
from Mirpur University of Science and Technology (MUST), Pakistan, and the M.Sc. degree in electrical engineering with specialization in digital systems and signal processing (DSSP) from the School of Electrical Engineering and Computer Science (SEECS), National University of Sciences and Technology (NUST), Islamabad, Pakistan. She is currently pursuing the Ph.D. degree in
electrical engineering with the Department of Electrical and Communication Engineering, United Arab Emirates University (UAEU), Al Ain, United Arab Emirates. Her research interests include continual learning, brain-inspired AI, machine learning, energy-efficient design, neuromorphic computing, digital signal processing, digital system design, and formal verification.
\end{IEEEbiography}

%%%%
\begin{IEEEbiography}[{\includegraphics[width=1in,height=1.25in,clip,keepaspectratio]{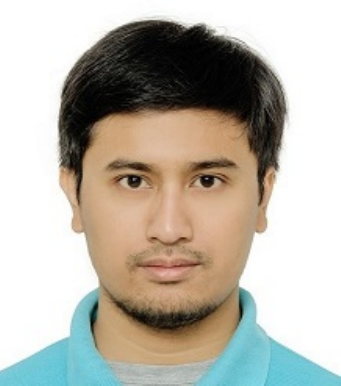}}]{Rachmad Vidya Wicaksana Putra}
(Member, IEEE) 
received the B.Sc. degree in electrical engineering and the M.Sc. degree in electronics engineering from Bandung Institute of Technology (ITB), Indonesia, and the Ph.D. degree in computer science from Technische Universität Wien (TU Wien), Vienna, Austria. 

In academia, he was a Teaching Assistant with the Electrical Engineering, ITB, a Research Assistant with the Microelectronics Center, ITB, and a Project Research Assistant with the Institute of Computer Engineering, TU Wien. Meanwhile, in industry, he also experienced working as an FPGA Engineer at PT. Fusi Global Teknologi, Indonesia, and TriLite Technologies GmbH, Austria. He is currently the Research Team Leader, eBrain Laboratory, New York University (NYU) Abu Dhabi, Abu Dhabi, United Arab Emirates. His research interests include neuromorphic and cognitive computing, computer architecture, integrated circuits and VLSI, system-on-chip design, emerging device technologies, robust and energyefficient computing, embedded AI, and electronic design automation. He received multiple HiPEAC paper awards and an ACM Showcase for his first-authored articles during the Ph.D. degree, and the Best Paper Nomination in ICARCV 2024.
\end{IEEEbiography}

\begin{IEEEbiography}[{\includegraphics[width=1in,height=1.25in,clip,keepaspectratio]{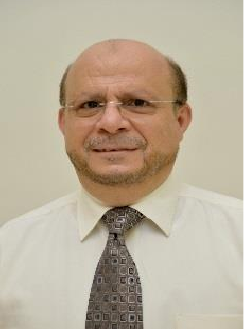}}]{Falah Awwad}
(Senior Member, IEEE)
received the M.Sc. and Ph.D. degrees in electrical and computer engineering from Concordia University, Montreal, Canada, in 2002 and 2006, respectively. He is currently a Professor with the Department of Electrical and Communication Engineering, College of Engineering, United Arab Emirates University (UAEU), Al Ain, United Arab Emirates, where he is also the Coordinator of the M.Sc. Electrical Engineering Program. He has co-authored one book and two book chapters and has published over 120 papers in journals and international conferences. His work has appeared in IEEE journals, Biosensors and Bioelectronics, Scientific Reports, and other reputable publications. He holds two U.S. patents. He has secured over 23 research grants, serving as the principal investigator on 17 of them. His service contributions include his roles as a consultant and a committee member at various universities and research institutions in United Arab Emirates. He has played a key role in developing and enhancing curricula for undergraduate and graduate programs in electrical engineering, computer engineering, and cybersecurity. He continues to advance his field through ongoing projects in the IoT security, energy-efficient computing, and advanced semiconductor devices. His research interests include VLSI circuits and systems, including hardware security, sensors, and biomedical applications.
\end{IEEEbiography}

\begin{IEEEbiography}[{\includegraphics[width=1in,height=1.25in,clip,keepaspectratio]{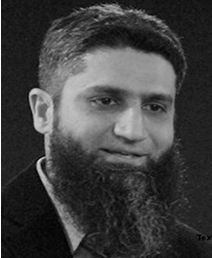}}]{Osman Hasan}
(Senior Member, IEEE)
received the B.Eng. degree (Hons.) from the University of Engineering and Technology Peshawar, Pakistan, in 1997, and the M.Eng. and Ph.D. degrees from Concordia University, Montreal,
Canada, in 2001 and 2008, respectively. 

He was an ASIC Design Engineer with LSI Logic Corporation, Ottawa, Canada, from 2001 to 2003, and a Research Associate with Concordia University, for 18 months after the Ph.D. degree. He joined the School of Electrical Engineering and Computer Science (SEECS), NUST, as an Assistant Professor, in September 2009. He was promoted to a Tenured Associate Professor and a Tenured Professor ranks in May 2015 and August 2019, respectively. He was the Head of the Department of Research, from 2015 to 2018, the Senior Head of the Department of Electrical Engineering, from 2018 to 2020, and the Principal and the Dean of SEECS, NUST, from 2020 to 2021. He is currently the Pro Rector (Academics) of NUST. He is the Founder and the Director of the System Analysis and Verification (SAVe) Laboratory, SEECS, and his research interests include embedded system design, formal methods, and e-health.
He has been able to acquire over Rs. 120 million of research grants from various national and international agencies and has published over 250 research articles, including six manuscripts, one patent, 20 book chapters, over 100 impact factor journal articles, and over 150 conference proceeding papers. 

Dr. Hasan is a member of ACM, the Association for Automated Reasoning (AAR), and Pakistan Engineering Council. He received the Quaid-e-Azam Award, Ministry of Education, Pakistan (1998), the Best University Teacher Award 2010 from HEC, Pakistan, the Ideal ICT Excellence Award 2012 from Ideal Distributions, Pakistan, the Excellence in IT Education Award 2013 from Teradata Pakistan, the Best Young Research Scholar Award from HEC, Pakistan (2011), the Excellence in IT Research and Development Award 2014 from Teradata Pakistan, the President’s Gold Medal for the Best Teacher of the University (2015) from NUST, the Best University Researcher Award (2015 and 2019), and the Research Productivity Award (2016) from Pakistan Council for Science and Technology. For his continued services in higher education, he was awarded the prestigious National Award of Tamgh-e-Imtiaz by the President of Pakistan in 2022.
\end{IEEEbiography}

\begin{IEEEbiography}[{\includegraphics[width=1in,height=1.25in,clip,keepaspectratio]{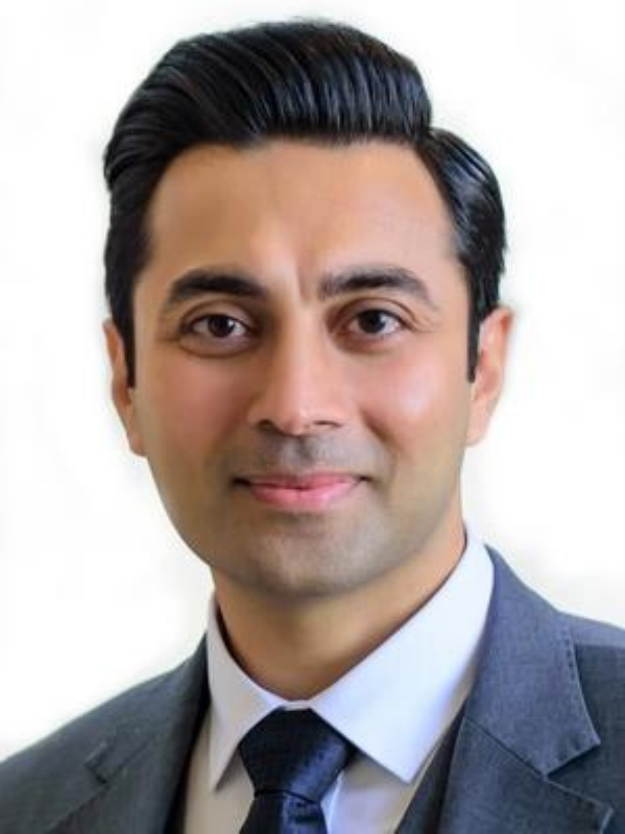}}]{Muhammad Shafique}
(Senior Member, IEEE)
received the Ph.D. degree in computer science from Karlsruhe Institute of Technology (KIT), Karlsruhe, Germany, in 2011.

He established and led a highly recognized research group at KIT for several years and conducted impactful collaborative research and development activities across the globe. In 2016,
he joined as a Full Professor of computer architecture and robust, energy-efficient technologies with the Faculty of Informatics, Institute of Computer Engineering, Technische Universität Wien (TU Wien), Vienna, Austria. Since 2020, he has been with New York University (NYU) Abu Dhabi, Abu Dhabi, United Arab Emirates, where he is currently a Full Professor and the Director of eBrain Laboratory. He is a Global Network Professor with the Tandon School of Engineering, NYU, New York, NY, USA. He is also a Co-PI/Investigator with multiple NYUAD centers, including the Center of Artificial Intelligence and Robotics (CAIR), the Center of Cyber Security (CCS), the Center for InTeractIng urban nEtworkS (CITIES), and the Center for Quantum and Topological Systems (CQTS). He holds one U.S. patent and has (co-)authored six books, more than ten book chapters, more than 350 papers in premier journals and conferences, and more than 100 archive articles. His research interests include AI and machine learning hardware and system-level design, brain-inspired computing, quantum machine learning, cognitive autonomous systems, wearable healthcare, energy-efficient systems, robust computing, hardware security, emerging technologies, FPGAs, MPSoCs, and embedded systems. His research has a special focus on cross-layer analysis, modeling, design, and optimization of computing and memory systems. The researched technologies and tools are deployed in application use cases from the Internet-of-Things (IoT), smart cyber–physical systems (CPSs), and ICT for development (ICT4D) domains.

Dr. Shafique is a Senior Member of the IEEE Signal Processing Society (SPS) and a member of ACM, SIGARCH, SIGDA, SIGBED, and HIPEAC. He has given several keynotes, invited talks, and tutorials, and organized many special sessions at premier venues. He served as the PC chair, the general chair, the track chair, and a PC member for several prestigious IEEE/ACM conferences. He received the 2015 ACM/SIGDA Outstanding New Faculty Award, the AI 2000 Chip Technology Most Influential Scholar Award in 2020, 2022, and 2023, the ASPIRE AARE Research Excellence Award in 2021, six gold medals, and several best paper awards and nominations at prestigious conferences.
\end{IEEEbiography}

\EOD

\end{document}